# AI based Algorithms of Path Planning, Navigation and Control for Mobile Ground Robots and UAVs

Jian Zhang

October 2021

# Abstract

As the demands of autonomous mobile robots are increasing in recent years, the requirement of the path planning/navigation algorithm should not be content with the ability to reach the target without any collisions, but also should try to achieve possible optimal or suboptimal path from the initial position to the target according to the robot's constrains in practice. This report investigates path planning and control strategies for mobile robots with machine learning techniques, including ground mobile robots and flying UAVs.

In this report, the hybrid reactive collision-free navigation problem under an unknown static environment is investigated firstly. By combining both the reactive navigation and Q-learning method, we intend to keep the good characteristics of reactive navigation algorithm and Q-learning and overcome the shortcomings of only relying on one of them. The proposed method is then extended into 3D environments. The performance of the mentioned strategies are verified by extensive computer simulations, and good results are obtained. Furthermore, the more challenging dynamic environment situation is taken into our consideration. We tackled this problem by developing a new path planning method that utilizes the integrated environment representation and reinforcement learning. Our novel approach enables to find the optimal path to the target efficiently and avoid collisions in a cluttered environment with steady and moving obstacles. The performance of these methods is compared with other different aspects.

In addition, another important navigation problem, reconnaissance and surveillance problem for UAVs, is studied and two algorithms are presented. It requires drones to fully cover the area of interest along their trajectories. In the first method, a two-phase strategy is presented and enable to operate with a given altitude. Furthermore, an occlusion-aware UAV reconnaissance and surveillance approach is developed, which takes both UAV kinematics constraints and camera sensing limitations into consideration. We have implemented all the proposed algorithms by illustrative computer simulations in different scenarios, and the results have confirmed the effectiveness of these approaches.

An extra study on the steering angle prediction algorithm for autonomous vehicles is presented. The proposed algorithm employs the convolutional neural network to extract features from the human driver and predicts the steering angle for autonomous driving. The performance of the algorithm is validated through simulations in different scenarios, we find the learned features can be transferred to the environment that has never been seen before.

# Contents

# Chapter 1

# Introduction

## Contents

## 1.1 Overview

As the demands of autonomous mobile robots are increasing in recent years, collision-free path planning/navigation problems have been the research subject of extensive researchers in the robotics field. It is an important problem in the navigation of mobile robots, which is to find an optimal collision-free path from a starting point to a goal in a given environment according to some criteria such as distance, time, or energy while distance or time being the most commonly adopted criterion [1].

They have become the necessary tools for a wide range of activities including but not limited to real-time monitoring, surveillance, reconnaissance, border patrolling, search and rescue, civilian, scientific and military missions, etc. Their advantage is unprecedented and irreplaceable especially in environments dangerous to humans, for example in radiation or pollution exposed areas. It is said as the third robotic revolution despite the fact that people generally assign them to a specific domain of the automotive industry [2].

It is undeniable that the space industry increasingly relies on technologies from other industries, especially the robotics industry. At the same time, the space industry could be a catalyst for the development of the robotics industry. Greater robotic independence and automation will be an important development. The space industry and the robotics industry are complementary to each other. In the future, there will be more and more valuable scientific research conducted by smart rovers and automation laboratories without direct interaction from humans, which will also benefit modern manufacturing. The mobile robots can be classified into ground mobile robots, flying robots, and underwater robots according to their working environments. We devote efforts to path planning for ground mobile robots and flying robots.

We assume that the ground robot studied in this research is a unicycle, which can be considered as a Dubins car [3] with non-holonomic constraint. Compared with the omnidirectional mobile robot model [4–7], the non-holonomic model is more practical.

## 1.2 Research Question

In this report, the main topic is to investigate path planning and control strategies for mobile robots with machine learning techniques and their applications, including ground mobile robots and flying UAVs.

As one of the most essential and important abilities of autonomous mobile robots, path planning has been and still is the focus of extensive research in the control field. The requirement of the path planning/navigation algorithm should not be content with the ability to reach the target without any collisions, but also should try to achieve a possible optimal or suboptimal path from the initial position to the target according to the robot's constraints in practice. So our first research question is how to improve the efficiency and effectiveness of the mobile robot in an unknown environment cluttered with obstacles.

Considering the unique nature and richness of Australia's biodiversity also carries national responsibility to protect and conserve native flora and fauna. A well-preserved environment is also the groundwork for the economy and in a way promotes economic growth. It requires constant vigilance against different risks (e.g. fires, resources extraction, goods movement, etc.) at massive scales. With the vast territory, sparse population on much of the continent, and land and sea-scape diversity, the challenge gets messier. Consequently, innovation tools are promising, like drones to perform reconnaissance and surveillance duty to help monitoring, intervention, remediation, and restoration at national, as well as local, scales. The Black Summer Bushfire in 2019-20 [1], as one of the most severe, has brought devastating, long-lasting impact on Australia. Under this condition, the development of an early bushfire detection system should absolutely be a priority. Although there exist satellite and a ground-based monitoring solution, the advantages of drones include lower cost of implementation, better resolution with more details, and high flexibility, which cannot be defeated. Our second research question is how to deploy drones to monitor the target area completely even under the influence of the occlusions.

---

[1]https://www.natureaustralia.org.au/what-we-do/our-insights/perspectives/australias-bushfire-crisis/

Autonomous vehicles, also known as robot cars, driverless or self-driving cars show great potential to change our society remarkably, not to mention the significant enhancements they could bring to the overall safety, efficiency, and convenience of transportation and transit systems. The final goal of the development of autonomous vehicles is to replicate the complex driving tasks of humans. However, there are still many problems that need to be settled efficiently, including arduous challenges, including obstacle perception, decision-making, and control. Thus, the full automation of vehicles is still on the road. The third question is how to facilitate the societal transition from human drivers to autonomous vehicles.

## 1.3 Contributions

We conduct extensive research from various perspectives to tackle the above questions. For the first research question, we develop collision-free path planning /navigation systems for mobile robots with machine learning techniques, which are targeting autonomous material movement and exploration applications. Two reactive navigation algorithms in the 2D environment and one in the 3D environment are developed for non-holonomic mobile robots (see Chapter 3, and 4). In particular, for the second research question, we provide an easily implementable method to estimate the minimum numbers and locations of waypoints to conduct the complete coverage for monitoring of the target area (see Chapter 5). In addition to that, we implement Dubins curves, Bezier curves, regular triangular patterns, and the clustered spiral-alternating method for route generation. Furthermore, the artifi-cial intelligence (AI) method is applied for route optimization (the shortest time possible). For the third research question, we provide a developed convolutional neural network to extract the features from the images and find the dependencies for forecasting the steering angle and throttle values for an autonomous vehicle (see Chapter 6). This report contributes to the area of path planning and control for mobile robots, such as ground robots, drones, and autonomous cars. The main contributions of this report can be described as follows:

- A hybrid reactive path planning algorithm for a non-holonomic ground robot is developed. Although there are reactive navigation algorithms, like randomized navigation algorithms that maneuver the robot in an unknown environment without any collisions, they cannot guarantee the optimality of the generated trajectories. In contrast, we introduce Q-learning in the decision-making process that can find out the shortest path efficiently. Compared with traditional Q-learning algorithms, our algorithm simplifies the complex calculation and converts the navigation problem into the selections of probability $p$ for every exit tangent point.

- A collision-free path planning algorithm for a non-holonomic flying robot is proposed. The objective is to enable the robot to reach the given destination by sense-and-avoid strategy without hitting any obstacles. This algorithm can be seen as a 3D extension of our hybrid reactive navigation algorithm. The obstacle avoidance is taken place on the avoiding plane constructed by the robot's heading and the direction to the center of the obstacle. While keeping a safe distance from obstacles, the presented control algorithm can navigate the robot via the optimal path successfully.

- The proposed hybrid reactive path planning algorithm is then modified for an unknown dynamic environment. One of the main benefits of our method is that the robot can find the path to the target and avoid collisions without any information of the obstacles' shapes or velocities. The selected path is efficient and effective in terms of minimum time, maximum safety. The robot learns from the environment by numerical evaluation function that assigns actions with different values.

- Two path planning algorithms for reconnaissance and surveillance are proposed, which ensure every point on the target ground area can be seen at least once in a complete surveillance circle. Besides, the geometrically complex environments with occlusions are considered in our research. Compared with many existing methods, we decompose this problem into a waypoint determination problem and an instance of the traveling salesman problem.

- A deep neural network is modified to tackle the path planning problem for an autonomous car. The objective of the convolutional neural network is to extracts the features from the images automatically and finds the dependencies for forecasting the steering angle to keep the vehicle running at the center of the lane. With the minimum training data from a human driver, the model learns how to drive in a known or unknown environment. Moreover, the learned features can be transferred to environments that have never been trained before. Compared with decomposing this problem into smaller pieces, such as lane detection, path planning, and control, this model optimizes all processing steps simultaneously.

- The performance and effectiveness of all algorithms have been confirmed by illustrative computer simulations in different scenarios. The simulation result demonstrated the effectiveness of the proposed algorithm by comparing it with other effective algorithms in the same environment.

- The collision avoidance, kinematic constraints of moving robots, and constraints of sensing are taken into consideration in all algorithms, which is more practical in real-time application.

## 1.4 Organization

The report is organized into seven chapters (see Figure 1.1):

Chapter 2 presents a comprehensive review of related work in the field of path planning for mobile robots, specifically different types of mobile robots, navigation, and machine learning. Chapter 3 provides a novel hybrid reactive navigation strategy for a non-holonomic ground mobile robot in an unknown cluttered environment. Our strategy combines both reactive navigation and Q-learning method. We intend to keep the good characteristics of the reactive navigation algorithm and Q-learning and overcome the shortcomings of only relying on one of them. This strategy is then extended in a 3D environment, and a sense-and-avoid type collision-free path planning algorithm is proposed. The flying robot traverses the clutter environment with static obstacles while keeping a safe distance from them. Compared with the last chapter, a more challenging environment with moving obstacles is considered in this chapter. In Chapter 4, an algorithm that combined the positive characteristics of the integrated representation of the environment and reinforcement learning is described. With our control algorithm, no approximation of the shapes of the obstacles or even any information about the obstacles' velocities is needed. Our Q-learning based algorithm can interact with the environment even with pedestrians walking around, and still be able to find the optimal path to the target efficiently and avoid collisions. Extensive computer simulations have been applied to confirm the performance of our path planning strategy. Following is Chapter 5, we extend the path planning research in the last two chapters into the complete coverage problem in the reconnaissance and surveillance of UAVs. Compared with other approaches, we take problems caused by occlusions into consideration. The quality of the generated trajectories is assessed by comparing them with two other algorithms. Then we employ the deep convolutional neural network for steering angle prediction in Chapter 6. With the trained model, the car can drive in both known and unknown environ-ments. One of the benefits of our method is to release the pressure of tremendous amounts of hand-coding. Instead, our model learns from the human driver. Lastly, Chapter 7 concludes this report and provides some possible directions for future work.

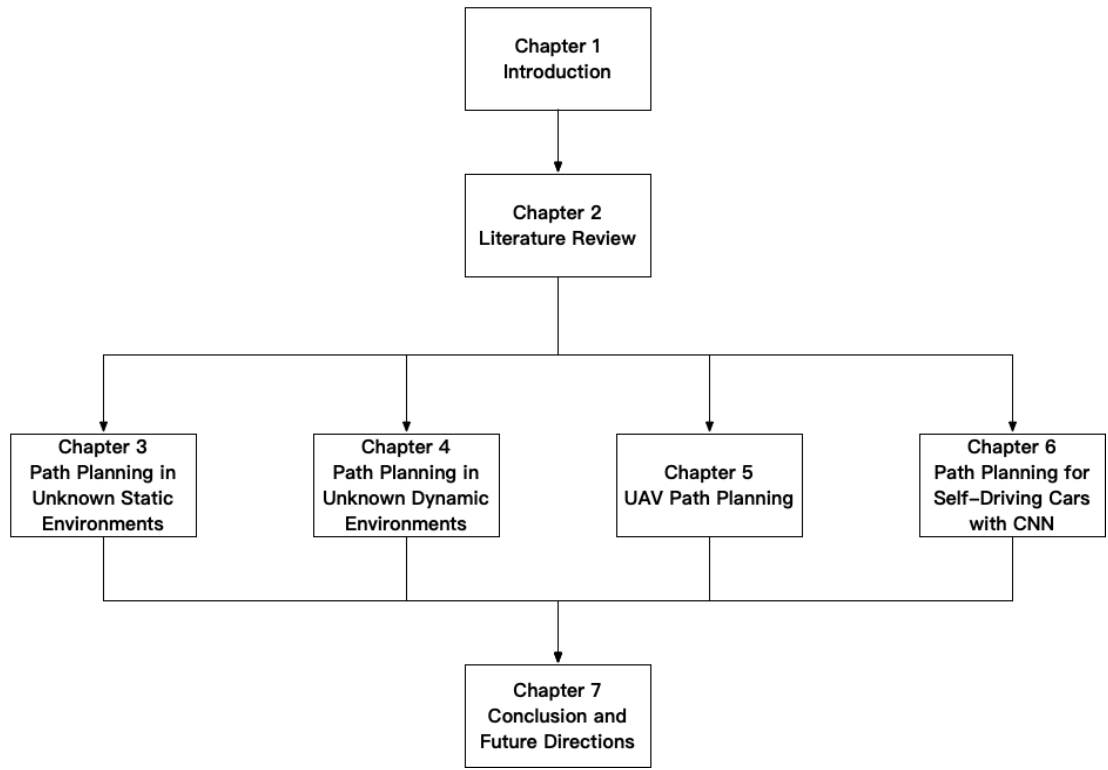

Figure 1.1: Interdependence of Chapters

# Chapter 2

# Literature Review

## Contents

## 2.1   Overview

Mobile robots have been applied to execute a wide range of crucial unmanned missions for civilian, scientific, and military usage over the past decades. Path planning has been one of the most important and fundamental research fields for mobile robots including ground mobile robots and flying UAVs since the 1960s. In this case, people cannot be satisfied with the simple ability to reach the target without collision, but also search for an optimal or suboptimal path from the initial state to the target state according to various criteria and kinematic constraints in practice. The development of path planning algorithms and machine learning techniques provides more possibilities for implementation of hybrid approaches.

Due to the breadth of the report, we only provide a survey of work related to our research problems mentioned in the last chapter, i.e., how to improve the effectiveness and efficiency of the generated path in an unknown environment, coverage problem for drone deployment with the impact of obstacle occlusions, and path planning problem for autonomous vehicles. We refer readers to [8–19] and references therein for surveys of corresponding extensive literature. The classification of the path planning and the positions of the proposed methods are shown in Figure 2.1. The flow chart Figure 2.2 shows the developing process of global/reactive path planning. The environmental modeling transforms the obtained environmental information into characteristic map information. The path search algorithm uses known environmental information to find a path from the initial state to the target state according to a variety of optimization criteria and kinematic constraints such as travel time, path length, speed and turning radius.

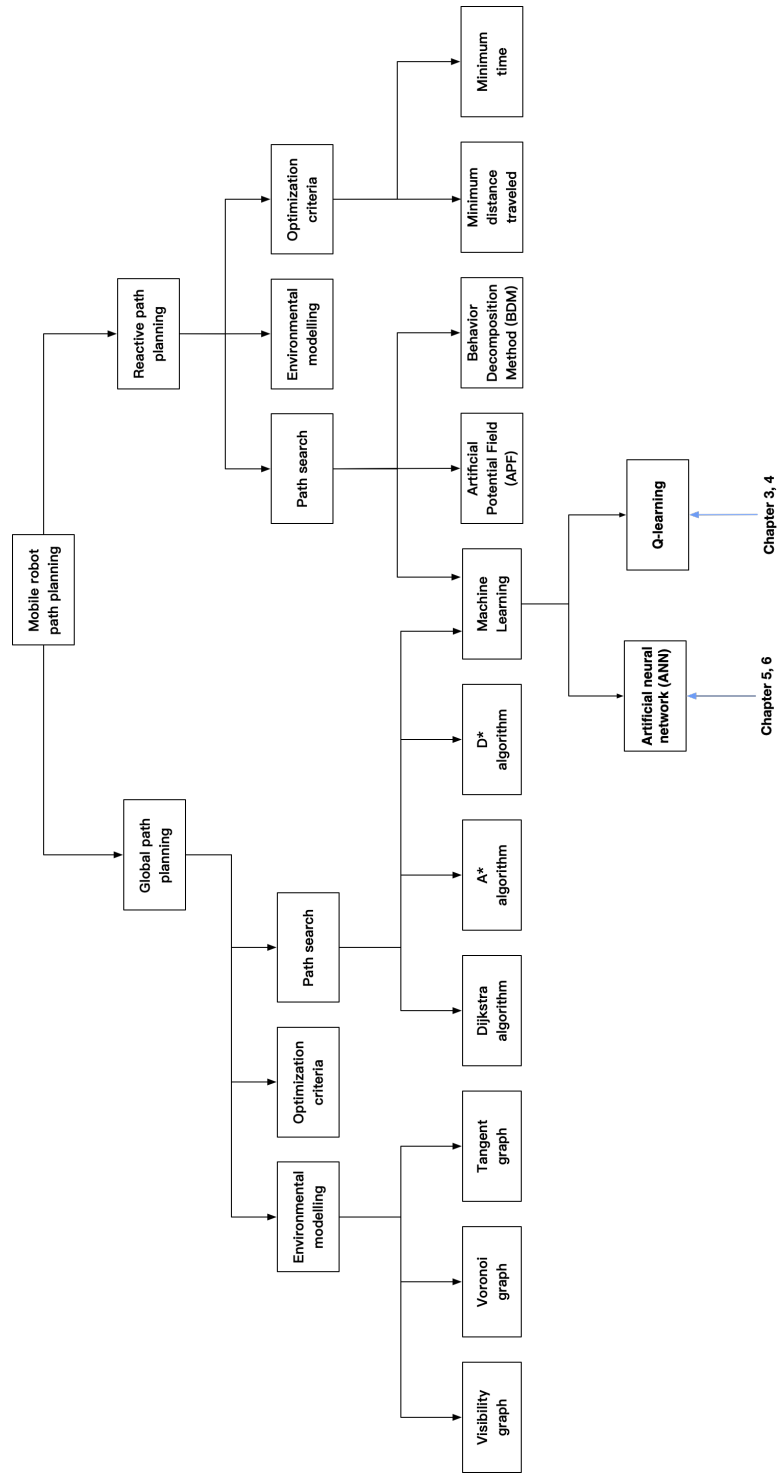

Figure 2.1: The classification of the reviewed approaches and the positions of our contributions

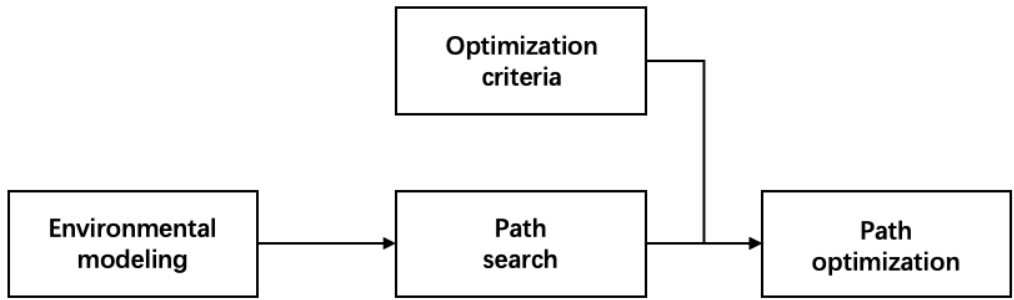

Figure 2.2: Developing process of mobile robot global/reactive path planning

## 2.2 Kinematics of Mobile Robots

The kinematic models of the most mobile robots can be divided into holonomic and non-holonomic model according to different kinematic constraints, see Figure 2.3 for examples.

- Holonomic model: All degrees of freedom of this kind of model are controllable, which has control capability in any direction. These constraints can be integrated into the position constraints. Common holonomic models include omnidirectional mobile robots and drones. The holonomic constraint can be mathematically expressed through positional variables, which imposes the relationship between the configuration of the model and its velocity, see [20–22] and references therein.

- Non-holonomic model: This kind of model has constraints that cannot be integrated into positional constraints, and requires a different relationship like the derivative of positional variables (direction and/or magnitude of velocities) [23, 24]. The non-holonomic models investigated in this work involved the unicycle-like model and car-like vehicle. The unicycle-like ground model consists two independently driven wheels on a common axis and one or more passive/castor wheels [25–27]. A typical car-like vehicle have a pair of steerable front wheels and a pair of fixed rear wheels. The maximum turning rate is proportional to the speed of the vehicle see [28–30]and references therein.

## 2.3   Unmanned Aerial Vehicles (UAVs)

The coverage control of mobile networked systems (e. g. UAVs for reconnaissance and surveillance) is important due to the growing use, especially in the defence, agriculture, and environment.

Unmanned Aerial Vehicles (UAVs), also known as aerial drones, are becoming increasingly present in our everyday lives [11]. Their extensive use recently jumped from military to hobby and professional applications [31]. They have become the necessary tools for a wide range of activities including but not limited to search and rescue [32,33], 3D reconstruction [32], delivering goods and merchandise, serving as mobile hot spots for broadband wireless access, and maintaining surveillance and security [34,35], infrastructure inspection [36], border patrolling [37], etc. The authors in [32] presented a vision-based autonomy system for UAVs equipped with bioradars that conduct USAR (UAV search and rescue) operations in post-disaster struck environments. Their autonomous navigation and landing system enables the UAV to localize itself, build three-dimensional occupancy maps, plan collision-free trajectories and autonomously explore highly complex unstructured environments while reconstructing a three-dimensional textured mesh of the scene. Li and Savkin [38] proposed a wireless sensor network-based collision-free navigation method for micro-UAVs in the Industrial Internet of Things (IIOT). A wireless sensor network (WSN) consisting 3-D range finder is involved to detect the static and dynamic obstacles in

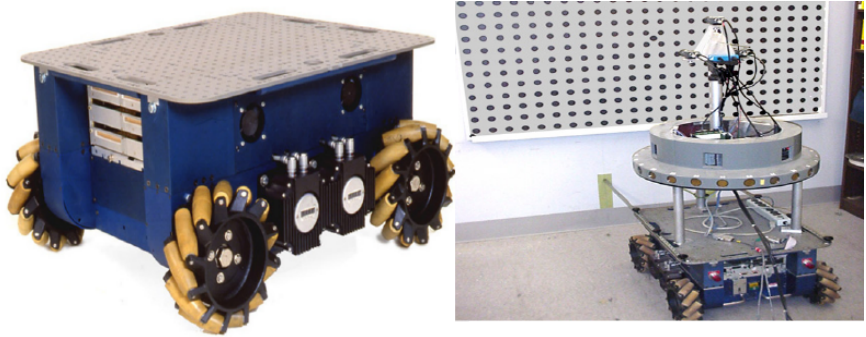

(a) Four-wheel omnidirectional robot "Uranus"[1]

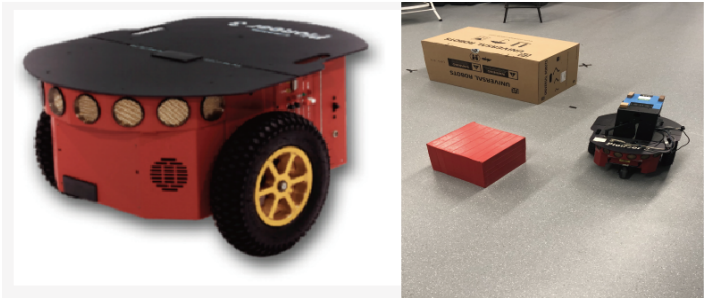

(b) The Pioneer 3-DX robot

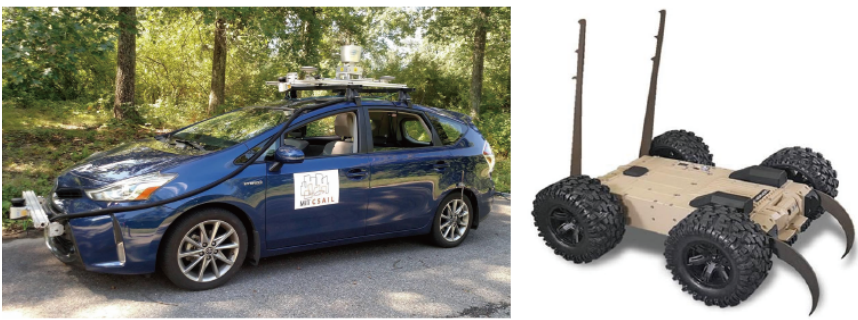

(c) MIT's self-driving car[2], and Car-like "Scorpio" Robot[3]

Figure 2.3: Robot examples of different kinematics

the workspace so that to reduce the weight of micro-UAVs. In general, the coverage problem was first put forward by [39] over a 2D grid environment. One important extension of this problem is for the UAVs, which requires new considerations due to battery life, mobility, and obstacles may cause occlusions. It can be generally classified into two main categories by different UAV motions. Some researchers [40, 41] focus on the deployment of the hovering UAVs to reconnoiter over certain terrains (static coverage). Savkin and Huang [41] focuses on monitoring every point on the target area while the UAVs keep hovering at certain locations respectively during all the mission time. While the standard static coverage problems typically pay attention to achieve complete or full coverage and track the intruders within it, some interesting research like [42] monitors the routes to access into the target area, which can be seen as an intrusion prevention and detection scheme. Another important problem is the reconnaissance and surveillance problem by moving UAVs (dynamic coverage) [43–45]. [45] developed a path planning strategy to maximize the coverage of the area of interest, and track multiple moving ground targets to avoid the surveillance of the UAVs. Savkin and Huang [46] proposed a distributed coverage-maximizing algorithm to find locally optimal positions of UAVs with the objective of maximizing the quality of coverage. Subject to that, the drones maintain a connected communication graph with the ground nodes.

In the common reconnaissance and surveillance scenario, the flying UAV equipped with a downward-facing video camera with a certain visibility angle can monitor the targets of interest on the ground, like vehicles, humans, animals, etc. [37,45,47]. The surveillance quality can be evaluated in terms of coverage and resolution [48]. As the video camera can only see the points within its cone-shaped field of view (FOV), A *full coverage* requires every point on the target area can be seen at least once in the complete surveillance circle. Rather than ideal flat terrain, in this research, we concern about a more challenging and realistic variant of this problem that reconnaissance and surveillance in geometrically complex environments, such as mountainous terrains and urban regions. Under this condition, the FOV can be reduced by any kind of obstacles, like mountains, hills, buildings, walls, etc. Furthermore, the lower altitude of the traveling path is preferred for a better resolution of the observed region of the terrain. This problem is likely to become especially significant for the small and micro unmanned aerial vehicles (SUAVs and MUAVs). Consequently, the optimized path planning for surveillance is indispensable to deliver outstanding performance encountered with the complex environment and limited resources. Although the occlusion-aware UAV path planning is addressed in [49], the resolution requirement is not considered, as all the camera locations are at the same altitude. Authors in [50] also assumes that the drones fly on the same height without the influence of the obstacle occlusions. Cheng *et al.* [51] focus on the problem to cover a 3-D urban structure using a single UAV flying around circular trajectories, however, the trajectory may not be optimal. Govindaraju *et al.* [52] proposes a probabilistic visibility model to identify near-optimal observation locations for UAV surveillance with both complete and partial occlusions.

## 2.4 Autonomous Vehicles

The autonomous vehicle is also known as the robot car, driverless or self-driving car. It is expected to mitigate driver's workload, reduce traffic accidents, and reduce traffic congestion. The profound impacts of autonomous vehicle technology could change our society remarkably, not to mention the significant enhancements they could bring to the overall safety, efficiency, and convenience of transportation and transit systems.

It dates back to 1920s, when the concept of self-driving car was first introduced by radio-controlled cars. In 1989, Pomerleau [53] pioneered the use of the neural network for autonomous vehicle navigation by building the Autonomous Land

Vehicle in a Neural Network (ALVINN) system. The neural network-based semi-autonomous vehicle model could predict the steering angle directly from pixel inputs in simple driving scenarios with obstacles. However, this model demonstrated its unlimited potential of implementing neural networks for self-driving vehicles and laid the foundation for contemporary control strategies. Caltagirone *et al.* [54] integrated LIDAR point clouds, GPS-IMU information, and Google driving directions to generate driving paths. The system is based on a fully convolutional neural network that jointly learns to carry out perception and path generation from real-world driving sequences and that is trained using automatically generated training examples. Bojarski *et al.* [55] created a neural-network-based system, known as PilotNet, which outputs steering angles given images of the road ahead. Besides g the obvious features such as lane markings, edges of roads, and other cars, PilotNet can learn more subtle features. Wang *et al.* [56] proposed a reinforcement learning-based approach to train the vehicle agent to learn an automated lane change behavior using a Deep Q-Network (DQN). Authors in [57–59] reviewed a wide range of research related to the control strategies for autonomous vehicles with the deep learning method.

'We are still at the 'horseless carriage' stage of this technology, describing these technologies by what they are not, rather than wrestling with what they truly are." [60] stated in 2012. With Extraordinary amounts of time, money, and effort invested from numerous companies including leading names both in tech and auto-making and the crucial breakthrough in AI since then, we seemed to see signs of hope. Companies such as General Motors, Google's Waymo, Tesla, BMW, Mercedes, Audi, and Lexus, Toyota, and Honda once announced that they would make autonomous vehicles by 2020 or sooner. But now, in 2021, we find that almost every company or institute rolled back their predictions, and the fully autonomous car is still out of the reach.

In general, the architecture of the autonomy system of self-driving cars is typ-ically organized into the perception system, and the decision-making system [18]. The perception task includes obstacle detection, localization, mapping, and decision-making task considers path planning, behavior selection and control, etc. The mounted sensors are indispensable to environment perception [61]. Sensory information is provided as input to complex models to identify details about its surroundings and generate appropriate navigation paths. The main sensor types include lidars [62], radars [63], optical cameras [53, 64] and ultrasound [65]. According to the harmonized classification system established by the Society of Automotive Engineers [66], the level of autonomous capabilities of autonomous cars is classified into six levels from "no automation" to "fully automation".

- Level 0: No Automation

- Level 1: Driver Assistance

- Level 2: Partial Automation

- Level 3: Conditional Automation

- Level 4: High Automation

- Level 5: Full Automation

There are technological limitations regarding its real-world implementation need to be solved, including but not limited with the vehicle-pedestrian interaction [67–70], lane change maneuvers [71–74], X-by-Wire driving in road condition [75–78], hardware issues [79], software failures [80, 81], rough weather conditions [82, 82, 83], light conditions [82], wheel slips caused by slippery and hazardous road conditions [84–86], and network connection [87, 88], etc. High degree of reliability in complex urban traffic is needed.

Researchers with the current technology have committed to tackling these questions. Clamann *et al.* [67] studied the intent communication between vehicles and pedestrians, and found the pedestrians still focus on legacy behaviors such as velocity rather than leverage the information on an external display. We refer readers to [68–70] for further reading. Researchers in [74] developed a dynamic potential field-based model to generate intention of lane-change (LC) maneuver for autonomous vehicles, and designed an integrated path planning and tracking control algorithm to evaluate and conduct safety of the LC process. [75] presented an adaptive sliding-mode (ASM) control methodology for a vehicle steer-by-wire (SbW) system. Zhang *et al.* [84] designed a nonlinear robust wheel slip rate tracking control law with lumped uncertainty observer is derived via the Lyapunov-based method. Matveev *et al.* [85] proposed two guidance laws for autonomous vehicles subjected to tire slips with a sliding mode approach and smooth nonlinear control. Falcone *et al.* [86] presented a model predictive control (MPC) method to compute the front steering angle in order to follow the trajectory on slippery roads such as snow-covered roads at the highest possible entry speed.

With numerous problems and challenges waiting to be solved, we still need decades until the self-driving car to be a part of our everyday lives. Litman [89] indicated that it will be at least 2045 before half of the new vehicles are autonomous, 2060 before half of the vehicle fleet is autonomous, and possibly longer due to technical challenges or consumer preferences. We must anticipate how new technologies and services are likely to affect road, parking, and public transit needs, and how to respond to minimize problems and maximize total benefits [89].

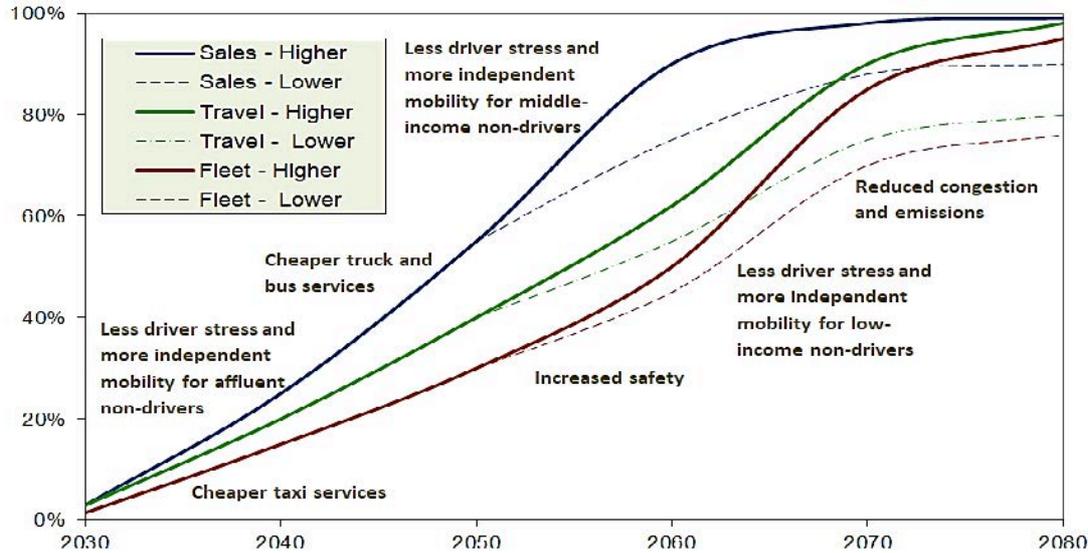

Figure 2.4: Autonomous vehicle sales, fleet, travel and benefit projections by [89]

## 2.5 Global Path Planning

In general, path planning can be divided into global path planning and reactive (local) path planning according to whether the *a priori* information of the environment is known or not. All the necessary information is known to the robot at the beginning for the global path planning. In contrast, reactive (local) path planning uses sensors to observe only part of the unknown environment each time.

### 2.5.1 Environmental Modeling

The common environmental modelling methods include visibility graph, Voronoi graph, tangent graph, cell decomposition approach, and probabilistic roadmap method.

### 2.5.1.1 Visibility graph

The principle of the visibility graph is shown in Figure 2.5. This technique was introduced by [90] in 1968 as the planning method for the famous "Shakey" robot in 1979 [91]. The mobile robot can move along edges that link all of its of its visible vertices of the obstacles. The shortest path may be obtained by exploring the visibility map. However, the efficiency of this method will be reduced significantly with increasing complexity of the problem. In this case, the path planning problem in 3D or higher is a NP-hard problem [92]. Some visibility graph methods [93–95] are provided for reader's reference.

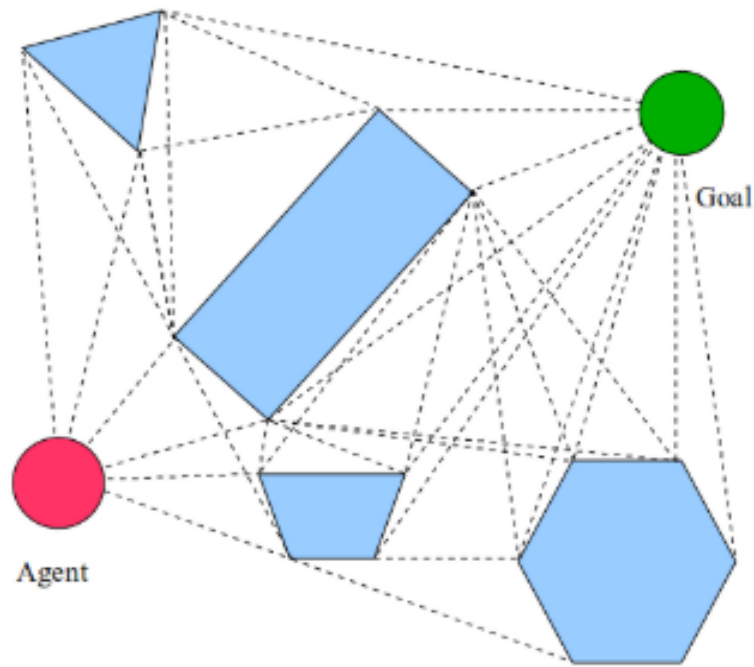

Figure 2.5: A sample visibility graph in [96]

### 2.5.1.2 Voronoi graph

In the Voronoi graph, the set of vertices are equidistant from at least three obstacles, and the edges are formed from are equidistant from at least two obstacles. The robot explore the free space of the environment [97,98]. Figure 2.6 shows the Voronoi graph of seven sites. Different from the visibility graph, the Voronoi graph keeps the robot to be as far away as possible from nearby obstacles. Generally speaking, Voronoi graph has fast calculation speed, but may also cause more mutational sites. The complexity of this method is affected by the obstacles and the constraints of the studied space.

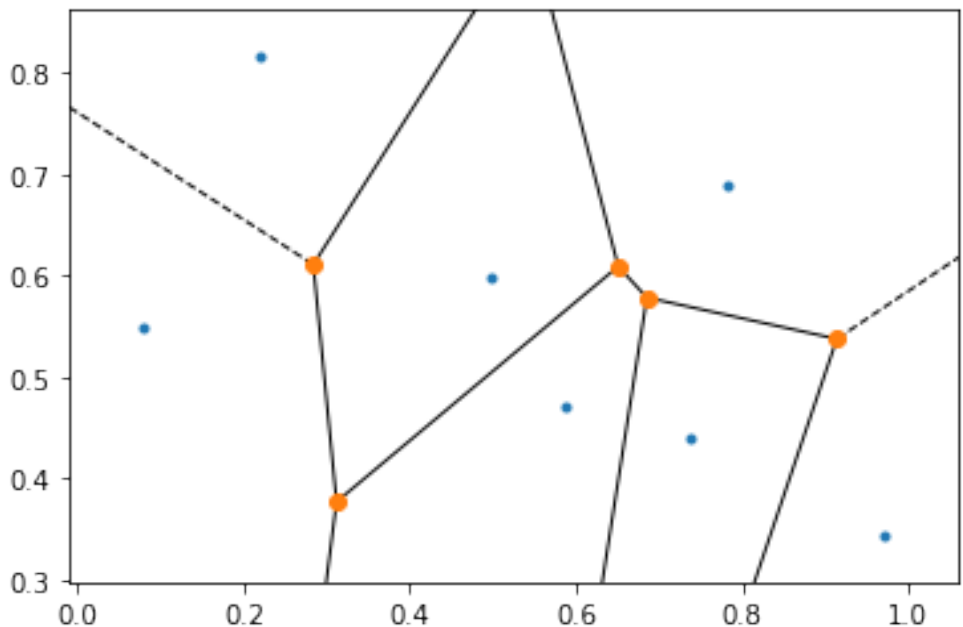

Figure 2.6: Voronoi graph

### 2.5.1.3 Tangent graph

The nodes of the tangent graph (Figure 2.7) are tangent points on the obstacles' boundaries, and the edges are conflict-free common tangents between the tangent points. The problem of finding the shortest path can be simplified to a finite path search problem as the shortest path consists of tangent graph [4, 99–103].

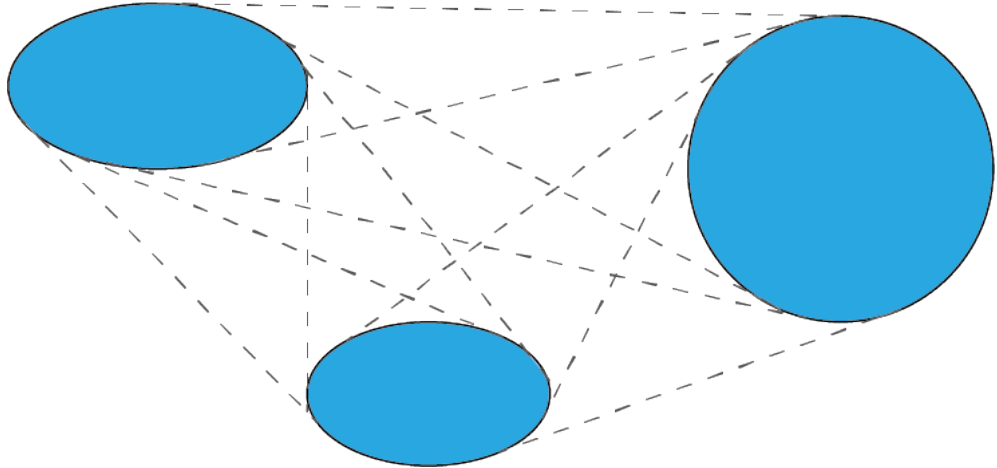

Figure 2.7: Tangent graph

## 2.5.2 Path Search Algorithm

The common global path planning algorithms such as Dijkstra algorithm, A* algorithm , D* algorithm and many more have been developed to handle the path search problem for mobile robots.

Dijkstra algorithm is an efficient breath-first search method to find a least-cost path between nodes in a graph. It was conceived by E.W. Dijkstra in 1959 [104]. The evaluation function of this method can be written as $f(n) = g(n) + 0$, which uses a negligible estimate of the distance to the goal. All vertices are divided into two groups, the first group is used for vertices of the shortest path, and the second group is used for vertices that have not been included yet. The initial state of the shortest path set only contains the starting point, while the another set initially includes all vertices except the initial point. Every time, a vertex closest to the starting point from the second set will move to the first set, and the path length will be minimized. The path length is the sum of the weight of edge from the starting point to the node. The search ends when all vertices are included in the first set. [105–107] have presented the path planning strategy for mobile robots using with Dijkstra algorithm. The main disadvantage of this method is that it does not use the information obtained from the environment, such as the location of the goal, so it will explore in all directions (uniformed search).

As an extension of Dijkstra algorithm, A* [108] is defined as the best-first algorithm. It is the most known path planning algorithm, which can be applied to the metric and topological configuration space. The principle of A* algorithm is to evaluate each state through the evaluation function $f(n)$, $f(n) = g(n) + h(n)$. $h(n)$ represents the heuristic estimate of the minimum cost from state $n$ to the goal state, and $g(n)$ is the current accumulated cost between the initial state to state $n$. Generally, the Manhattan or the Euclidean distance is used when calculating the heuristic cost. The evaluation function $f(n)$ notably represents the minimum estimated cost from the initial state to the goal state passing through state $n$. A* algorithm is more effective than the Dijkstra algorithm in path search, because it considers the location of the goal and mainly explores the goal state (informed search). Nonetheless, which may exist several states with the minimum $f(n)$ value, which can not guarantee the optimality of the generated path.

There also exist the extensions of A* algorithm like D* algorithm [109,110], field D* algorithm [110,111], Focussed D* [112],LPA* (Lifelong Planning A*) [113], D* Lite [114], basic Theta* algorithm [115], and Phi* [116].

D* algorithm [109] stands for the dynamic version of the A* algorithm, which means it can change the cost of arc while operating (online replanning). The backward path is searched from the goal to the current position. Every state expect goal state to have a *backpointer* to the next state, so that the cost from each state to the the goal state is known. The calculation ends when the next state to be evaluated turns out to be the start state. D* algorithm is more efficient than the A* algorithm in complex environments as it can ignore the influence of local changes in the environment and avoid the cost of backtracking computation. However, the high cost of memory is the main disadvantage of D* algorithm.

The details of the intelligence algorithms will be illustrated in the next section.

## 2.6 Reactive Path Planning

Due to the limitation of environmental information, the problem of reactive collision-free navigation is challenging. In order to reach the target in a complex environment without collisions, the robot should be able to detect and avoid obstacles. The environmental modeling refers to the last section.

### 2.6.1 Optimization Criteria

There are many optimization criteria can be implemented in path planning algorithms considering different circumstances. The commonly used options are listed in the following:

- Minimum distance: This standard is widely applied in many path planning strategies as it can be separated from the vehicle's speed profile. In this circumstance, the results are available for vehicles including but not limited with Dubins vehicles [117, 118], omnidirectional vehicles [119, 120], and actuated speed vehicles [121, 122]. Zhang [101] constructes the tangent graph of numbers of possibly non-convex obstacles with constraint on the curvature of their boundaries and searches the shortest path through these edges.

- Minimum time: This criterion is conducted with extensive research because it is highly related to productivity in industrial environments. The minimum traveling time path planning problem is considered under the kinematic and dynamic constraints of the vehicle. It can be combined with phase plane [123, 124], dynamic programming [125, 126], etc.

- Optimal surveillance frequency: This metric can also be described as the information age of the surveillance area over a period of time [127]. The goal is to minimize the average time of collecting information, so as to monitor the target area more frequently.

- Minimum Energy: The energy criterion is important in the following situations. It is not just about economic considerations, but crucial limited energy environments such as spacecraft and submarines. At the same time, smoother trajectories relieve the pressure of both the robot's mechanical structure and actuators [128, 129].

### 2.6.2 Path Search Algorithm

The path search algorithms discussed in this section include the behavior decomposition methods, artificial potential field method, and machine learning algorithm.

#### 2.6.2.1 Behavior Decomposition Method

Behavior Decomposition Method (BDM), also called behavior-based path planning. This common method of local path planning decomposes the whole behavior tasks into basic behavior units. Generally speaking, behavior decomposition is a method that breaks down the complicated path planning problem into several smaller independent units, such as target pursuit, collision avoidance, etc. Each unit has its own sensors and actuators, and the ultimate goal is achieved by different units' cooperations. Gilimyanovr *et al.* [130] show the complex behaviors can be feasible through the coordination of simple behaviors. The study in [131] presents a solution to the parking problem of a non-holonomic mobile robot with a stable switching control strategy. The favored operation results can be obtained through sub-behaviors with fast processing speed.

### 2.6.2.2 Artificial Potential Field

Artificial Potential Field (APF), also known as potential field, was introduced by [132]. It is one of the simplest reactive path planning methods, in which the robot is regarded as a point under the influence of the artificial potential field. The mobile robot is subjected to the attractive force and repulsive force. The mobile robot moves to the goal by the attractive force from the goal point and avoid collisions by the repulsive force from the obstacle. The advantages of APF include simple operation, easy implementation, fast calculation speed and the ability to generate the safe path to the goal with real-time requirements. Another disadvantage of the basic A* algorithm is that it takes a long time to execute, while the APF is one of the fastest path search methods. Ren *et al.* [133] and Triharminto *et al.* [134] apply APF to the non-holonomic model. The destination is reached by the gradient descent method to reduce the potential while avoiding the obstacles. However, this method suffers from the local minima and the complexity of the model. The local minima may occur when the attractive force and the repulsive force cancel each other. Thus this method may not perform well in congested environments. Besides, it can be difficult to find the optimal path by tuning the parameters. Many researchers were committed to remove the local minimum limitations. Techniques like "random motion," "valley-guided motion," and "constrained motion" are presented in [135] to escape the local minimum. Masehian *et al.* [136] combine Voronoi graph, visibility graph, and potential field to solve the path planning problem. A parametric tradeoff between safety and length is provided, and the generated path is shorter than the Voronoi graph and potential field methods, and faster than the visibility graph. Other variants of APF were published in [137–139].

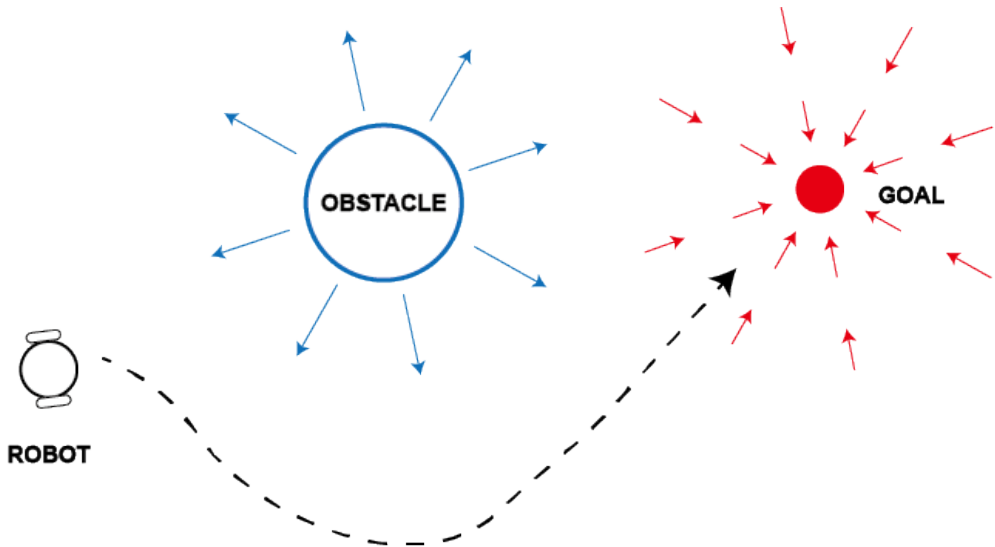

Figure 2.8: The scheme of APF

### 2.6.2.3 Machine Learning

The contents of machine learning algorithms would be illustrated in detail in the next section.

## 2.7 Machine Learning Algorithm

## 2.7.1 Artificial Neural Network (ANN)

Artificial Neural Networks (ANNs) or Neural networks (NN) are computational processing systems inspired by the biological nervous system (refers to the human brain), which consists of interconnected processing elements (refers to neurons) via weights. As a popular framework to perform machine learning, neural networks aim to recognize complicated patterns from data and make intelligent decisions. The path planning problem can be seen as a map between the perceptual space and action space. Thus researchers tried to see this problem from the perspective of neural networks.

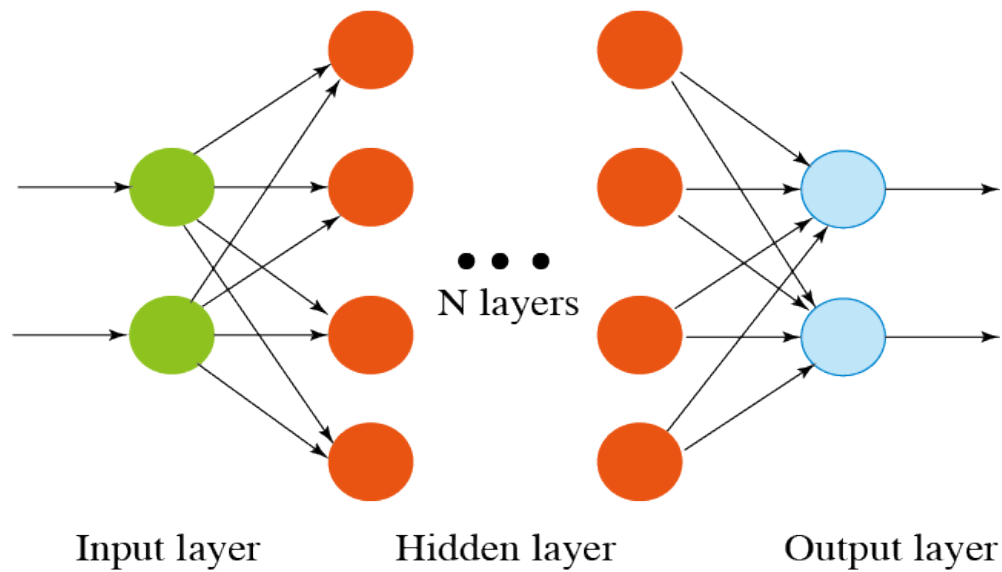

Figure 2.9: The basic structure of a FNN

Figure 2.9 shows the basic structure of a Feedforward Neural Network (FNN). It contains neurons arranged by layers, and the information can only transfer in one direction. Each node represents a neuron, and the edge represents the synapse that transmits the signal from the output of the prior neuron to the input of the next neuron associated with a weight. In general, neurons are aggregated into layers, and a network consists of an input layer, hidden layers, and an output layer. The input layer just brings the information outside into the neural network but doesn't perform any computation at this stage. The hidden layer has no direct concoction with the outside and locates between the input layer and the output layer. It can have more than one layer. The hidden layer conducts non-linear transformations of the input through the activation function, and transfer that information to the output layer. The output layer is also responsible for the function of computation and transfers the information from the neuron network to the outside.

Neural network with multiple hidden layers is called Deep Learning or Deep Neural Network (DNN), which is a subfield of machine learning. One of the main breakthroughs of DNN is the Convolutional Neural Network (CNN or ConvNet), which mimics the tuning properties of the visual cortex. Its name also indicates that this type of network has layers that employ convolution. The most beneficial aspect of CNN is that it can extract features from input data like images, and reduce the numbers of parameters without losing its characteristics. Compared with traditional ANN, CNN is an efficient algorithm for pattern recognition especially for image data with precise but simple architecture. The neuron of CNN is no longer fully connected to the previous layer, which reduces the numbers of parameters and speeds up the convergence. In addition, the parameters are further diminished by the interconnected weight sharing and the down-sampling dimensionality reduction. Excellent results are achieved for applications with CNN in the fields of image classification [140–142], object detection [143–145], human action recognition [146–148], object recognition [149, 150], Natural Language Processing (NLP) [151, 152], autonomous driving [64, 153], etc.

Besides the input layer and the output layer, the hidden layers of CNN include convolutional layers, flatten layers, pooling layers, fully connected layers. As found in the classic ANN, the input layer feeds the pixel values of the preprocessed (resizing, normalization, etc.) image into the neural network. Then, neurons extract local features from different parts of the input data from the previous layers with the same local receptive fields' size. The feature map is produced with the achieved local features and their positional relationships. The numbers of the parameters are reduced as the neurons from the same mapping plane share weights. Besides, it reduces the complexity of the neural network. This kind of spatial structure makes CNN more like a biological neural network and has unique benefits for image processing and speech recognition.

The activation function, loss function, and optimizers are introduced to determine which feature should be extracted. The activation function helps neurons to learn the real-world non-linear representations by conducting non-linear mapping between the input and output, which greatly enhanced the adaptivity of the neural network. The common activation functions include ELU [154], ReLU [155], Sigmoid [156], and tanh [157]. The loss function or cost function can be used as the learning criterion that shows the difference between the predicted (learned) value and the actual (expected) value. The regression problems and the classification problems are solved by minimizing the value of the loss function. Common loss functions include Mean Square Error (MSE) [158, 159], Mean Absolute Error (MAE) [160, 161], Cross Entropy (CE) loss [158, 162], contrastive loss [163, 164], center loss [165], triplet loss [166, 167], large-margin softmax loss [162], etc. Therefore, we need optimizers to minimize the loss function with acceptable training time. Common optimizers include Adaptive Moment Estimation (Adam) [168], Adamax [168, 169], Adagrad [170], Adadelta [171], Mini-Batch Gradient Descent (MBGD) [172], Momentum [173], Nesterov Accelerated Gradient (NAG) [174, 175], Root Mean Square prop (RMSprop) [172], etc. The pooling layer downsamples the input data with image local correction along the spatial dimensions (width, height), which further reduces for further layers the numbers of parameters and obviates redundancy by removing trivial features. The flatten layer converts the data from the previous layer into a 1D array and transfer it to the final classification model at the fully connected layer. The fully-connected layer connects like the traditional neural network, that each neuron connects to every neuron in the previous layer as well as the next layer.

Pothal *et al.* [176] proposed the Adaptive Neuro Fuzzy Inference System (AN-FIS) Controller for navigation by combing mutual network and fuzzy logic, which takes the benefits of both. [177] presented a goal-directed, collision-free navigation method by combining the potential field method and neural network. Gautam *et al.* [178] proposed a path planning algorithm that combines Genetic Algorithms (GA) and ANN to avoid collisions. Li *et al.* [179] utilized Q-learning to collect training data for the ANN path planning controller. And the proposed method outperformed the results by using only one of these two methods. Caltagirone *et al.* [54] developed a fully convolutional neural network model to generate the driving path from LIDAR point clouds. The proposed method utilizes a nonuniform sampling distribution generated from a CNN model based on learning quantities of successful planning cases from the A* algorithm.

Self-Organizing Map (SOM) is an unsupervised neural network proposed by [180]. SOM has been treated as a solution for specific optimization problems such as the Travelling Salesman Problem (TSP) [181–183]. The beauty of the SOM is the fact that the individual neurons adaptively tend to learn the properties of the underlying distribution of the space in which they operate [184].

SOM enables to observe unlabelled input data to identify hidden patterns and regularities and adjust its neurons to form topology-preserving desired patterns. The characteristic of unsupervised learning that learning without the reference from the supervisor can be an advantage, since the algorithm can search for patterns or regularities from different perspectives. Jin *et al.* [182] develop an integrated SOM with the genetic algorithm (GA) for TSP. Three learning mechanisms of different SOM-like neural networks are integrated to generate the proposed learning rule. A genetic algorithm is successfully specified to determine the elaborate coordination among the three learning mechanisms as well as the suitable parameter setting. Faigl *et al.* [185] presented a SOM-based strategy for Dubins traveling salesman problem (DTSP) to solve its multi-vehicle variant generalized for visiting target regions called k-DTSP with Neighborhoods (k-DTSPN). Zhu *et al.* [186] proposed an integrated biologically inspired SOM algorithm for task assignment and path planning of an autonomous underwater vehicle (AUV) system.

## 2.7.2 Learning-Based Approaches

Generally speaking, the key learning paradigms of machine learning include supervised learning, unsupervised learning, and reinforcement learning. For supervised learning, the learning process is based on pre-labeled data input. The term supervised originates from the fact that desired outputs are provided by an external teacher/supervisor [187]. The supervised learning algorithm analyses the training labeled data and infer a function to map new examples. On the contrary, unsupervised learning does not need a supervisor, it can extract an internal representation of the statistical structure implicit in the inputs without any labels [188]. For reinforcement learning, the agent receives feedback from the environment based on the agent's action and learns the optimal policy by trial and error. Immediate feedback (reward or punishment) will provide to the agent every time it acts.

Unlike imitation learning, reinforcement learning learns from the environment's feedback with no prior training data, which means it does not need a supervisor. As one of the most used reinforcement learning methods, Q-learning [189] can be applied to solve the path planning problem for mobile robots. Konar *et al.* [190] proposed an extension of the extended Q-learning (EQL) [191] for path planning of a mobile robot. The proposed method, called improved Q-learning (IQL), only needs to update the entries in Q-table once, which reduces the processing time significantly. The simulation results in different environments show the outperformance of the proposed method in comparison to the extended Q-learning (EQL), and classical

Q-learning (CQL) algorithms.

The rapid development of the neural network makes it possible to combine the learning-based methods with different types of neural networks in deep learning [192]. The integration of reinforcement learning and neural networks has a long history [193]. Tai and Liu [194] presented a robot exploration method based on the Deep Q-Network (DQN) framework, where a convolution neural network structure was adopted in the Q-value estimation of the Q-learning method. Baker and Gupta [195] introduced an autonomous method for CNN architecture selection through a novel Q-learning agent, and performed the effectiveness on given machine learning task without human intervention. The model-free method Deep deterministic policy gradient (DDPG) in [196] can learn competitive policies for all of our tasks using low-dimensional observations (e.g. cartesian coordinates or joint angles) using the same hyper-parameters and network structure.

# Chapter 3

# Path Planning in Unknown Static Environments

## 3.1 Motivation

This chapter is based on the publications [101] and [197]. It is fair to say that the autonomous mobile robot is increasingly important with the development of technology. Unmanned flying and underwater mobile robots have been widely used in the last few decades. Their advantage is unprecedented and irreplaceable especially in environments dangerous to humans, for example in radiation or pollution exposed areas. They have become the necessary tools for a wide range of activities including but not limited to real-time monitoring, surveillance, reconnaissance, border patrolling, search and rescue, civilian, scientific and military missions, etc. The capability to avoid obstacles without human intervention is crucial. As one of the most essential and important abilities of the autonomous mobile robot, avoidance of collisions has been and still is the focus of extensive research. The reactive collision-free navigation problems are challenging due to the limitation of the envi-ronment information. In order to reach the target in a complex environment without collisions, the robot should enable to detect and avoid obstacles.

In general, the existing path planning strategies can be classified into two main categories: global and reactive (local) approaches [61]. The path planning problems with the *a priori* known information of the environment are usually considered with global navigation strategies [198, 199]. Common approaches include tangent graph [4, 198], roadmap methods [200], probabilistic roadmap [201, 202]. One of the most classical approaches is called Model predictive control(MPC) [61]. It can predict the path for the next $N$ steps at each time interval. Its main advantage compared with potential fields [203, 204] inspired by [132] is more conservative when expanded to higher order models. It is can be applied in both known and unknown environment. It is an ideal way to seek for the optimal path, but computationally expensive. On the contrary, reactive navigation algorithms use on-board sensors to observe small fraction of the unknown environment at each time [198, 199, 205–208]. The ability to perceive and detect obstacles and avoid them during operation is expected. Compared with the global navigation methods, the reactive navigation algorithms are more practical and has been widely accepted due to its low requirements of *a priori* environment information. The artificial potential fields method like [209] introduced by [132] is one of the popular reactive collision-free navigation approaches. The main idea is to calculate the gradient of weighted sum of potentials, assuming that repulsive potentials exerted by obstacles while the goal point exerts an attractive potential. The local minima limit the applications of potential field, which causes the unwanted stop at unintended locations. Using polar histogram to choose velocity makes it possible to solve the stability problem. Dynamic window [210] solves the reactive navigation problem in the unknown environment with steady obstacles bu choosing safe directions among a set of valid trajectories. The collision-free problem with moving obstacles are much more painful. Collision cone [211] is applied to address this problem. However, most approaches based on it don't take non-holonomic model into consideration. It is likely to be computation expensive and requires determinative information of the obstacles.

Although the navigation algorithms show the ability to maneuver the robot in an unknown environment with good robustness, and enable to reach a steady target without any collisions, they cannot make sure the optimality of the trajectory. In other words, they cannot guarantee to find out the best possible trajectory/path (shortest in the length) in each operation, due to the limitation of environment information. On the other hand, in those complex environment that is difficult to model, the navigation algorithm with reinforcement learning methods [212] may be applicable to plan the best possible path for the robot, but need a huge number of trails for training [6, 213–216]. Another severe limitation is that many other researchers' papers [4–6, 216] ignore the non-holonomic constraints of the moving robots, which cause severe limitation in practice.

The shortcomings mentioned above motivated researchers to propose hybrid approaches by combing reactive navigation and reinforcement learning methods to accomplish better navigation results [217]. However, several issues need further investigation. In [218], a reinforcement learning method with the concept of sliding mode control is proposed, and the researchers proved their new control method has advantages of both the robustness and stability of the sliding mode control and applicability to complex systems that are hard to model. It is still a long way to apply this method to the real navigation environment. Another possible approach to solve the navigation problem is to modify the traditional reinforcement learning algorithm, such as [6], which developed a quantum-inspired reinforcement learning algorithm to improve the robustness. The applicability in practice and the training time cannot be settled, which is a huge problem in practice [204, 219].

The remainder of this chapter is organized as follows. Firstly, we will introduce a hybrid reactive algorithm for ground mobile robots, and then a similar flying robot's version extension. In the first section, we propose a novel hybrid reactive navigation strategy for non-holonomic mobile robots in cluttered environments. Our strategy combines both reactive navigation and Q-learning method. We intend to keep the good characteristics of reactive navigation algorithm and Q-learning and overcome the shortcomings of only relying on one of them. The second section presents a collision-free 3D path planning strategy for the non-holonomic mobile robot. The non-holonomic mobile robot travels through an unknown 3D environment with obstacles and reaches a given destination safely with no collisions. In addition, our approach enables to find the optimal (shortest) path to the target efficiently based on the avoiding plane selected.

## 3.2   A Hybrid Reactive Navigation Strategy for a Ground Mobile Robot

In this section, we propose a novel hybrid reactive collision-free navigation algorithm under an unknown environment cluttered with random shape and smooth (possibly non-convex) obstacles with constraints on the curvature of their boundaries. And the simulation results confirm the validity of our algorithm and show that the robot is able to seek the shortest path to reach the steady target with only several trails. We assume that the ground robot studied in this chapter is a unicycle, which can be considered as a Dubins car [3] with non-holonomic constraint. Compared with the omnidirectional mobile robot model [4–6], the non-holonomic model is more practical. This model has been widely used to describe the movements of different mechanical systems such as wheeled robots, unmanned aerial vehicles, missiles and references therein [24, 198, 206, 220–222].

The next subsection formulates the problem we are going to solve. The naviga-tion policies and Q-learning based path planning algorithm are presented in Subsection 3.2.2. The computer simulations are performed to validate the performance of the presented approach, and the results can be found in Subsection 3.2.3. In these simulations, two scenarios were considered with random-shaped obstacles. Subsection 3.2.4 includes brief conclusions.

## 3.2.1 Problem Statement

We consider a planar robot modelled as a unicycle, and we can represent its state as the configuration $X = (x, y, \theta) \in SE(2)$, where $(x, y) \in \mathbb{R}^2$ is the robot's Cartesian coordinates, and $\theta$ is the robot's heading. The robot travels with speed $v$ and angular velocity $u$, both constrained by given constants. The dynamics of the robot can be represented as follows:

$$
\begin{aligned}
\dot{x}(t) &= v(t) \cos \theta(t) \\
\dot{y}(t) &= v(t) \sin \theta(t) \\
\dot{\theta}(t) &= u(t)
\end{aligned}
\tag{3.1}
$$

where $v(t) = v$, and $v > 0$.

$$
u(t) \in [-u_M, u_M]
\tag{3.2}
$$

$u_M$ is the maximum angular velocity, the non-holonomic constraint can be de-

---

*SE(2): the special Euclidean group in the plane*

scribed as

$$|u(t)| \leqslant u_M \tag{3.3}$$

And the minimum turning radius of the robot satisfies

$$R_{min} = \frac{v}{u_M} \tag{3.4}$$

Any path $(x(t), y(t))$ of the robot as shown in (3.1) is a plane curve satisfying the following constraint on its so called average curvature (see [198]): let $P(s)$ be this path parametrised by arc length, then

$$\|P'(s1) - P'(s2)\| \leqslant \frac{1}{R_{min}}|s1 - s2| \tag{3.5}$$

Here $\|\cdot\|$ denotes the standard Euclidean vector norm. Note that we use the constraint (3.4) on average curvature because we cannot use the standard definition of curvature from differential geometry [223] since the curvature may not exist at some points of the path of the robot.

This unicycle model is moving in an unknown planar area, with several disjoint steady obstacles, $D_1, ..., D_k$, the safety margin of obstacle is given as $d_{safe} > 0$, and then the safety boundary of the obstacle $\partial D_i(d_{safe})$ can be found, see Figure 3.1(a). For the challenge of reactive navigation in unknown environment crowded by numbers of possibly non-convex obstacles with constraint on the curvature of their boundaries, the shortest path consists of edges of the so-called tangent graph is proven in [4].

**Assumption 3.2.1.** *Any two elements do not intersect.*

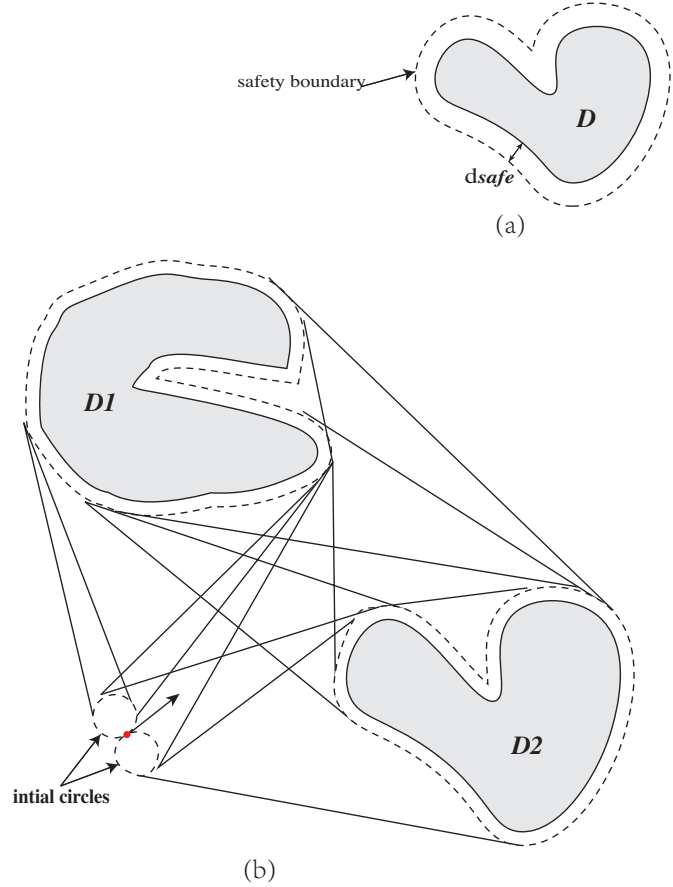

Figure 3.1: An illustrative example of (a) safety margin (b) tangent graph

The tangent line can be defined as a straight line that tangent to two elements simultaneously and do not intersect with each other. The points of elements belonging to tangent lines are called tangent points. And the curve between two tangent points on the same element is called arc. The tangent graph, denoted as $G(V, E)$, where vertex set $V$ consists of finite set of tangent points and edge set $E$ consists of finite set of arcs and tangent lines. Figure 3.1(b) shows an example of a tangent graph. Therefore, the trajectory of the robot at different position can be described easier.

**Definition 3.2.1.** *There are two circles with the radius $R_{min}$ that cross the initial robot position $(x(0), y(0))$ and tangent to the robot initial heading $\theta(0)$. We will call*

*them initial circles.*

**Assumption 3.2.2.** *For all $i$, the boundary $\partial D_i$ of the obstacle $D_i$ is a closed, non-self-intersecting analytic curve.*

**Assumption 3.2.3.** *For all $i$, the boundary $\partial D_i(d_{safe})$ is a closed, non-self-intersecting analytic curve with curvature $k_i(p)$ at any point $p$ satisfying $k_i(p) \leqslant \frac{1}{R_{min}}$.*

**Assumption 3.2.4.** *The minimum distance between obstacles is big enough for robot to travel safely.*

**Assumption 3.2.5.** *The robot's location $(x(t), y(t))$ and heading $\theta(t)$ are obtained by some localization technologies, like odometry, GPS, etc.*

## 3.2.2 Path Planning Algorithm

In this section, we consider the case when the robot does not know the location of the target or the obstacles *a priori*. The robot has several ultrasonic-type sensors around it, which help the robot to detect its current distance between the target and obstacles, and odometer to estimate its position. So that the robot is able to determine the relative coordinates of the targets and the boundaries of obstacles.

We will need the following assumptions and definitions to simplify the case.

**Assumption 3.2.6.** *Each tangent point only belongs to one tangent line.*

**Definition 3.2.2.** *There are two circles with the radius $R_{min}$ that cross the initial robot position $(x(0), y(0))$ and tangent to the robot initial heading $\theta(0)$. We will call them initial circles.*

### 3.2.2.1 Navigation Policies

According to [198], we approach this problem by reasoning over policies:

$$\xi = \{initial, pursuit, follow\} \tag{3.6}$$

Where *initial* refer to the policy that the initial condition of the robot is move along one of these two initial circles randomly. When the robot moves along the initial circle or obstacle boundary and reach an exit tangent point to the target, it begins to operate *pursuit* policy to move along the corresponding tangent line between its current tangent point and the target. Moreover, the robot will use the path planning algorithm in next section to decide between *pursuit* and *follow* policies, which means move to corresponding tangent line and follow current initial circle or obstacle boundary arc, respectively. Otherwise, the robot operates *follow* when it moves along a corresponding tangent line between initial circle and obstacle or between two obstacles, and reaches the tangent point on the obstacle boundary.

This navigation strategy was realised as a sliding mode control law by switching between the boundary following approach and the pure pursuit navigation approach from [198] as follows:

$$u(t) = \begin{cases} \pm u_M & R1 \\ \Gamma sgn[\phi \tan(t)] u_M & R2 \\ \Gamma sgn[\dot{d}_{min} + \chi(d_{min}(t) - d_{safe})] u_M & R3 \end{cases} \tag{3.7}$$

Where R1 refers to *initial* mode, R2 refers to *pursuit* mode, and R3 refers to *follow* mode. The variable $\Gamma = +1$ when robot's heading direction intersects with obstacles, and $\Gamma = -1$ otherwise.

The linear saturation function is defined as follows:

$$\chi(r) = \begin{cases} lr & if\ \|r\| \leqslant \delta \\ \delta l sgn(r) & otherwise \end{cases} \tag{3.8}$$

Where $sgn(r) := 1$, when $r > 0$, $sgn(r) := 0$, when $r = 0$, and $sgn(r) := -1$, whereas $r < 0$. $l, \delta > 0$ are tunable constants.

And the navigation mode transition rules can be illustrated as:

$$\begin{aligned} R1 \to\ R2 : &\quad reach\ the\ tangent\ point \\ &\quad\quad to\ target\ or\ obstacle \\ R2 \to\ R3 : &\quad d_{min}(t) < d_{trig}, \dot{d}_{min}(t) < 0 \\ R3 \to\ R2 : &\quad illustrate\ in\ next\ section \end{aligned} \tag{3.9}$$

According to (3.7), the navigation is fulfilled by switching between modes R1 - R3. At the beginning of each episode, R1 is active, and transits to other modes are determined by (3.9). Mode R1 describes the robot motion that move along the initial circle with maximal actuation. Mode R2 describes the robot motion that pursuit along the tangent segment, where $\phi \tan(t)$ is defined as the angle between the vehicle's heading and a line segment connecting the vehicle and currently tracked tangent edge. Mode R3 describes boundary following behaviour, where the control calculation is based on the minimum distance to the nearest obstacle, defined as $d_{min}(t)$. A constant $d_{trig} > d_{min}(t)$ is also introduced, which determines when the control system transitions to boundary following mode. In the following section, we will describe the procedure to transit from R3 to R2. We introduce Q-learning to find the best possible $p$ for the certain environment.

#### 3.2.2.2 Path Planning Algorithm

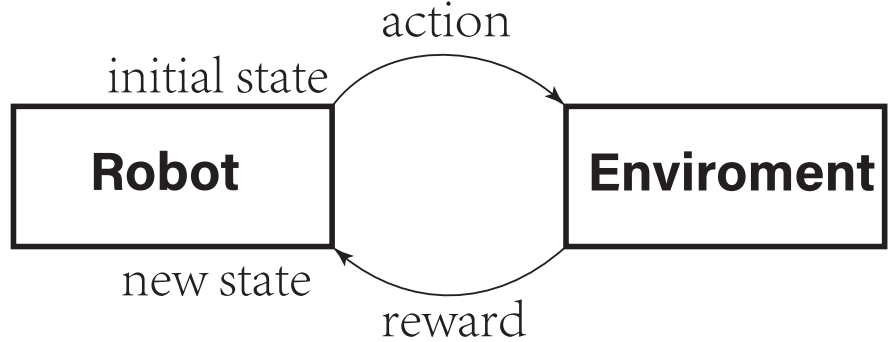

Figure 3.2: Schematic of Q-learning

In this section, we apply Q-learning [189] to find the best possible $p$ at each exit tangent point. The reinforcement learning is a real-time, online learning method [213]. By introducing Q-learning, the robot is able to sense more about the environment and learn to choose the optimal actions to achieve the goal, in our work, reach the target. The strategy that the robot learns from the environment and perform properly is by matrices and the numerical evaluation function, which assigns numerical values to different distinct actions at each distinct state. The policy of the robot's learning is to achieve as high reward scores as it can with long-term interest. The robot will be reward by "good" behaviour and punished by "bad" one. In our case, the robot can perceive the distinct state as a set of $\mathcal{S}$, $\forall s \in \mathcal{S}$, and its actions as a set of $\mathcal{A}$, $\forall a \in \mathcal{A}$ at each discrete time step $t$, respectively. At each discrete time step $t$, the robot senses the current state $s_t$, chooses a current action $a_t$ and performs it, the environment responds by returning a reward $R(s_t, a_t)$ and by producing the immediate successor state value $Q(s_t, a_t)$, who depend only on the current state and action. In other words, the robot regularly updates its achieved rewards based on the taken action at a given state. The schematic of Q-learning shows in Figure 3.2.

As it is impossible to predict in advance the exact outcome of applying an arbitrary action to an arbitrary state, we apply $Q(s_{t+1}, a_{t+1})$ (i.e. the Q-value) of the robot, which indicates the maximum discounted reward that can be achieved starting from $S$ and applying action $A$ first. With the action $a_t$ (moving along the boundary or the corresponding segment) at a given state $s_t$ (current exit tangent point's relative coordinate), the future total reward $Q(s_{t+1}, a_{t+1})$ of the robot is computed using

$$Q(s_{t+1}, a_{t+1}) = (1 - \alpha)Q(s_t, a_t) + \alpha[R(s_t, a_t) + \gamma Q_{max}(s_t, a_t)] \qquad (3.10)$$

where $R(s_t, a_t)$ is the immediate reward of performing an action $a_t$ at a given state , $Q_{max}(s_t, a_t)$ is the maximum Q-Valve and $\alpha$ is the learning rate, which is between 0 and 1, and $\gamma \in [0, 1)$, which is the discount factor. By following an arbitrary policy that produces the greatest possible cumulative reward over time, there exits the cumulative discounted reward $G_t$ achieved from an arbitrary initial state $s_t$ as follows:

$$\begin{aligned} G_t &= R_{t+1} + \gamma R_{t+2} + \gamma^2 R_{t+3} + \dots \\ &= \sum_{k=0}^{\infty} \gamma^k R_{t+k+1} \end{aligned} \qquad (3.11)$$

The discount factor $\gamma$ determines the relative value of delayed versus immediate rewards, $R_{t+k+1}$ are rewards of successor actions generated by repeatedly using the policy.

---

**Algorithm 1** Path Planning Algorithm

---

1: Set the $\gamma$ parameter and environment rewards matrix $R$

2: Initialize matrix $Q(s_t, a_t)$ to zero

3: **for** each episode **do**

4:      Select a random state $s_t$ from $\mathcal{S}$

5:      **while** the goal state hasn't been reached **do**

6:          Choose one possible action $a_t$ from current state $s_t$ by matrix $R$

7:          Consider moving to the next possible state $s_{t+1}$

8:          Based on all possible actions, computer the maximum $Q$ value for $s_{t+1}$ from current matrix $Q$

9:          Refresh matrix $Q$ by updating the value of $Q(s_t, a_t)$ derives from $Q(s_t, a_t) = R(s_t, a_t) + \gamma Q_{max}(s_t, a_t)$

10:         Set $S_{t+1}$ as the current state

11:      **end while**

12: **end for**

---

From the above algorithms, we can make the robot more concentrate on the immediate reward or the previous experience by increasing $\alpha$ or $\gamma$ and vice versa. In other word, larger $\alpha$ will make the robot more concern about the immediate reward, whereas larger $\gamma$ will lead to actions which pays more attention to previous experience, see Figure 3.3 for an example.

The training can be finished by numbers of training episodes. For each episode, it contains series of observations, actions, and rewards at each state from the initial observation to the terminal observation. The robot selects one among all possible actions for the current state, using this possible action to consider going to the next state. Furthermore, the robot tries to get maximum Q value for this next state based on all possible actions by (3.10), upload the new Q value to the Q-table (i.e. a large table with separate entry for each distinct state-action pair) and set the next state as the current state. The robot is going to iterate the above learning process until the goal state has been reached, which is the end of one episode. Provided that the system can be modelled as a deterministic Markov Decision Process (MDP), which means for a distinct state, the output is a certain action, the training episodes will be iterated until the robot's estimate Q table converges to the actual Q, and actions are chosen so that every state-action pair is visited infinitely often.

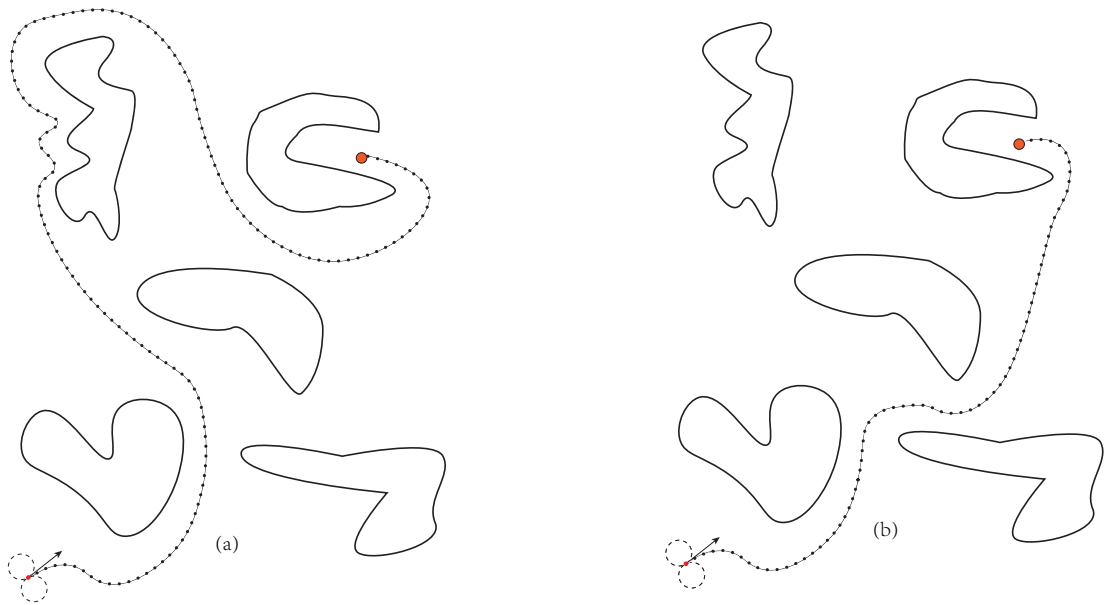

Figure 3.3: (a) One random trajectory generated without proposed algorithm and (b) the navigation trajectory after proposed algorithm's training

### 3.2.3 Simulation Results

To validate the effectiveness of the proposed algorithm, two simulation scenarios are carried out considering different conditions and environments. Firstly, the $25m \times 25m$ unknown environment filled with random-shaped obstacles is considered. The more crowded environment is handled in the second. The performance of proposed algorithm is evaluated by comparing the trajectories with and without the algorithm.

The control law was simulated using the perfect discrete time model, updated at a sample time of 0.1 s. The control parameters are shown in Table. 2. The robot has no *a priori* information about the environment, it can detect its surroundings only by 10 mounted ultrasonic sensors. The robot travels with constant speed $v$ and is controlled by the bounded angular velocity $u_M$ in both scenarios. According to the proposed navigation strategy, the robot switches between the boundary following approach and the pure pursuit mode with different angular velocities. The actual velocity of the robot is influenced by its corresponding angular velocity, which indicates the oscillations and reduction of magnitude. Hence, the maximum velocity of the robot never exceeds 0.5 m/s. Simulations were carried out on V-REP, interfaced with MATLAB and Python.

In order to figure out how these two algorithms works, we simulate the first scenario, and second scenarios as follows. Compare with the randomized navigation algorithm, for the robot in our proposed algorithm, the introducing of Q-learning can help robot to make the best decision at each exit tangent point.

Table 3.1: Simulation parameters

| Parameter | Value |
|---|---|
| Maximum angular velocity  $u_M$ | 60 degree/s |
| Maximum velocity  $v$ | 0.5 m/s |
| Safety margin  $d_{safe}$ | $> 0$m |
| Tunable constant  l | 0.08 |
| Tunable constant  $\delta$ | 0.5 |
| Transition margin  $d_{trig}$ | 0.8 m |
| Discount factor  $\gamma$ | 0.8 |
| $\varepsilon$ | 0.4 |
| range of probability  $p$ | [0,1) |

### 3.2.3.1 Unknown cluttered environment with random-shaped obstacles

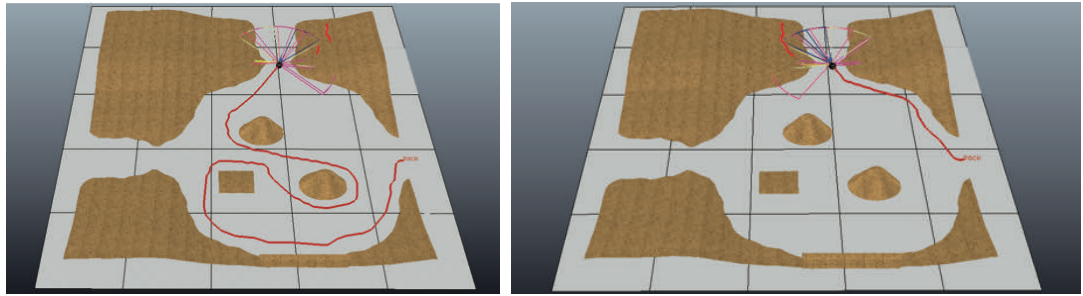

(a) The trajectory of the robot    (b) The trajectory of the robot

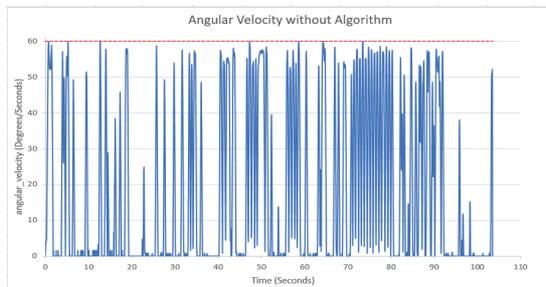 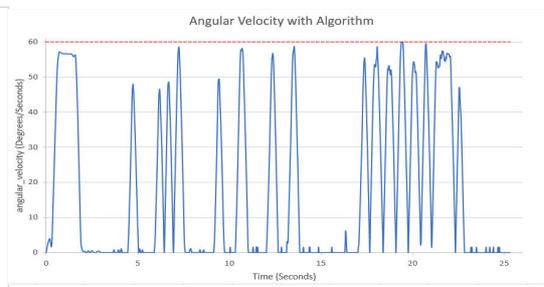

(c) Robot's absolute value of angular velocity (d) Robot's absolute value of angular velocity

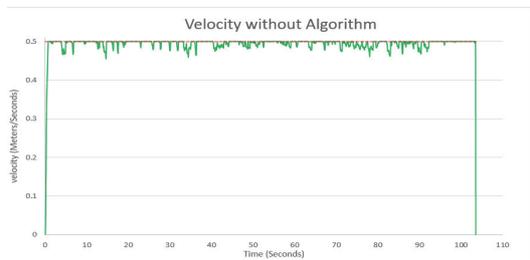 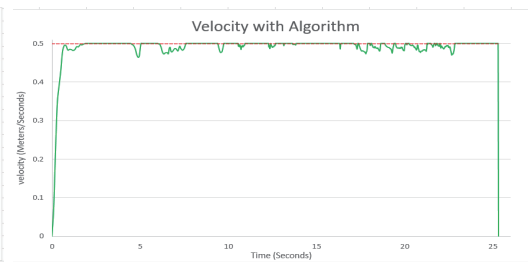

(e) Robot's velocity    (f) Robot's velocity

Figure 3.4: (a), (c), (e) Robot's trajectory, its absolute angular velocity, and velocity respectively, generated without our proposed algorithm, and (b), (d), (f) are robot's trajectory, its absolute angular velocity, and velocity respectively after proposed path planning algorithm's training in an unknown cluttered environment with random-shaped obstacles.

The first scenario aims to validate the proposed navigation algorithm in an unknown cluttered environment with random-shaped obstacles. The situation in Figure 3.4 is evaluated. It consists of random-shaped sand dunes as the obstacles. Figure 3.4 also shows how the robot plans its path, and its velocity and angular velocity changes with time before and after the proposed algorithm's training, respectively. The absolute value of the angular velocities and the speeds of the robot are within the constrains. And the average value probability $p$ after training is less than its give constrain for this particular cluttered environment. With our simulation results, we found out that the average value of the probability $p$ is susceptible to the unknown environment. As can be seen from Figure 3.4, the robot's total length of the trajectory for this environment is significantly deducted due to the application of our algorithm.

### 3.2.3.2 More challenging cluttered environment with random-shaped obstacles

The second scenario intends to discuss the performance of the algorithm with more cluttered deployed obstacles shown in Figure 3.5, which means the distance between dunes becomes closer. The simulation results could be attributed to that the proposed algorithm can always find out the shortest path of the operating area within a specific time, while the randomized algorithm may spend more time to the target, thus may use longer path. By comparing Figure 3.5 (a), (c), (e) with (b), (d), (f) respectively, we can find the advantage of proposed algorithm becomes more evident for the operating area with both first and second scenarios, which is more difficult for the random search to operate. Compared with traditional Q-learning algorithm, our proposed algorithm simplify the complex calculation and react swiftly.

We show the benefits of introducing Q-learning in terms of navigation performance by efficiently finding out the shortest path in an unknown complex steady environment. The merit of our proposed algorithm can be reflected even by simply comparing the length of the path, and the total time of process with the randomized navigation algorithm.

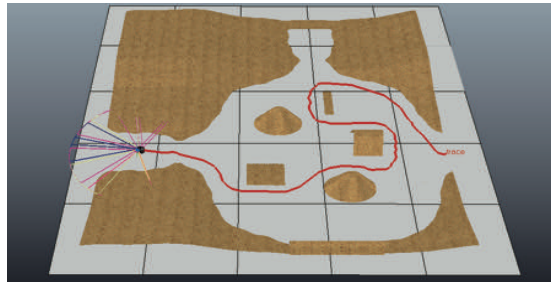
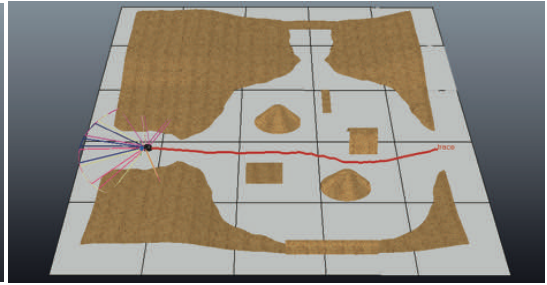

(a) The trajectory of the robot      (b) The trajectory of the robot

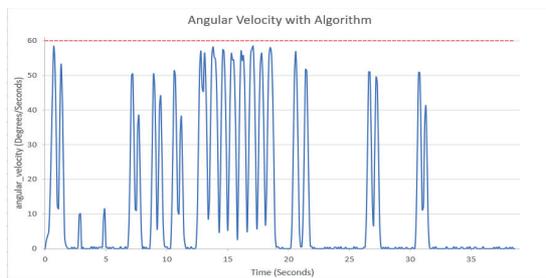
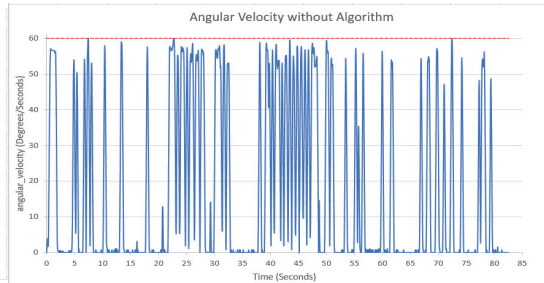

(c) Robot's absolute value of angular velocity (d) Robot's absolute value of angular velocity

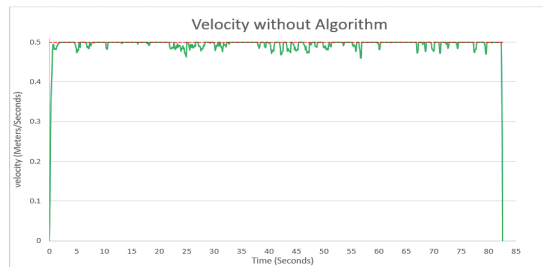
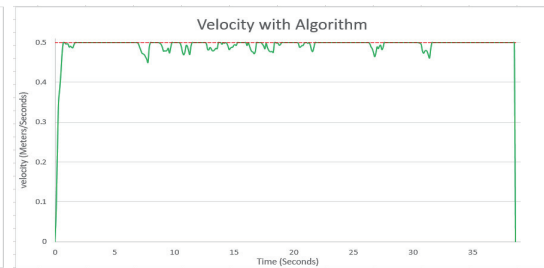

(e) Robot's velocity                 (f) Robot's velocity

Figure 3.5: (a), (c), (e) Robot's trajectory, its absolute angular velocity, and velocity respectively, generated without our proposed algorithm, and (b), (d), (f) are robot's trajectory, its absolute angular velocity, and velocity respectively after proposed path planning algorithm's training in a more challenging cluttered environment with random-shaped obstacles.

### 3.2.4   Section Summary

In this section, we considered the collision-free reactive navigation problem in the environment cluttered with obstacles constrained on the curvature of their bound-aries. We proposed a hybrid reactive navigation algorithm that can be operated in cluttered environments and the results successfully validated that the robot can find the shortest path to the steady target without any collisions. A series of navigation policies are adopted as the fundamental of our collision-free navigation. What's more, a proposed path planning algorithm taught the robot how to switching between these policies by the selections of probability $p$ at each exit tangent point, in order to make sure the robot can find out the shortest path during its reactive navigation. The performance and effectiveness of our algorithm has been confirmed by extensive computer simulations in different scenarios.

It should be pointed out that, as the usage of the autonomous ground vehicles are more often, the study of decentralized control of multiple vehicles has gained much attention in recent years. This type of systems with several vehicles can be viewed as an example of networked control systems, and have been studied extensively [28, 224, 225].

## 3.3 A Collision-Free Path Planning Strategy for Flying Robots

A considerable amount of collision-free navigation approaches have been presented see [198, 199, 226–229] and references therein. However, substantially all existing research in this area focuses on the planar mobile robot in a 2D environment. Only few concentrates on the much challenging case of 3D collision-free navigation see [230–234] and references therein.

The proposed strategy can be seen as the extension of the planar strategy of last section to the much challenging 3D navigation problem. In this section, we proposed a sense-and-avoid type navigation strategy, which can drive a single non-holonomic mobile robot to a given destination in an unknown three-dimensional environment without hitting any obstacles. The performance of the proposed 3D robot navigation algorithm is verified by several illustrative simulation examples in Virtual Robotics Experimentation Platform (V-REP).

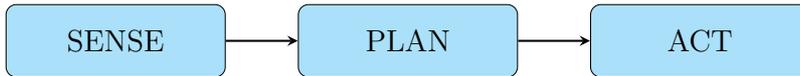

Figure 3.6: Sense-and-avoid strategy

The remainder of this section is organised as follows. Subsection 3.3.2 describes the problem we are going to solve. The navigation strategy is presented in Subsection 3.3.2. The computer simulations are performed to validate the performance of the presented approach, and the results can be found in section 3.3.3. Section 3.3.4 includes brief conclusions.

### 3.3.1   Problem Statement

In this work, we consider a three-dimensional non-holonomic mobile robot, whose Cartesian coordinates can be described as $c_R(t) := [x(t), y(t), z(t)]$. The 3D mathematical model is as follows [230]. Some assumptions and definitions had been claimed in Section 3.2 so they are used directly without explaining.

$$\dot{c_R}(t) = V(t)\widetilde{v}(t) \tag{3.12}$$

The robot travels with speed $V(t) \in \mathbb{R}$, and its orientation $\widetilde{v}(t) \in \mathbb{R}^3$, and $||\widetilde{v}(t)|| = 1$ for all $t$.

$$\dot{\widetilde{v}}(t) = u(t), u(t) \in \mathbb{R}^3 \tag{3.13}$$

The robot travels with constants:

$$||u(t)|| \leq u_{max}, V(t) \in [V_{min}, V_{max}], \widetilde{v}(t) \cdot u(t) = 0 \tag{3.14}$$

for all $t$. And the minimum turning radius of the robot satisfies

$$R_{min} = \frac{V}{u_{max}} \tag{3.15}$$

Here $||\cdot||$ denotes the standard Euclidean vector norm. And we can easily see that $\widetilde{v}(t)$ and $u(t)$ are always orthogonal. Moreover, the above kinematics model has been widely used to describe many unmanned underwater and flying vehicles [234] and references therein.

We study the optimal navigation problem for a mobile robot $R$ moving in a 3D environment with $1, 2, ..., k$ obstacles $D_k$, $k = \{1, 2, ..., k\}$. We assume that there is a moving *covering sphere*, $O_k$, $k = \{1, 2, ..., k\}$ with radius $R_o$, the safety margin of obstacle is given as $d_{safe} > 0$, and then the safety boundary of the obstacle's covering sphere is $\partial O_k(d_{safe})$. And the obstacle is always inside the covering sphere, see Figure 3.7(b). The objective of the proposed algorithm is to drive the robot to the stationary target $T$ without hitting any obstacles. What's more, the direction $H(t)$ from robot's coordinates to the target is known to the robot, see Figure 3.7(a). The distance $d(t)$ between the robot and the covering sphere is defined as

$$d(t) := \min_{P \in O(t)} ||P - c_r(t)|| \tag{3.16}$$

The value of $\dot{d}(t)$ is available to the robot as well. In addition, the robot is able to find out the optimal path to the target in the environment cluttered with stationary obstacles if required.

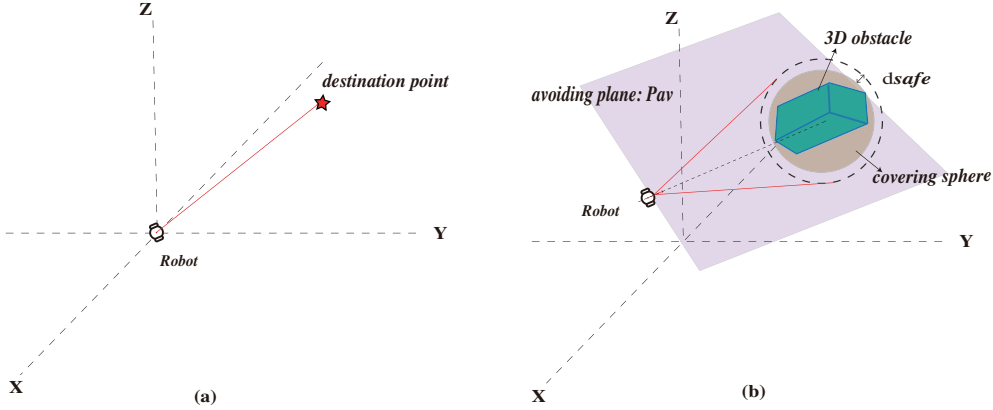

Figure 3.7: (a) direction $H(t)$ between robot and destination; (b) 3D obstacle, covering sphere, avoiding plane $P_{av}$

## 3.3.2 Path Planning Algorithm

In this section, we present the proposed 3D navigation algorithm by extending from the concepts from [198, 230]. The robot updates its data at discrete time $0$, $\delta$, $2\delta$, $3\delta$..., where the sampling period $\delta > 0$.

We assume that there is $P_{av}(t)$, called *avoiding plane*, in where the strategy of avoidance take place. This plane is constructed by vector $\tilde{v}(t)$ and $h_k(t)$, and going through the robot's coordinates $c_R(t)$, where $h_k(t)$ is the direction from the robot's coordinates $c_R(t)$ to the center of the covering sphere and $\tilde{v}(t)$ is robot's orientation, see Figure 3.7(b).

We will need the following assumptions and definitions to simplify the case.

**Definition 3.3.1.** *The navigation strategy is said to be destination reaching with obstacle avoidance if there exits a time $t_f > 0$ such that $c_R(t_f) = T$ and $d_i(t) \geq d_{safe} > 0 \forall i = 1, 2, ..., k, \forall i \in [0, t_f]$. [230]*

**Assumption 3.3.1.** *The robot's location $c_R(t)$ and heading $\tilde{v}(t)$ are obtained by some localization technologies, like odometry, GPS, etc.*

**Assumption 3.3.2.** *For all the boundary $\partial O_k$ of the obstacle's covering sphere $O_k$ is a closed, non-self-intersecting analytic curve.*

**Assumption 3.3.3.** *For all $i$ the boundary $\partial D_i(d_{safe})$ is a closed, non-self-intersecting analytic curve with curvature $k_i(p)$ at any point $p$ satisfying $k_i(p) \leqslant \frac{1}{R_{min}}$.*

**Assumption 3.3.4.** *The minimum distance between obstacles is big enough for robot to travel safely.*

**Assumption 3.3.5.** *For each tangent point on its avoiding sphere $P_{av}(t)$ at time $t$ only belongs to one tangent line.*

We approach this problem by reasoning over policies:

$$\xi = \{R1, R2\} \tag{3.17}$$

Where R1 refers to *pursuit* mode, and R2 refers to *avoid* mode.

**R1:** Pure pursuit strategy towards destination $T$ with $V_{max}$

**R2:** The avoidance manoeuvre is given in (3.18) and operates on the pre-determined $P_{av}$

$$V(t) \in [V_{min}, V_{max}]$$
$$u(t) = \Gamma sgn[\dot{d}(t) + \chi(d(t) - d_{safe})]u_{max}\vec{i}(t) \tag{3.18}$$

The variable $\Gamma = +1$ when $\tilde{v}(t)$ intersect with obstacles, and $\Gamma = -1$ otherwise.

$$\vec{n}_{p_{av}}(t) = h_k(t) \times \tilde{v}(t)$$
$$\vec{i}(t) = \tilde{v}(t) \times \vec{n}_{p_{av}}(t) \tag{3.19}$$

where a unit vector $\vec{i}(t) \in p_{av}$ is vertical to $\tilde{v}(t)$. It make sure $\Gamma\vec{i}(t)$ is always pointing away from the obstacle. The saturation function $\chi(r)$ is defined as follows:

$$\chi(r) = \begin{cases} lr & if \, \|r\| \leqslant k \\ lk \, sgn(r) & otherwise \end{cases} \tag{3.20}$$

Notice that $sgn(r) := 1$, when $r > 0$, $sgn(r) := 0$, when $r = 0$, and $sgn(r) := -1$, where $l, k > 0$. A constant $d_{trig}(t) > d_{safe}(t) > 0$ is also introduced. The robot's initial mode is R1, and will keep moving under R1 until some certain condition. This navigation strategy was realised as a sliding mode control law by switching between the above two modes as follows:

$$\begin{aligned} R1 \rightarrow \ R2: \qquad & d(t) = C, \dot{d}(t) < 0, \\ & C \ is \ a \ constant \\ R2 \rightarrow \ R1: \qquad & d(t) < d_{trig}, \\ & \tilde{v}(t) \ is \ towards \ H(t) \end{aligned} \tag{3.21}$$

When solving the question in a 3D dynamic environment, we need to add following assumption: $B(t) = [x_D(t), y_D(t)] \in D(t), \left\|\dot{B}(t)\right\| \leq v$, which implies that $D(t)$ cannot move or change faster than $v$.

### 3.3.3 Simulation Results

The mean purpose of this paper is navigating the robot to the target in a 3D environment with obstacles without hitting any of them. In this section, computer simulations are carried out to validate the performance of the presented strategy. The simulation in Figure 3.8 considers the collision-free navigation in an 3D environment cluttered with stationary obstacles. And we can see that the presented algorithm is able to navigate the robot in 3D environment with obstacles, go through and find the target.

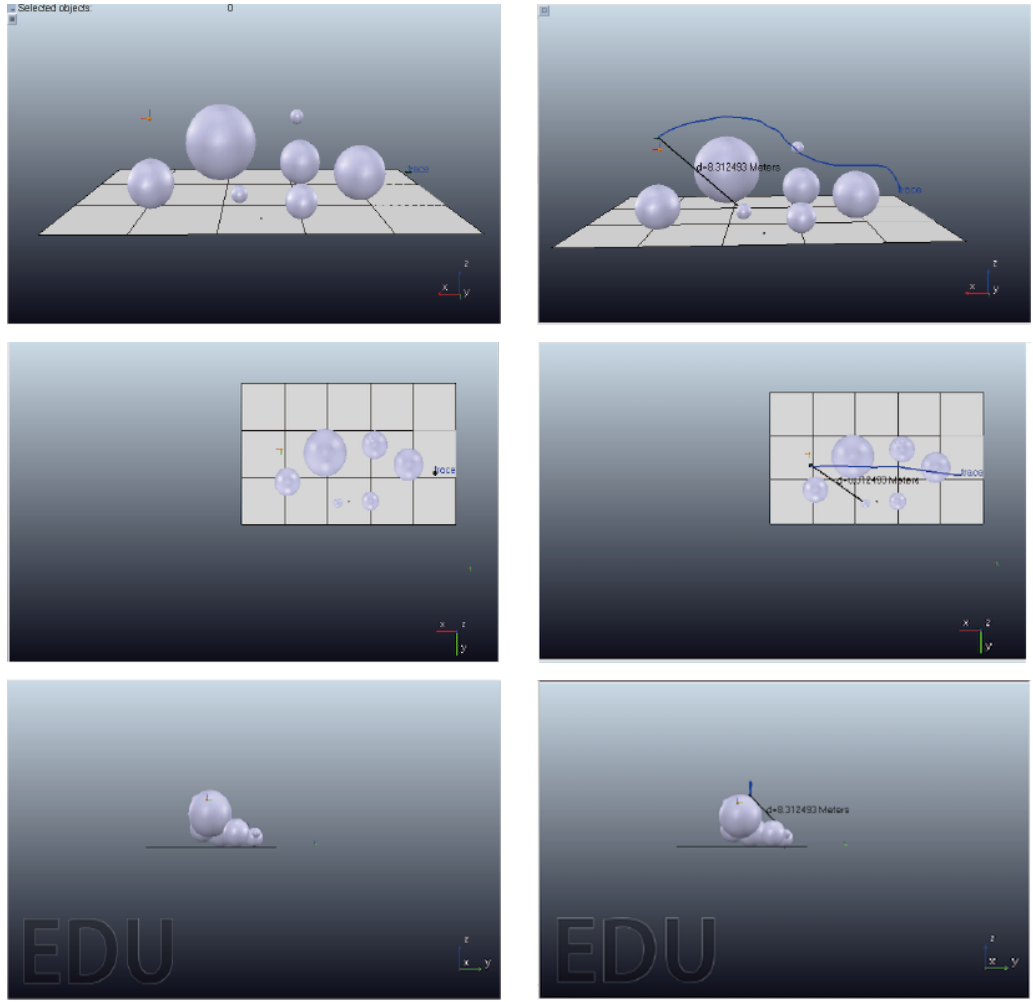

Figure 3.8: (a), (c), (e) are simulation environments with cluttered stationary obstacles in different angles, and (b), (d), (f) are corresponding simulation results with cluttered stationary obstacles

## 3.3.4 Section Summary

Collision avoidance is an essential important ability for the autonomous mobile robot, and it has been and still is one of the most important research topics for researchers. The main contribution of this paper is to develop a 3D safe navigation strategy for the non-holonomic mobile robot. The proposed algorithm enables the robot to traverse the cluttered environment with obstacles, and reach the given steady target. And the performance of the proposed algorithm has been proven by computer simulations. The presented algorithm can navigate the robot via the optimal path successfully while keeping a safe distance from obstacles in a 3D environment. The application extension of the strategy to deal with moving and even deforming obstacles is currently under investigation. Moreover, the relationship between the selection of the avoiding plane $P_{av}$ and the navigation performance are considered in future work.

# Chapter 4

# Path Planning in Unknown Dynamic Environments

## Contents

## 4.1 Motivation

This chapter is based on the publication [102]. Chapter 3 presented a hybrid reactive navigation strategy in cluttered environments. However, it is not suitable for an environment containing moving obstacles. The path planning algorithm in the last chapter is improved for a more challenging environment.

In order to overcome the mentioned problems, the integrated representation of the environment and reinforcement learning approaches have been employed in the path planning of mobile robot control. There are many control solutions like tangent graph based path planning [198, 235], the curvature velocity [236], potential filed [209] inspired by [132], and dynamic window [210]. However, most of them only consider the case of stationary obstacles. Methods like [198, 236, 237] have failed to be fulfilled in a partially known or totally unknown environment. Collision-free control algorithms like [235] based on the tangents of the obstacles enable to generate collision-free path between the robot and the target with obstacles around. However, one of the huge drawbacks is that they have not taken the robot kinematic factors into consideration so that it is hard to apply to real robots. It is too simple to be executed or extended in dynamic environments. The potential fields methods like [209] introduced by [132] are also a useful tool in collision-free path planning. The main idea is to calculate the gradient of the weighted sum of potentials, assuming that repulsive potentials exerted by obstacles while the goal point exerts an attractive potential. The local minima limit the applications of potential field, which causes the unwanted stop at unintended locations. Many other researchers' papers [5, 6, 216] rarely take the non-holonomic constraints on a robot's motion control into consideration, which causes severe limitations in practice. The collision-free robot control problem with moving obstacles are much more challenging. Collision cones [211] and velocity obstacles [238] are applied to address this problem. However, these methods are likely to be computation expensive and require determinative information on the obstacles. Methods such as [239] has been recently used to overcome the above difficulties in optimum path selection by the recursive method. However, it is not easy to apply this method for both polygons and curved obstacles. [240] developed an optimal path planning algorithm among certain shaped steady obstacle. [241] deployed a sensor network consists of some range finder to detect static and moving obstacles, which combined mobile robots and sensor networks together.

Recently, many researchers proposed path planning algorithms using reinforcement learning. [237] applied reinforcement learning to control quadrotors, while the path has been generated using random shooting without taking the dynamic of the vehicle into consideration. At the same time, path planning algorithms with reinforcement learning [212] may be applicable to safely maneuver the robot in an unknown dynamic environment and even plan the best possible path for the robot. We cannot ignore the need for huge numbers of trails for training [6,215,232]. Other intelligent control methods such as Fuzzy logic [242], neural network [243] and heuristic approaches are applied to solve the environmental uncertainty in path planning. However, the performances of fuzzy logic and neural network highly depend on the selection of the membership function, and the construction of connecting and optimal selection of nodes, respectively.

The goal of this work is to combine the positive characteristics of several previous approaches with new ideas to generate a new control algorithm that provides an effective solution to the robot path planning problem in terms of minimum time, maximum safety. As the algorithm navigates the non-holonomic model, through the environment, it receives updated information by sensor measurements concerning both the environment itself and the dynamic elements within. With the algorithm presented, no approximation of the shapes of the obstacles or even any information about the obstacles' velocities is needed. Extensive computer simulations have been applied to confirm the performance of our path planning strategies. For comparison and validation purposes, the presented method has been subjected to the path planning algorithm [199] based on an integrated representation of the information about the environment.

## 4.2 Problem Statement

The modeled mobile robot is considered as a Dubins car as Chapter 3. The robot travels with speed $v > 0$, angular velocity $u$ constrained by $u \in [-u_M, u_M]$, $u_M$ is the given maximum angular velocity, then the minimum turning radius of the robot is $R_{min}$. The robot updates its $u(t)$ at discrete time 0, $\Delta T$, $2\Delta T$, $3\Delta T$..., where the
sampling period $\Delta T > 0$.

We can represent its state as the configuration $X = (x, y, \theta) \in SE(2)$, where
$c_r(t) := [x(t), y(t)] \in \mathbb{R}^2$ is the robot's Cartesian coordinates, and $\theta$ is the robot's
orientation measured from the x-axis in counterclockwise direction, $\theta \in (-\pi, \pi]$.

Some assumptions and definitions had been claimed in Chapter 3 so they are
used directly without explaining. We assume that the robot is moving in a dynamic
planar environment $D(t) \in \mathbb{R}^2$, which is unknown to the robot. In other words,
we can treat $D(t)$ as a time-varying subset contains deformable steady and moving
obstacles. The safety margin of the obstacle is given as $d_{safe} > 0$, and then the
safety boundary of the obstacle $\partial D_i(t)$ can be found, see Figure 4.1. And $\partial D(t) =
\partial D_1(t) + \partial D_2(t)$.

**Definition 4.2.1.** *All obstacles $D_i(t)$ for $i = 1, 2, ...., n$ together form the time-
varying subset $D(t)$.*

**Assumption 4.2.1.** *$\partial D(t)$ is closed with piecewise analytic boundarys.*

The distance $d(t)$ between the robot and the environment is defined as

$$d(t) := \min_{P \in D(t)} \|P - c_r(t)\| \qquad (4.1)$$

Here $\|\cdot\|$ denotes the standard Euclidean vector norm.

The angle from the robot to the steady target $\tau$ is known to the robot as $\theta_\tau(t)$
for all t. The goal of the proposed strategy is navigating the robot to the target $\tau$
in the unknown dynamic environment $D(t)$, and $d(t) \geq d_{safe}$ all the time.

**Assumption 4.2.2.** *For any point $B(t) = [x_D(t), y_D(t)] \in D(t), \left\|\dot{B}(t)\right\| \leq v$, which
implies that $D(t)$ cannot move or change faster than $v$.*

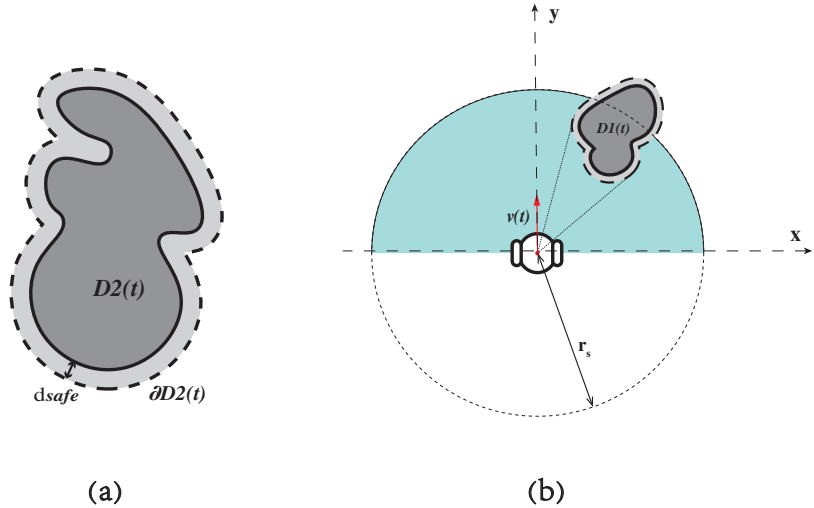

Figure 4.1: (a) An illustrative example of the mobile robot model, safety margin $d_{safe}$, obstacle $D_1(t)$ (dark grey), and the safety boundary of the obstacles (lighter grey) $\partial D_1(t)$, (b) The semicircle shaped sensing disc of radius $r_s$ in light blue

**Assumption 4.2.3.** *The semicircle shaped sensing disc $S(t)$ of radius $r_s$ is made of ultrasonic rays, whose are centered at the center of robot's position at time $t$. See Figure 4.1(b) for a demonstration. The obstacles inside the disc will be detected.*

## 4.3 Path Planning Algorithm

In this section, we consider the case when the robot does not know the environment *a priori*. The basic idea of this work is developed from [101]. The robot has ultrasonic-type sensing disc around it, which help the robot to detect the current distance and angle from the robot to the target and obstacles, respectively. The odometer is applied. The robot is able to determine the relative coordinates of the targets and the boundaries of obstacles.

According to [199], we address this problem by the binary function $M(\alpha, t)$:

$$M(\alpha, t) := \begin{cases} 1 & S_b(t) \leq r_s \cos(\alpha) \\ 0 & \text{otherwise} \end{cases} \tag{4.2}$$

Where $S_b(t)$ is the distance from the emitting position of the ray to a point B of the safety boundary of the obstacle $\partial D(t)$ at time $t$ and the angle between the robot's heading and point B is $\alpha$. The example can be fund in Figure 4.2. When $M(\alpha, t) = 1$, at least one obstacle is detected by the sensing disc at $t$.

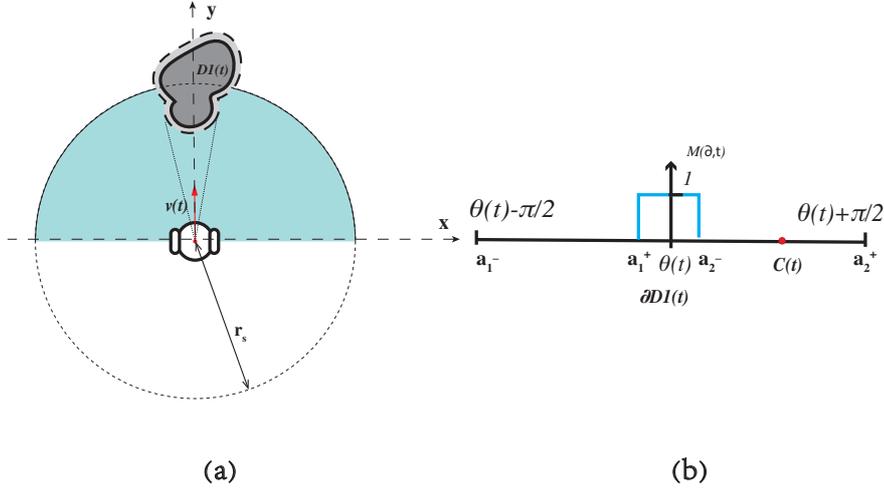

(a)                                          (b)

Figure 4.2: Illustrative example of $M(\alpha, t)$ function, and direction $C(t)$

The function $m(t)$ defined as

$$m(t) := \begin{cases} 0 & M(\alpha, t) = 0, \forall \alpha \in (\theta(t) - \frac{\pi}{2}, \theta(t) + \frac{\pi}{2}) \\ 1 & \text{otherwise} \end{cases} \tag{4.3}$$

For some $\alpha \in (\theta(t) - \frac{\pi}{2}, \theta(t) + \frac{\pi}{2})$, if $m(t) = 1$ and $M(\alpha, t) = 0$, there exists a finite numbers of non-overlapping non-empty obstacle-free interval $[\alpha_i^-, \alpha_i^+], \alpha_i^- < \alpha_i^+$. If $M(\theta(t), t) = 0$, $\theta(t) \in (\alpha_i^-, \alpha_i^+)$, otherwise $j(t)$ is the index that is closed to $\theta(t)$ is

calculated by:

$$j(t) := argmin\{\mid \alpha_i^- \mid, \mid \alpha_i^+ \mid\} \tag{4.4}$$

over all $i$. And we define the middle of the interval closest to $\theta$ is

$$C(t) := \frac{\alpha_{j(t)}^- + \alpha_{j(t)}^+}{2} \tag{4.5}$$

In Figure 4.2, there is one obstacle close to the robot at time $t$. In the case $M(\alpha, t) = 0$, there consists two non-overlapping non-empty obstacle-free interval $(\alpha_i^-, \alpha_i^+)$, where $\alpha_1^- = -\frac{\pi}{2}$, $\alpha_2^+ = -\frac{\pi}{2}$. In the case $M(\alpha, t) = 1$, the direction $C(t)$ is determined by the interval $(\alpha_2^-, \alpha_2^+)$ closest to $\theta(t)$ with probability $p$, therefore $C(t) := \frac{\alpha_2^- + \alpha_2^+}{2}$; or by the next-closest interval $(\alpha_1^-, \alpha_1^+)$ with probability $(1 - p)$, therefore $C(t) := \frac{\alpha_1^- + \alpha_1^+}{2}$.

And the path planning rules can be illustrated as:

$$u(t) := \begin{cases} u_M sgn(\theta_\tau(t) - \theta(t)) & m(K\Delta T) = 0 \\ u_M sgn(C(t) - \theta(t)) & m(K\Delta T) = 1 \end{cases} \tag{4.6}$$

Where $sgn(r) := 1$, when $r > 0$, $sgn(r) := 0$, when $r = 0$, and $sgn(r) := -1$, whereas $r < 0$, $\theta_\tau(t)$ and $C(t)$ are angle to the steady target, and the middle value of the interval, respectively.

According to Eq. (4.4) (4.5) (4.6), the path planning is fulfilled by adjusting the value of the angular velocity $u(t)$. The path planning algorithm can be illustrated as following: if the sensing disc senses the obstacle(s), the vehicle turns to the middle value of the closest obstacle-free zone with probability $p$, to the middle of the next-closest interval with probability $(1 - p)$. Otherwise, the vehicle heads to the target.

The distance to the nearest obstacle, defined as $d(t)$. The strategy mentioned enable to find the target with $d_{safe} \geq d(t)$ for any $t \geq 0$. We will describe the procedure to select the optimal collision-free interval with probability $p$.

In this paper, we apply one of the most efficient reinforcement learning method, called Q-learning, to find the best possible $p$ for each interval. The robot is able to "learn" the optimal $p$ by interacting with the environment. The robot learns from the environment and performs properly by the numerical evaluation function, which assigns numerical values to different distinct actions at each distinct state.

The policy of the robot's learning is to achieve as high reward scores as it can with long-term interest. At each discrete time step $\Delta T$, the robot senses the current state $s_t$, chooses a current action $a_t$ and performs it, the environment responds by returning a reward $R(s_t, a_t)$ and by producing the immediate successor state value $Q(s_t, a_t)$, who depend only on the current state and action. In other words, the robot regularly updates its achieved rewards based on the taken action at a given state. In our case, the robot can perceive the distinct state as a set of $S$, $\forall s \in S$, and its actions as a set of $A$, $\forall a \in A$ at each discrete time step $\Delta T$, respectively. With the action $a_t$ (moving along the boundary or the corresponding segment) at a given state $s_t$ (current exit tangent point's relative coordinate), the future total reward $Q(s_{t+1}, a_{t+1})$ of the robot is

$$Q(s_{t+1}, a_{t+1}) = (1-\alpha)Q(s_t, a_t) + \alpha[R(s_t, a_t) \\ + \gamma Q_{max}(s_t, a_t)] \tag{4.7}$$

where $R(s_t, a_t)$ is the immediate reward of performing an action $a_t$ at a given state , $Q_{max}(s_t, a_t)$ is the maximum Q-Valve and $\alpha$ is the learning rate, which is between 0 and 1, and $\gamma \in [0, 1)$, which is the discount factor. By following an arbitrary policy that produces the greatest possible cumulative reward over time, there exists the cumulative discounted reward $G_t$ achieved from an arbitrary initial state $s_t$ as follows:

$$\begin{aligned} G_t =& R_{t+1} + \gamma R_{t+2} + \gamma^2 R_{t+3} + \ldots \\ =& \sum_{k=0}^{\infty} \gamma^k R_{t+k+1} \end{aligned} \qquad (4.8)$$

The discount factor $\gamma$ determines the relative value of delayed versus immediate rewards, $R_{t+k+1}$ are rewards of successor actions generated by repeatedly using the policy.

Increasing $\alpha$ will make the robot more concern about the immediate reward, whereas $\gamma$ will lead to actions which pay more attention to previous experience. We can make the robot concentrate on immediate reward or the previous experience more by tuning $\alpha$ or $\gamma$.

The training contains multiple training episodes, and each episode is formed with numbers of observations, actions, and rewards at each state from the initial observation to the terminal observation. The proposed algorithm is shown in Algorithm 2. The robot selects one among all possible actions for the current state, using this possible action to consider going to the next state. The purpose of the robot is to get maximum Q value for this next state based on all possible actions by Eq. (4.7), upload the new Q value to the Q-table (i.e. a large table with separate entry for each distinct state-action pair) and set the next state as the current state. The robot is going to iterate the above learning process until the goal state has been reached, which is the end of one episode. The system can be modeled as a deterministic Markov Decision Process (MDP).

---

**Algorithm 2** Reactive Path Planning Algorithm

---

1: Set the $\gamma$ parameter and environment rewards matrix $R$

2: Initialize matrix $Q(s_t, a_t)$ to zero

3: **for** each episode **do**

4:     **while** the goal state hasn't been reached **do**

5:         **if** $m = 1$ and $M(\alpha, t) = 0$ when $\alpha \in [\theta(t) - \frac{\pi}{2}, \theta(t) + \frac{\pi}{2}]$ **then**

6:             Select action $a_t$ as the closest collision-free interval from current state $s_t$ by matrix $R$ with possibility $p$

7:             or Select action $a_t$ the next-closest interval from current state $s_t$ by matrix $R$ with possibility $(1 - p)$

8:         **else**

9:             Select action $s_t$ as the target direction from from current state $s_t$ by matrix $R$

10:         **end if**

11:         Consider moving to the next possible state $s_{t+1}$

12:         Based on all possible actions, computer the maximum $Q$ value for $s_{t+1}$ from current matrix $Q$

13:         Refresh matrix $Q$ by updating the value of $Q(s_t, a_t)$ derives from $Q(s_t, a_t) = R(s_t, a_t) + \gamma Q_{max}(s_t, a_t)$

14:         Set $S_{t+1}$ as the current state

15:     **end while**

16: **end for**

---

## 4.4 Simulation Results

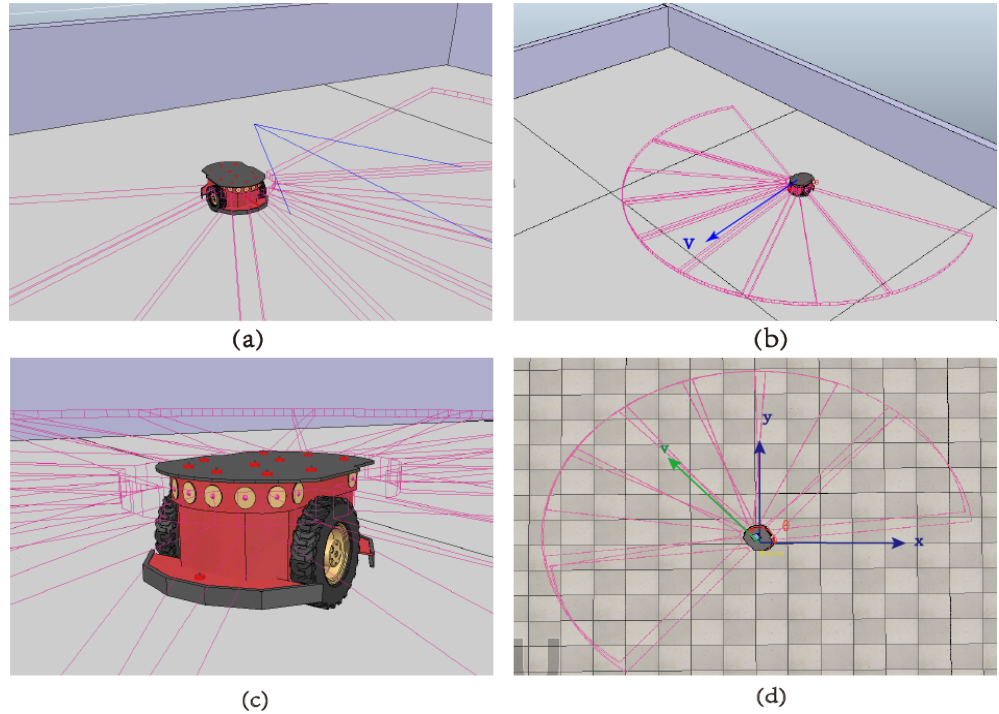

Figure 4.3: (a) the mobile robot model; (b) sensing disc of the model;(c) sensors of the model;(d) planform of the model

In this section, two simulation scenarios are carried out. Firstly, the unknown environment with cluttered stationary obstacles is considered. The more challenging one with both stationary and dynamic obstacles is handled in the second scenarios. In order to show the effectiveness of the proposed method, a comparison with path planning algorithm [199], which shows the superior performance over another popular approach called velocity obstacle approach (VOA), has been performed under the same obstacle configuration. The robot has no *a priori* information about the $15m * 25m$ planar environment, it uses ultrasonic-ray sensing disc to detect its surroundings, which is emitted by mounted sensors, shown in Figure 4.3. The robot's sensors' range is shown as the pink sector around the robot see Figure 4.3(b), and Figure 4.3(c) shows the yellow round parts are the sensors in the front of the robot.

The perfect discrete time model is applied in the simulations, updated at a
sampling period of 0.1s. The simulation parameters can be found in Table 4.1, and
the results of simulations in Table 4.2. Simulations were carried out on V-REP,
interfaced with MATLAB.

## 4.4.1   Reactive Path Planning in a Steady Environment

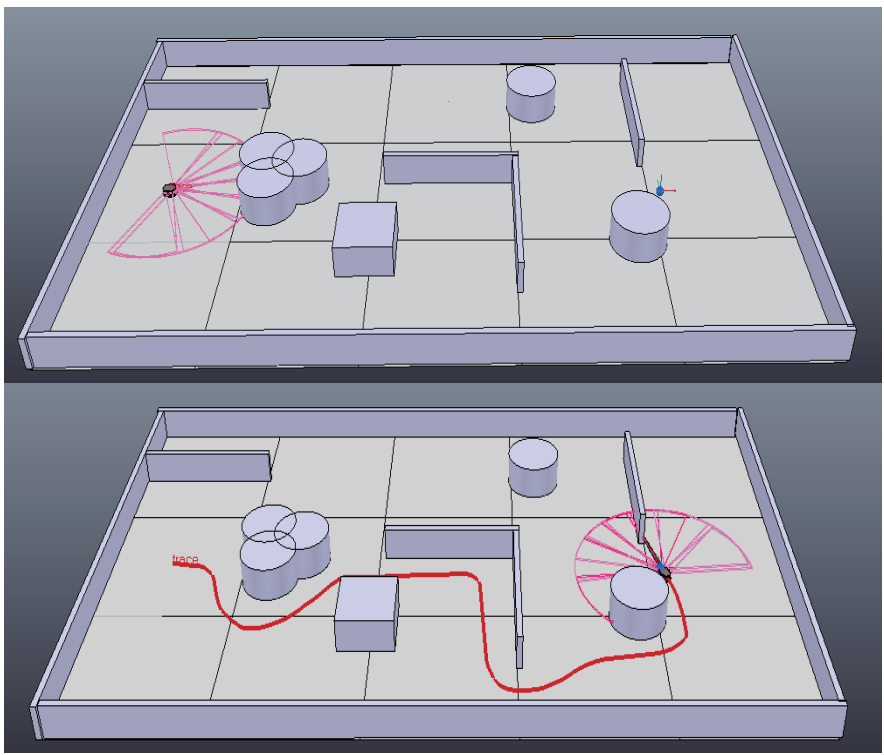

Figure 4.4: Robot path planning result when the integrated environment represen-
tation is applied

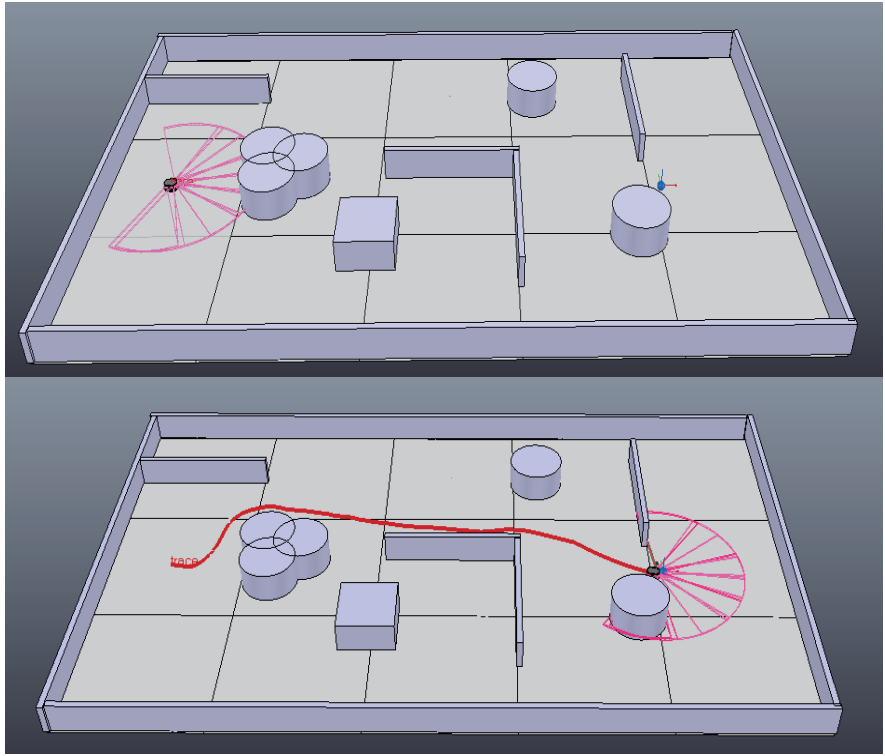

Figure 4.5: Robot path planning result when the the proposed algorithm is applied

Firstly, the situation in the first scenario shown in Figure 4.4 and Figure 4.5 is considered. Arbitrary obstacles are randomly deployed in a $15 \times 25$ m$^2$ planar field. The simulation results of the proposed algorithm are shown in Figure 4.5, compared with the results of the algorithm in Figure 4.4. The robot moves towards the direction of the steady target, and only change directions to the closest collision-free interval to the current heading when probability $p$. Otherwise, the robot will turn to the next-closest interval.

In Figure 4.5, the robot learns from its training results for achieving maximum Q value for this next state based on all possible actions by selecting the best interval facing different obstacles. A improvement of 28.1% was obtained over the integrated environment representation method for the average time from Table 4.2. By comparison, it is evident that our presented method's performance is superior to the compared algorithm, i.e., integrated environment representation method.

## 4.4.2 Unknown Cluttered Environment with Moving Obstacles

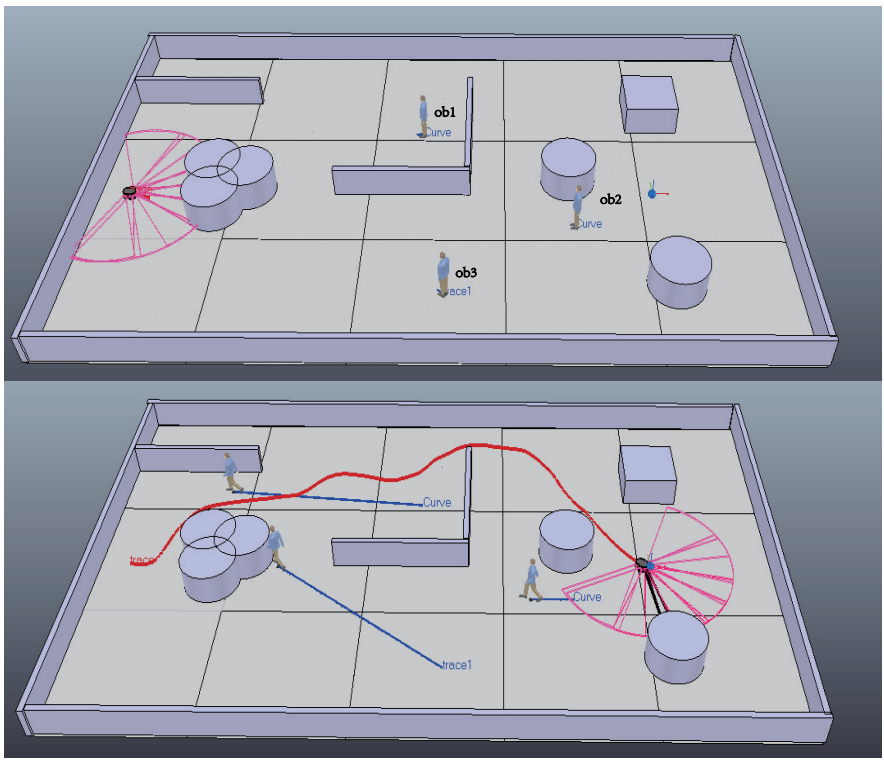

Figure 4.6: Robot path planning in a complex dynamic environment with the integrated environment representation method

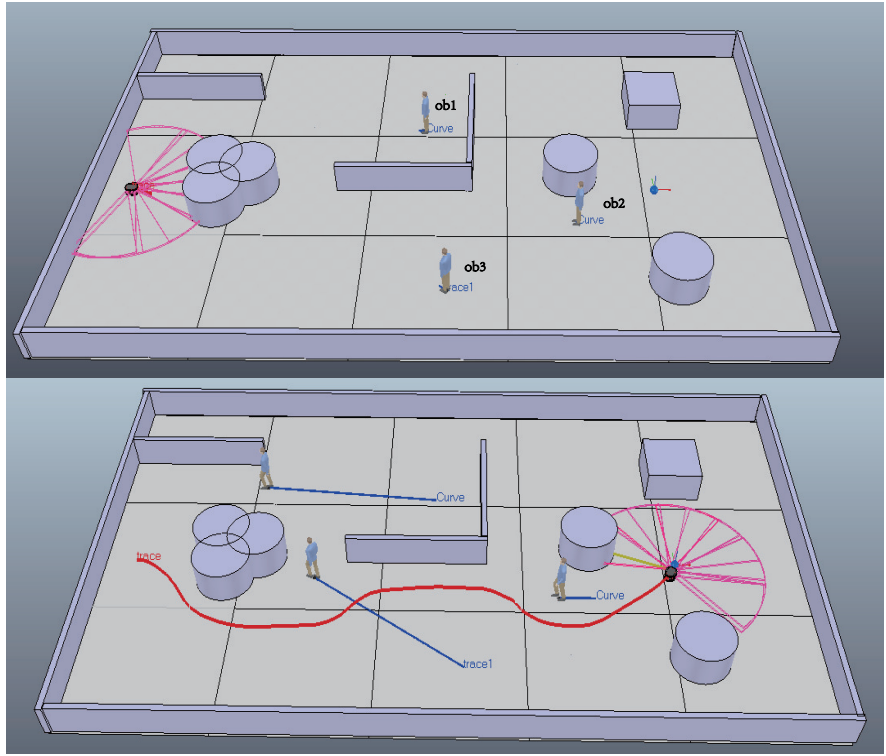

Figure 4.7: Robot path planning in a complex dynamic environment with the proposed algorithm

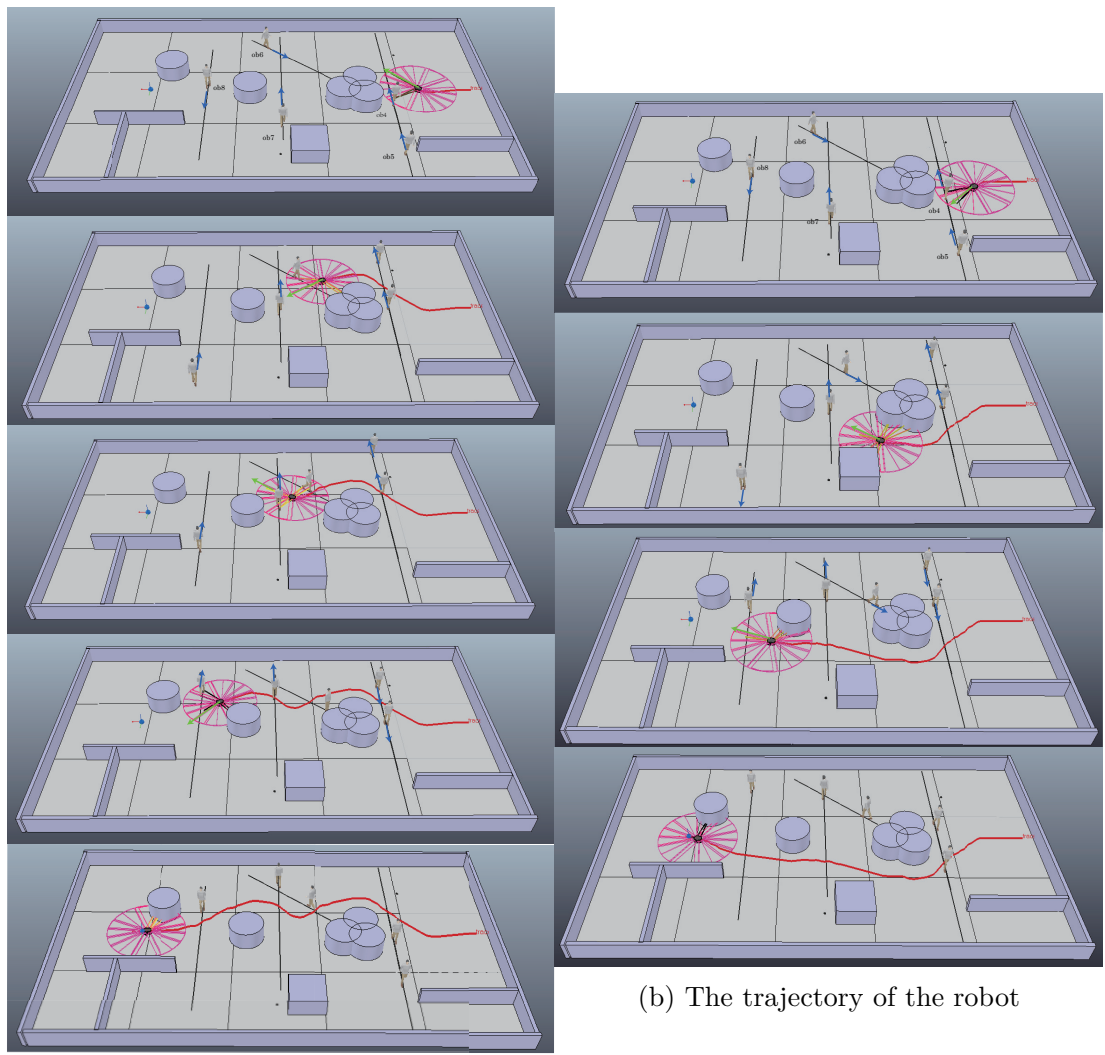

(a) The trajectory of the robot

(b) The trajectory of the robot

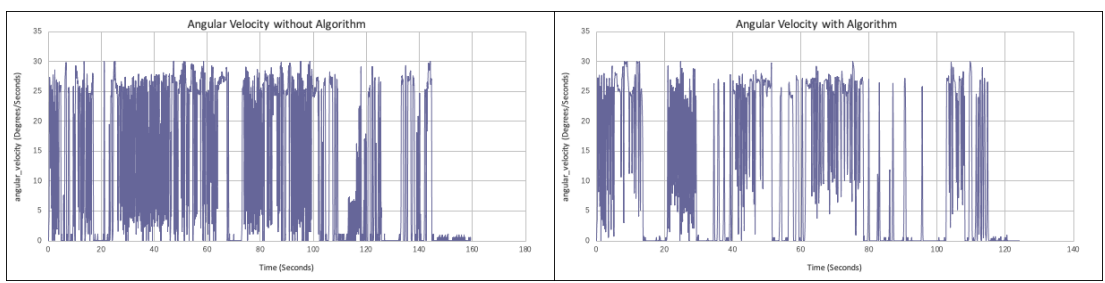

(c) Robot's absolute value of angular velocity (d) Robot's absolute value of angular velocity

Figure 4.8: (a),(c)Robot's trajectory, and its absolute angular velocity with the integrated environment representation method;(b),(d)robot's trajectory, and its absolute angular velocity with the proposed algorithm

Table 4.1: Simulation Parameters

| Parameter | Value |
|---|---|
| Maximum angular velocity $u_M$ | 30 degree/s |
| Velocity $v$ | 0.25 m/s |
| Safety margin $d_{safe}$ | 1.0m |
| Sensing range $r_s$ | 2.5m |
| ob1's velocity | 0.1 m/s |
| ob2's velocity | 0.05 m/s |
| ob3's velocity | 0.1 m/s |
| ob4's velocity | 0.1 m/s |
| ob5's velocity | 0.1 m/s |
| ob6's velocity | 0.05 m/s |
| ob7's velocity | 0.15 m/s |
| ob8's velocity | 0.12 m/s |

To validate the effectiveness of the proposed algorithm, two simulations in different complex environments with both steady and dynamic obstacles are conducted shown in Figure 4.6, Figure 4.7 and Figure 4.8. The robot is travelling in both environments with different numbers of steady and dynamic obstacles (ob1-3 and ob4-8), and the control parameters are shown in Table 4.1. All the moving obstacles go and walk back and forth along their own path. Figure 4.7 demonstrate excellent performance over Figure 4.6 by the length of the path, an average of 16% was obtained over the compare algorithm, i.e., integrated environment representation method. In Figure 4.7, ob4 and ob5 have same path with the velocities, which can be seen as a chain of moving obstacles. ob7 and ob8 cause a "trapped" situation for the robot, and the robot goes through the middle of the gap if and only if the

Table 4.2: Simulation Results

| Environment | Average time (s) | | Improvement (%) |
|---|---|---|---|
| | Compared algorithm | Proposed algorithm | |
| Static | 114.4 | 82.2 | 28.1 |
| Dynamic1 | 102.5 | 86.1 | 16.0 |
| Dynamic2 | 159.9 | 124.4 | 22.2 |

space is sufficient.  The proposed method achieves an improvement of 22.2% over
the integrated environment representation method for the average time.

## 4.5   Summary

In this chapter, we develop a new path planning method which utilizes integrated
environment representation and reinforcement learning to control a mobile robot
with non-holonomic constraints in unknown dynamic environments. With the con-
trol algorithm presented, no approximating the shapes of the obstacles or even any
information about the obstacles' velocities is needed. Our novel approach enables
to find the optimal path to the target efficiently and avoid collisions in a cluttered
environment with steady and moving obstacles. We carry out extensive computer
simulations to show the outstanding performance of our approach.

It is also worth mentioning that another rapidly emerging research area is the
cooperative control of multi-vehicle networked systems using inter-vehicle communi-
cation. With limited information about each other and the environment, the vehicle
has to make decisions independently, see, e.g., [244–249].

# Chapter 5

# UAV Path Planning for Reconnaissance and Surveillance

## 5.1 Motivation

This chapter is based on the publications [250] and [35]. In this chapter, we extend the path planning problem into the coverage problem in the reconnaissance and surveillance of Unmanned Aerial Vehicles (UAVs). UAVs, also known as aerial drones, are becoming increasingly present in our everyday lives [11]. As their extensive use recently jumped from military to hobby and professional applications [31], the complete coverage becomes a necessary function for activities including but not limited to border patrolling [37, 41], search and rescue [33], 3D reconstruction [32, 99, 251], infraction inspection [36], and surveillance and security [35, 50, 197, 252], etc.

In general, the coverage problem was first put forward over the 2D grid environment by [39]. We can classify this problem into two main categories based on vehicle's movements. The static coverage focuses on the deployment of the hovering UAVs to reconnoiter over certain terrains [40, 41, 50], while the dynamic coverage addresses the reconnaissance and surveillance problem by moving UAVs [35, 47, 197].

In the common reconnaissance and surveillance scenario, the flying vehicle equipped with a downward-facing video camera with a certain visibility angle can monitor the targets of interest on the ground, like vehicles, humans, animals, etc. [37, 45, 47]. We can evaluate the quality of the surveillance in terms of coverage and resolution [48]. In this case, the lower altitude of the traveling path is preferred for a better resolution of the observed region of the terrain. One of the most significant technical challenges is to completely cover a given target of ground with the minimum number of drone's waypoints, which requires every point on the target area can be seen at least once by the onboard camera during one complete surveillance circle. However, these two evaluation terms need to reach a suitable compromise to perform an ideal surveillance duty.

This chapter presents two novel path planning algorithms to address the aforementioned gaps. In contrast to the existing literature, our approach takes both UAV kinematics constraints and camera sensing constraints into consideration. In the first algorithm, we consider the fixed-wing vehicle case, assuming the vehicle flies at a given altitude with constant speed and limited turning radius. This Dubins aircraft model is similar to those in Chapter 3. We present a two-phase strategy to solve this surveillance problem. Firstly, an easily implementable estimation algorithm is developed at a given altitude, and the minimum number and locations of waypoints are determined to provide the complete coverage of the target area. The second phase addresses the distribution of the achieved locations over one or more UAVs and creates the shortest paths to reconnoiter the corresponding area of interest. The Dubins paths consist of straight lines and arcs of the circle of a constant radius. To achieve this, regular triangular patterns and the clustered spiral-alternating method are implemented, respectively. Our second algorithm concerns the surveillance problem over geometrically complex environments with varying altitudes and occlusions, such as mountainous terrains and urban regions. In the first stage, the challenge is to find a set of camera locations called the *vantage waypoint set* to provide full coverage of the area of interest, which can be viewed as a 3D Art Gallery Problem using drones as the observers. In the second stage, one or several UAVs are determined to conduct the full coverage reconnaissance and surveillance duty along individual routes respecting their kinematic constraints in the optimization criterion (the shortest time possible). This variant of the combinatorial traveling salesman problem is solved by introducing unsupervised learning and Bézier curves.

The remainder of the chapter is divided into two sections for surveillance problems at given and different altitudes, respectively. Each section starts with the addressed problem with the necessary background, followed by details of the proposed surveillance solution. The performance of the algorithm is then evaluated using computer simulations, and finally the summary.

## 5.2 Surveillance Algorithm at given altitude

This section provides a two-stage approach similar to [35, 48, 49] but more realistic and efficient. In the first stage, the waypoints to be visited are determined by constructing a triangulation of the ground terrain consisting of congruent equilateral triangles, and every point on the target area can be seen from at least one equilateral triangle. In the second stage, the clustered spiral-alternating path planning strategy calculates a short and smooth trajectory along which the vantage waypoint set needs to be visited. The trajectory is parameterized as a sequence of straight segments and Dubins curves, and every point on the target area can be covered at least once in the complete surveillance circle. This section takes advantage of the methods for the definition and assumption in [41,50]. Some assumptions and definitions are used directly without explaining. This section is derived from [250].

## 5.2.1 Problem Statement

We consider the UAV surveillance problem over one or multiple disjoint surface areas. We control one or a fleet of fixed-wing UAVs to travel with constant speed $v$ and angular velocity $u$ at a given altitude, which is referred to as Dubins vehicle mentioned in previous chapters.

$$\dot{x}(t) = v(t)\cos\theta(t)$$
$$\dot{y}(t) = v(t)\sin\theta(t) \qquad\qquad (5.1)$$
$$\dot{\theta}(t) = u(t)$$

$\theta$ is the vehicle's heading.

Let $(x_g, y_g)$ denote the Cartesian coordinates of a point on the ground. Let $z_g = F(x_g, y_g)$ be the perpendicular elevation of the point $(x_g, y_g)$ on the terrain. Moreover, let $\mathcal{G}$ denote a given Lebesgue measurable region [253] on the ground with zero elevation, i.e., $z_g = 0$. In addition, this region $\mathcal{G}$ has piecewise boundary. Our goal is to survey the region $\mathcal{G}$ by one or more UAVs. Let $Z_{max}$ and $Z_{min}$ be the given constants representing the maximum and minimum altitudes of UAVs. Any vantage point $(x, y, z)$ must make the following constraints hold:

$$(x, y) \in \mathcal{G}, z \in [Z_{min}, Z_{max}], 0 < Z_{min} < Z_{max}. \qquad\qquad (5.2)$$

Each UAV carries a fixed EO/IR camera, which is downward-facing to the ground with a certain visibility angle $0 < \alpha < \pi$. The sensing process is based on pinhole perspective projection [254], and the quality of the resolution is influenced by both the distance to the object and the physical parameters. In particular, if we set the camera's center as UAV's position $(x, y, z)$, the UAV can only see points along the $z$ axis with a limited angle of view. So that a cone-shaped field of view (FOV) with radius

$$R := z \cdot \tan(\frac{\alpha}{2}) \qquad (5.3)$$

is constructed (see Figure 5.1). A point is invisible if it falls out of the FOV.

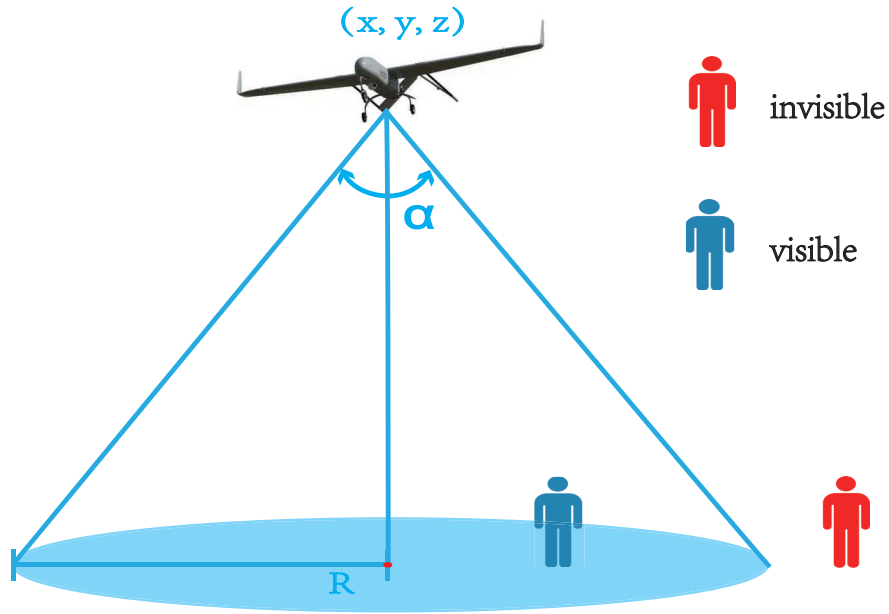

Figure 5.1: The visibility cone

As aforementioned, a surveillance mission should deliver a coverage of the area of interest with a satisfactory resolution. Moreover, *complete coverage* means that every point of the region $\mathcal{G}$ is seen at least once in a complete surveillance circle of UAVs. One of the most significant technical challenges is how to completely cover the target region by the minimum number of waypoints. The visibility cone enlarges when the UAV flies higher, but the resolution of the onboard recording is worse. We can choose a certain altitude that can achieve the required resolution. Here, we select the lowest possible altitude $z$, i.e., $Z_{min} < z < Z_{max}$ for the best resolution. At this altitude, the generated waypoint set, which provides complete coverage of the target area with a resolution that cannot be further better, is preferred to conduct an ideal surveillance mission.

**Definition 5.2.1.** *The complete coverage means constrain (5.2), (5.4) is satisfied and every point on the target area $\mathcal{G}$ can be seen at least once in a complete surveillance circle.*

In the herein addressed surveillance problem, we consider one waypoint location $p_i \in \mathbb{R}^3$ from vantage waypoint set $P = \{p_1, ..., p_n\}$ where the drone should visit, and the record can be taken within $\delta$ distance from $p_i$, i.e., the photo/video of a certain part of the target area can also be taken within $\delta$ distance from the particular location $p_i \in \mathbb{R}^3$. Thus the vantage waypoint set should contain the waypoint location $p_i^*$ that $||(p_i^*, p_i)|| \leq \delta$.

Moreover, the distance $q_{ij}$ between waypoint $i$ and $j$, and the minimum distance $q_i$ between waypoint $i$ and the terrain should hold the following constraints to avoid collisions:

$$q_{ij} \geq c_1, \quad q_i \geq c_2 \tag{5.4}$$

where $c_1 > 2\delta > 0$ and $c_2 > \delta > 0$ are given safety margins.

Under this condition, the problem becomes to find the trajectory to visit $\delta$-neighborhood of all assigned vantage points, which involves the optimization of the sequence of visits, i.e., an instance of Dubins-vehicle Travelling Salesman Problem (DTSP). The final trajectory is a sequence of straight line segments and circular arcs with minimum radius that traverse the $\delta$-neighborhood of all assigned vantage points.

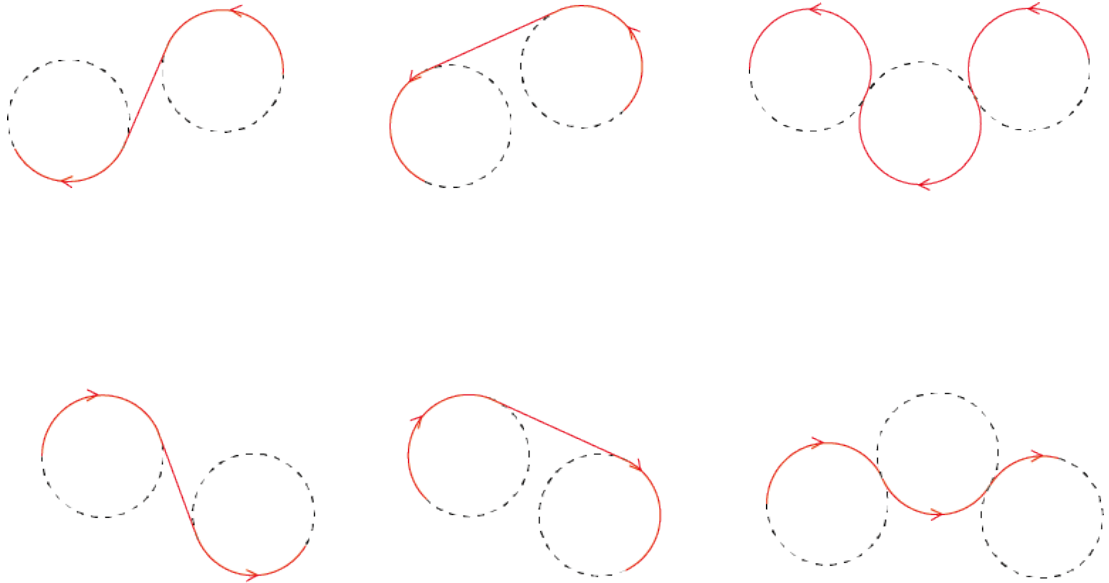

Figure 5.2: Six cases of Dubins curves

In the DTSP, the shortest path between two configurations must be one of the six possible combinations: LSL, LSR, LRL, RSL, RSR, and RLR, where 'S' means going straight and 'L' and 'R' denote left and right turn with a minimum turning radius, respectively. The situation is illustrated in Figure 5.2.

## 5.2.2 Surveillance Algorithm

In this section, we propose the following strategy based on a decomposition of the surveillance problem into a complete coverage problem and an instance of Dubins-vehicle Travelling Salesman Problem (DTSP). We try to find the waypoints to fully cover the area of interest first, and then plan trajectory along with these locations at the same altitudes, such that the length of the path to visit all the locations is minimal. It should be noted that our design is not necessarily unique or optimal, but a preliminary approach is sufficient to prove our point.

### 5.2.2.1 Stage One: Waypoints Generation

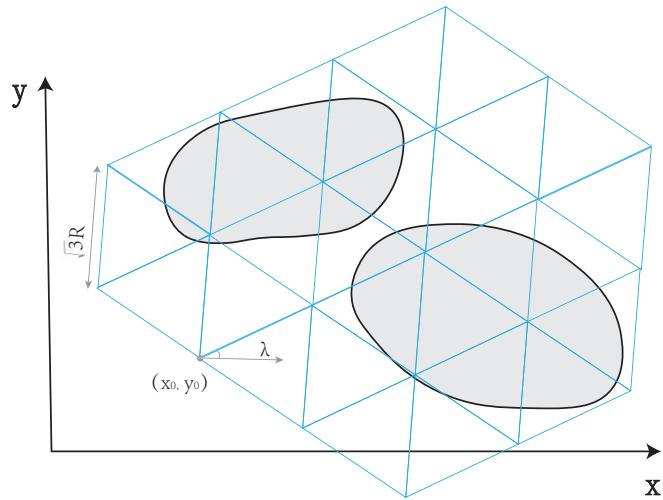

Figure 5.3: Triangulation $\Delta(\lambda, x_0, y_0)$.

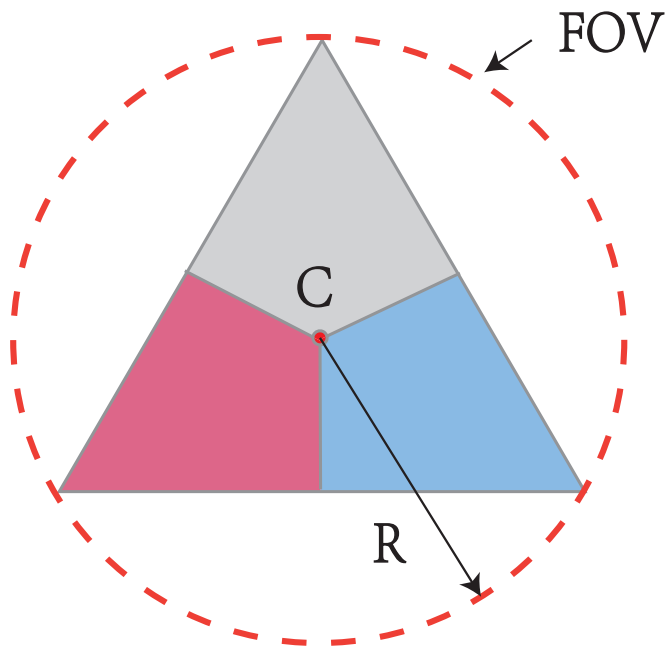

Figure 5.4: Equilateral triangle is made of three congruent Voronoi cells.

In this stage, we use a method similar to [50] to estimate the minimal number of waypoint locations with coordinates on a given plane $z_p = z$ that is parallel to the ground first, then to find the minimum altitude $z$ of sensing with the achieved 2D coordinate. The vantage waypoint set is illustrated as $P = \{p_1, ..., p_n\}, p_i \in \mathbb{R}^3$ with 2D coordinate $(x_i, y_i)$.

The constructed triangulation consists of equilateral triangles, and the length of the side of all the triangles is $\sqrt{3}R$, where $R$ is the radius of the vision-cone as defined in (5.3). We denote the direction of a side of a triangle with respect to a certain direction (such as the $x$-axis) as the angle of the triangle and we use $\lambda \in [0, \frac{\pi}{3})$ to represent this angle. Additionally, we denote $(x_0, y_0)$ as the coordinate of one of the vertices (see Figure 5.3). Then, a triangulation can be represented by $\Delta(\lambda, x_0, y_0)$. Moreover, in any triangle, the line segments connecting the center point, denoted by $C$, and the vertices divide the triangle into three congruent Voronoi cells. For instance, we use different colors to indicate these cells in Figure 5.4. In an equilateral triangle with side length $\sqrt{3}R$, the distance between any two points in the same Voronoi cell is within $R$ the radius of the FOV. It is obvious that the drone located at the same Voronoi cell to any point of the area of interest can cover at least this point of the area.

Starting from $z = Z_{min} = 120m$, $\lambda, x_0, y_0$ are randomly generated, and the algorithm in [50] are applied. The minimum number and the positions of waypoints are determined while decreasing the altitude $z$ by $1m$. For comparison, the Voronoi cell-based approach [40] is applied to the same environment. The corresponding number of waypoints with altitude is provided in Figure 5.5(c) and (d). It is apparent that the method of [40] in Figure 5.5(a) needs more waypoints than the proposed approach in Figure 5.5(b) to cover the same area at a similar altitude. The proposed method is better.

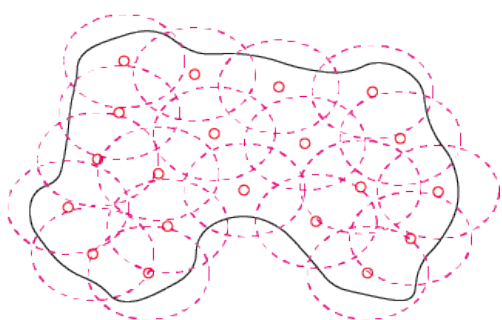

(a) Locations of 20 waypoints at 109m by [40].

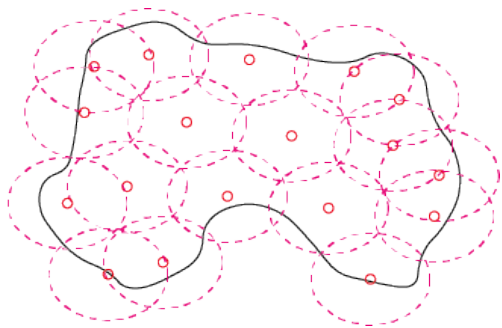

(b) Locations of 18 waypoints at 109m by the proposed algorithm.

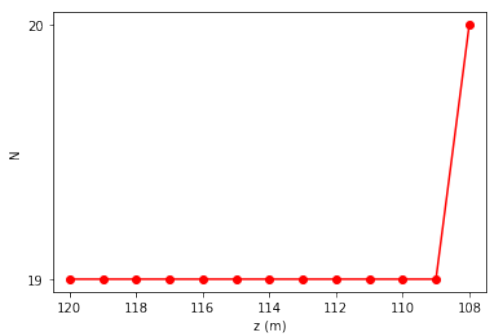

(c) The number of waypoints $N$ versus altitude $z$ by [40].

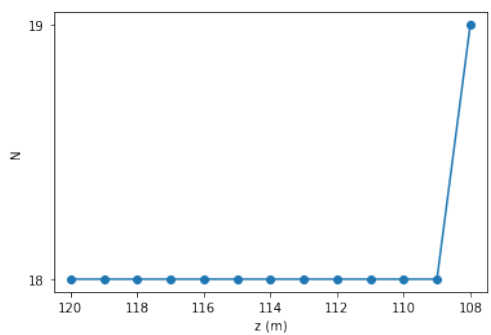

(d) The number of waypoints $N$ versus altitude $z$ by the proposed algorithm.

Figure 5.5: Locations of waypoints by two different algorithms. The red dash circles are the coverage areas of UAVs.

### 5.2.2.2 Stage Two: UAV Path Planning for Surveillance

We consider that the UAVs fly along Dubins paths. DTSP, as an extension of the Traveling Salesman Problem (TSP), assumes the salesman travels along the Dubins path, where the travel cost is proportional to length of the path [255]. The algorithms to derive the Dubins path can determine the shortest path connecting vantage waypoints generated in the first phase.

We briefly discuss two algorithms to address the DTSP problem, which are the alternating algorithm [255] and the spiral algorithm [49]. The alternating algorithm is an approximation algorithm to solve DTSP with given lower and upper bounds on solution quality. This algorithm computes the sequence of visits first uses the optimal solver for Euclidean TSP (ETSP) and then connects two waypoints by either an alternating straight-line and Dubins arc segments. The spiral algorithm is another popular DTSP surveillance algorithm. It links the chain of the convex hull of the given sets of waypoints. The shape of the resulting path resembles a spiral respecting the turning radius constraints.

When the waypoints are spare, i.e., when two waypoints are far enough with respect to the turning radius, the alternating algorithm computes the shortest path. By contrast, when the waypoints are dense, the spiral algorithm has been shown to be superior. In addition, the spiral algorithm is also more suitable for surveillance duty at low altitudes. The clustered spiral-alternating algorithm [256], which is actually a combination of the above two algorithms, is used in this paper to improve the effectiveness and efficiency of surveillance. To apply this algorithm we need to cluster the waypoints manually or by other methods. After clustering, waypoints in a cluster are dealt by the spiral algorithm. After this, the spiral paths for different clusters are combined, and the alternating line segments are replaced with Dubins segments to form the shortest path. From the observed result of total path length, the clustered spiral-alternating algorithm is always shorter than the original spiral algorithm in clusters. Compared to the alternating algorithm, the clustered spiral-alternating algorithm has fewer sharp turns.

The proposed scheme is also applicable when the target area consists of multiple disjointed areas (no overlaps). One of the most straightforward ways is to assign as many UAVs as the disjoint areas. The merits of the method include its simplicity, adaptability, and robustness against restricted communication among UAVs.

### 5.2.3 Simulation Results

In this section, computer simulation results are presented to demonstrate the effectiveness of the proposed approach. To confirm the performance of our surveillance strategy in a complex environment, we carry out verification in the following scenario.

#### 5.2.3.1 Single Area Surveillance

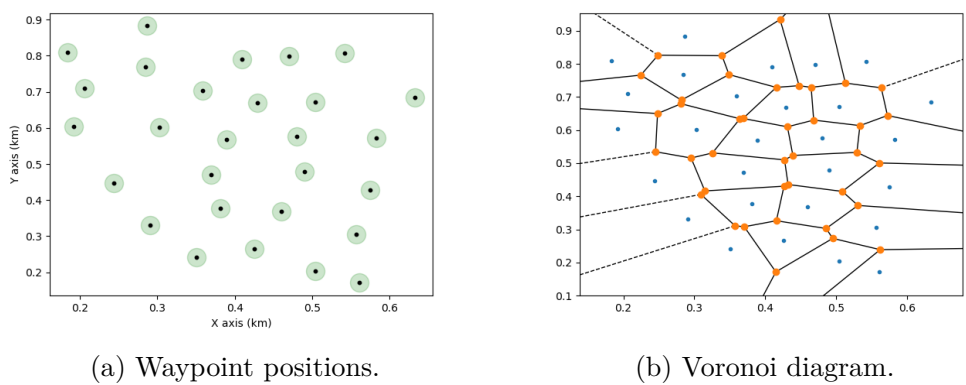

(a) Waypoint positions.              (b) Voronoi diagram.

Figure 5.6: Single-area surveillance scenario.

We first consider the vantage waypoints in a $700 \times 900$ m$^2$ single area as shown in Figure 5.6(a), and Figure 5.6(b) demonstrates the Voronoi diagram of the waypoints.

Figure 5.7 shows the single UAV surveillance path generated by the spiral algorithm [255], the alternating algorithm [49], the genetic algorithm [257], Grey Wolf Optimizer (GWO) algorithm in [258], and the clustered spiral-alternating algorithm [256], respectively. We evaluated the selected clustered spiral-alternating algorithm with mentioned four algorithms in terms of path length and number of sharp turns. The path generated by clustered spiral-alternating algorithm Figure 5.7e is shorter than the spiral algorithm (SA) Figure 5.7a and Grey Wolf Optimizer (GWO) algorithm Figure 5.7c, and fewer sharp turns than the alternating algorithm (AA) Figure 5.6b and genetic algorithm (GA) Figure 5.7d. The straight-line segments and the Dubins segments are represented by yellow and red segments, respectively. The average execution times for the above methods are shown in Table 5.1. The average execution times for the above methods are shown in Table 5.1. The SA and AA are faster than the GWO, GA, and the proposed method. However, the proposed method has a far shorter path length than the SA method and fewer sharp turns than the AA method. The proposed method is available as an option when there is a tradeoff between the execution time and optimal path.

Table 5.1: Average execution times (seconds) for the different algorithms

| Algorithm | Time (s) |
| --- | --- |
| SA | 3.21 |
| AA | 3.69 |
| GWO | 16.14 |
| GA | 10.36 |
| Proposed | 4.84 |

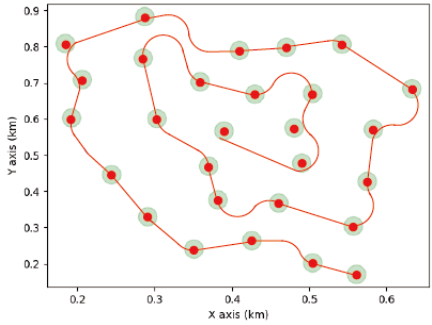

(a) Spiral algorithm: 2258.16 m.

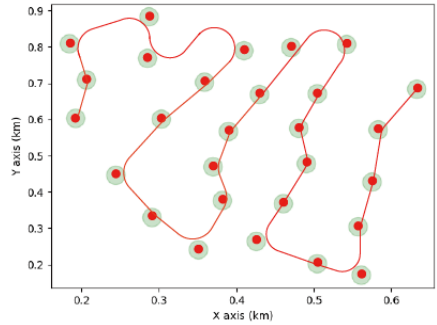

(b) Alternating algorithm: 1825.80 m.

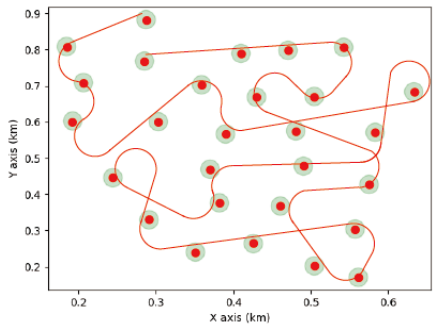

(c) Grey Wolf Optimizer (GWO) algorithm: 3200.70 m.

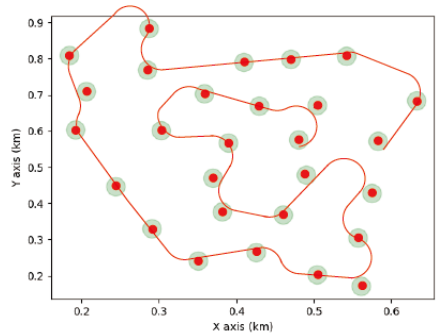

(d) Genetic algorithm:1900.30 m.

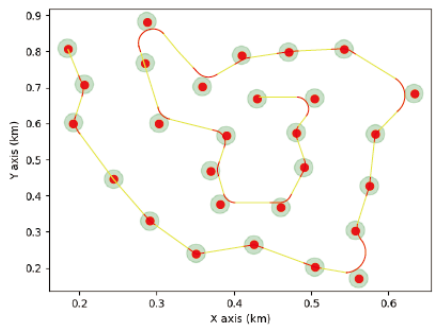

(e) Cluster spriral-alternating algorithm: 1827.32 m.

Figure 5.7: UAV surveillance trajectory—Single UAV.

### 5.2.3.2 Multiple Disjoint Areas Surveillance

The multiple disjoint surface area in a $600 \times 600 \text{ m}^2$ terrain is shown in Figure 5.8. In the first stage, the minimum number and position of the waypoints of our region of interest are generated and shown in Figure 5.8a, and the $\delta$-neighborhood of each individual vantage waypoint location is in Figure 5.8b. Then, the path is achieved using clustered spiral-alternating algorithm and Dubins curves. For our results in Figure 5.8c, two clusters are assigned to two UAVs, respectively. In addition, the coverage of the area of interest is achieved when both finish their surveillance circle. It makes the quick parallel surveillance possible and the subsequent task allocations simple.

In the single UAV scenario, we can use only one drone to carry out the surveillance. With the consideration of computation efficiency, we decide to keep one obtained path and regenerate the other cluster's path. The final surveillance planning with Dubins curves by the proposed algorithm shows in Figure 5.8d. The simulation result demonstrated the validation of the algorithm.

## 5.2.4 Section Summary

We consider the dynamic coverage problem over the target area, where UAVs keep flying through each pre-defined waypoint during surveillance and take video. We present a two-phase strategy to completely cover the target area with a given altitude. We propose the following strategy based on a decomposition of the surveillance problem into a variant of the waypoint determination problem and an instance of the Dubins Travelling Salesman Problem. Dubins Travelling Salesman Problem can be used to determine the shortest path connecting these waypoints. This instance of Dubins-vehicle Travelling Salesman Problem (DTSP) is solved by introducing

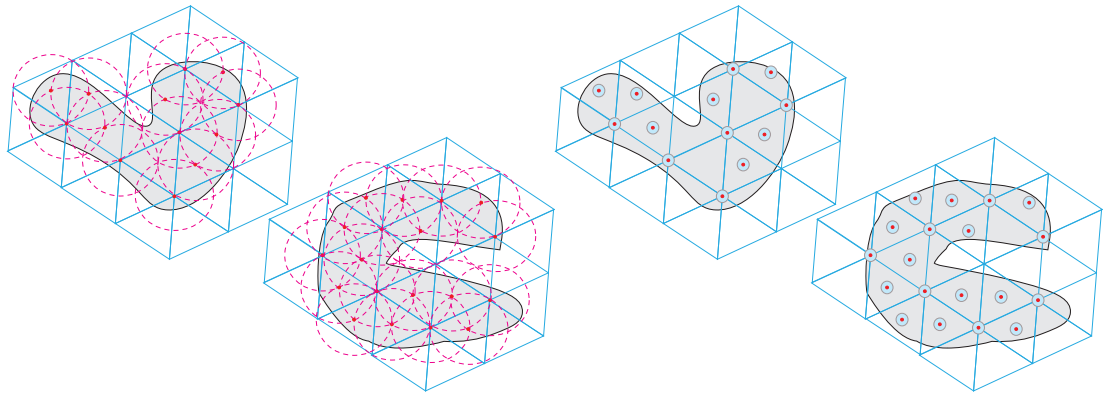

(a) Vantage waypoint set of the target area (b) $\delta$-neighbourhood of vantage waypoint set

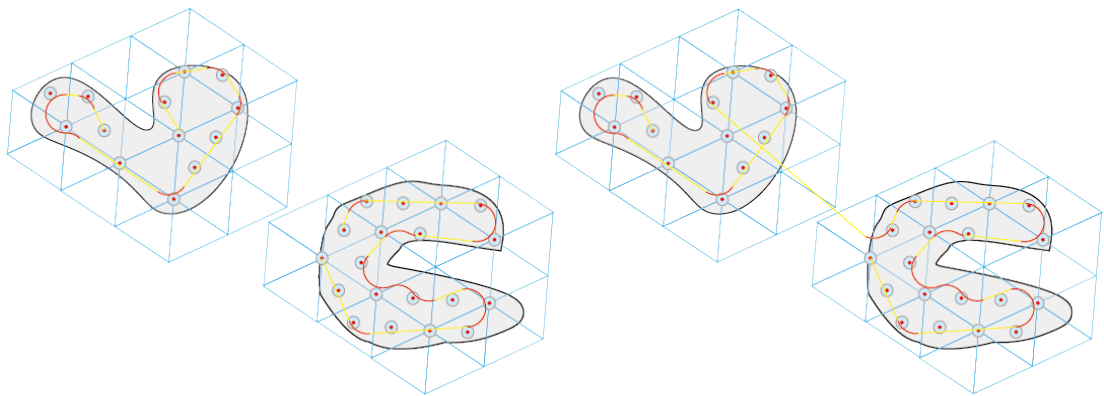

(c) Two UAV, two paths          (d) Proposed UAV path

Figure 5.8: Proposed UAV surveillance in multiple disjoint surface area

clustered spiral-alternating algorithm, and the trajectory respecting the kinematic constraints of the UAV. The main contribution of this section is to determine the minimum number of waypoints necessary to be visited and make trajectory at a given altitude to monitor the region. The performance and effectiveness of our algorithm have been confirmed by extensive computer simulations in different scenarios.

## 5.3 Surveillance Algorithm at different altitudes

This section considers a more challenging and realistic situation involving surveillance in geometrically complex environments, such as mountainous terrains and urban regions. Because the video camera can only see the points within its cone-shaped field of view (FOV), the FOV can be reduced when facing any kind of obstacle. Similar to our previous section, we address the mentioned problems separately, decoupling them into the waypoint generation part and trajectory determination part. Some assumptions and definitions had been claimed in Section 5.2 so they are used directly without explaining. This section is derived from [35].

### 5.3.1 Problem Statement

We consider the dynamic coverage problem in the target area, where UAVs keep flying through each pre-defined waypoint during surveillance and take video. The vehicle's kinematic model at the point $p = (x, y, z)$ can be described as follows:

$$\begin{bmatrix} \dot{x} \\ \dot{y} \\ \dot{z} \end{bmatrix} = v \begin{bmatrix} \cos\theta \cos\psi \\ \sin\theta \cos\psi \\ \sin\psi \end{bmatrix} \tag{5.5}$$

where $\theta$ and $\psi$ are the turning angle and the pitch (climb/dive) angle, respectively. The state of our drone is $s = (p, \theta, \psi)$.

The model of the terrain based on several assumptions is described in the last section. In the addressed reconnaissance and surveillance scenario, our UAV equipped with a downward-facing video camera with a certain visibility angle. It can monitor the relatively small targets of interest on the ground with the required level of details within its FOV. The onboard camera with a given visibility angle $0 < \alpha < \pi$ is given in Figure 5.9. The drone at $(x, y, z)$ can only see points $(x_g, y_g, z_g)$ of the target area that are inside the cone-shaped field of view (FOV) of radius

$$R(z_g) := (z - z_g) \cdot \tan(\frac{\alpha}{2}),$$

$$z > z_g.$$

(5.6)

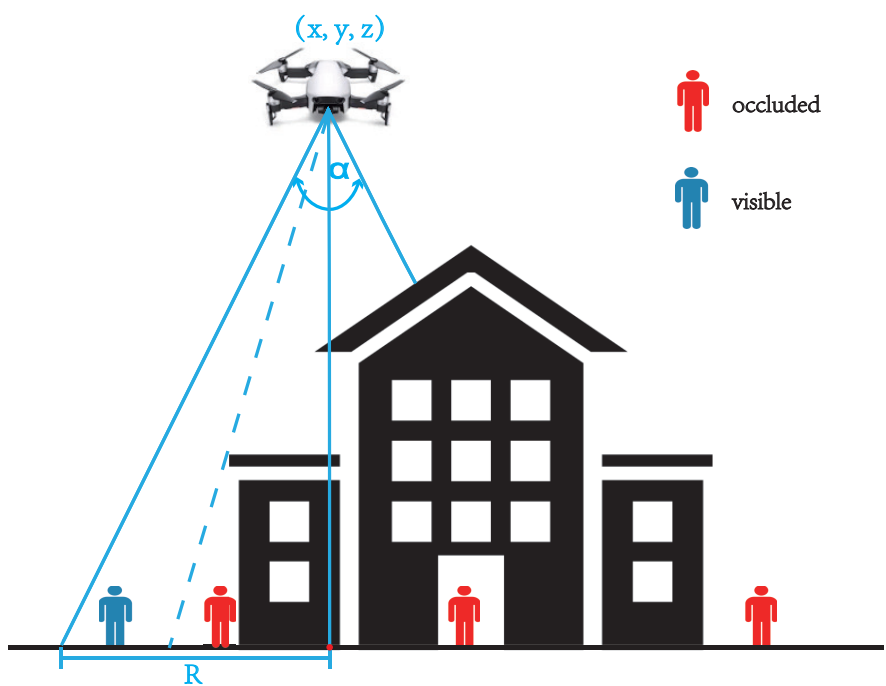

Figure 5.9: Occlusion effects on camera sensing

The problem becomes to find the trajectory to visit $\delta$-neighborhood of all assigned vantage points, which involves the optimization of the sequence of visits, i.e., a variant of the combinatorial traveling salesman problem. The final trajectory is connected by a sequence of piecewise cubic Bézier curves that traverse the $\delta$-neighborhood of all assigned vantage points. A $n$ degree Bézier curve can be

parametrized as:

$$\mathcal{T}(t) = \sum_{i=0}^{n} \mathbf{P}_i b_{i,n}(t) \tag{5.7}$$

where $\mathbf{P}_i$ stands for the i-th control point, and $0 \leq t \leq 1$. $b_{i,n}(t)$ is named Berstein polynomial and defined as follows:

$$b_{i,n}(t) = \begin{pmatrix} n \\ i \end{pmatrix} t^i (1-t)^{n-i} \tag{5.8}$$

and $0 \leq t \leq 1$. The binomial coefficient is given by

$$\begin{pmatrix} n \\ i \end{pmatrix} = \frac{n!}{i!(n-i)!} \tag{5.9}$$

The utilized cubic Bézier curve is defined by four control points ($\mathbf{P}_0, \mathbf{P}_1, \mathbf{P}_2, \mathbf{P}_3$), and can be expanded as

$$\mathcal{T}(t) = \mathbf{P}_0 (1-t)^3 + 3\mathbf{P}_1 t(1-t)^2 + 3\mathbf{P}_2 t^2 (1-t) + \mathbf{P}_3 t^3. \tag{5.10}$$

As the final trajectory $\mathcal{T}$ is closed and smooth curve, which consists of $n$ Bézier curves, two consecutive curves $\mathcal{T}_i$ and $\mathcal{T}j$ with control points ($\mathbf{P}_0^i, \mathbf{P}_1^i, \mathbf{P}_2^i, \mathbf{P}_3^i$) and ($\mathbf{P}_0^j, \mathbf{P}_1^j, \mathbf{P}_2^j, \mathbf{P}_3^j$) should be connected at the same endpoint:

$$\mathbf{P}_3^i = \mathbf{P}_0^j \tag{5.11}$$

And the tangents $\mathbf{t}_a^i$ and $\mathbf{t}_b^i$ with length $|t_a^i|$ and $|t_b^i|$ of $\mathcal{T}_i$ and $\mathcal{T}j$ must point to the same direction to connect a sequence of ézier curves into a smooth path:

$$l_a^j t_b^i = l_b^i t_a^j \tag{5.12}$$

where

$$\mathbf{t}_a^i = \mathbf{P}_1^i - \mathbf{P}_0^i, \quad \mathbf{t}_b^i = \mathbf{P}_3^i - \mathbf{P}_2^i, \quad l_a^i = \left| t_a^i \right|, \quad l_b^i = \left| t_b^i \right|. \tag{5.13}$$

Compared with the Dubins model with constant velocity, the multi-motor model can decelerate to make turns and accelerate on fairly straight paths. For the UAV model applied, we prefer the multi-rotor UAV to the curvature-constrained Dubins model. Therefore, the aim is to find the fastest route for the UAV with the maximal vehicle velocity and acceleration limits, rather than the shortest path with Dubins velocity constraints. Therefore, instead of minimizing the length of the trajectory, the expected time to travel the path is considered in this section.

## 5.3.2 Surveillance Algorithm

In this section, we try to find the vantage waypoint set to fully cover the area of interest first, and then plan a smooth trajectory along with these locations at different altitudes as fast as possible, such that the completion time to visit all the locations is minimal. We decompose the surveillance problem into drone version 3D Art Gallery Problem and an instance of the combinatorial traveling salesman problem.

### 5.3.2.1 Stage One: Vantage Waypoint Set Generation

The problem of finding the vantage waypoint set can be viewed as a drone version of the 3D Art Gallery Problem, which has been shown to be NP-hard [259]. The approximation approach is always employed, in this section, we use a method similar with [41] to estimate the minimal number of the waypoint locations by 3-coloring method [260], and generate the vantage waypoint set $P = \{p_1, ..., p_n\}, p_i \in \mathbb{R}^3$ in two major steps. In the first step, the objective is to determine the 2D coordinates $(x_i, y_i)$ of each vantage waypoints. The second step is to find the best altitudes $z_i$ of sensing with the achieved 2D coordinates $(x_i, y_i)$.

We assumed the terrain $\mathcal{G}$ is a polygon with $n$ vertices, and $\mathcal{G}_1, ..., \mathcal{G}_i$ are obstacles inside $\mathcal{G}$ with $n_1, ..., n_i$ vertices, respectively. Let $\mathcal{E}_1, ..., \mathcal{E}_i$ be inside polygons of $\mathcal{G}_1, ..., \mathcal{G}_i$, respectively. And each $\mathcal{E}_i$ has same number of vertices $n_i$ as $\mathcal{G}_i$. We can consider $\mathcal{E}_i$ as and "top" face, and $\mathcal{G}_i$ as the corresponding "bottom" face of the each obstacle polyhedron model. $\mathcal{A}$ is obtained from $\mathcal{G}$ without $\mathcal{G}_1, ..., \mathcal{G}_i$, which is a non-convex polygon with $i$ "holes". Let $c > 0$ be a constant denoted the relative even area's altitude, and $c + c_2 \leq Z_{min}$ holds. And let $\hat{d}_e := \max \left\{ d_e, \hat{d}_l \right\}$, $d_e$ is the maximum distance between $(x_i, y_i)$ to the corresponding vertices of $\mathcal{E}_i$, and $b$ is the maximum altitude of the terrain points corresponding to the $\mathcal{E}_i$ and its side quadrilaterals. Let $d_l$ be the maximum length of the triangulation triangles sides whose vertex is one of the two end points. The developed vantage waypoint set generation algorithm can be found in the Algorithm 3.

---

**Algorithm 3** Vantage Waypoint Set Generation

---

1: **procedure** STEP I

2:     Construct polygon $\mathcal{P}$ without "holes" by $i$ non-intersect diagonals with $n + n_1 + ... + n_i + 2i$ vertices

3:     Cut $\mathcal{P}$ into triangulation $\mathcal{T}$, whose vertices are $\mathcal{P}$'s, and sides are either $\mathcal{P}$'s or its diagonals.

4:     Build the dual graph of $\hat{\mathcal{T}}$ by enlarging $\mathcal{T}$ by the vertices of $\mathcal{E}_i$

5:     Paint the vertices of the triangulation $\hat{\mathcal{T}}$ by 3-coloring method in [260]

6:     The minimum number of vertices subset of the three is selected as the 2D coordinates $(x_i, y_i)$ of the vantage waypoint set

7: **end procedure**

8: **procedure** STEP II

9:     **while** $z_i \leq Z_{min}$ **do**

10:         **if** $(x_i, y_i)$ is not a vertex of any polygons $\mathcal{E}_i, \mathcal{G}_i$ **then**
            $z_i := \max\left\{Z_{\min}, c + \frac{d_l}{\tan\left(\frac{\alpha}{2}\right)}\right\}$

11:         **else if** $(x_i, y_i)$ is a vertex of some polygons $\mathcal{E}_i, \mathcal{G}_i$ **then**
            $z_i := \max\left\{Z_{\min}, b + c_2 + \frac{\hat{d}_e}{\tan\left(\frac{\alpha}{2}\right)}\right\}$

12:         **else**
            $z_i := \max\left\{Z_{\min}, a + c_2 + \frac{\hat{d}_l}{\tan\left(\frac{\alpha}{2}\right)}\right\}$

13:         **end if**

14:     **end while**

15: **end procedure**

16: **return** vantage waypoint set $P = \{p_1, ..., p_n\}, p_i = (x_i, y_i, z_i)$

---

### 5.3.2.2    Stage Two: UAV Path Planning for Surveillance

As mentioned in the previous section, the final trajectory will go through the $\delta$-neighbourhood of each individual vantage waypoint location to reduce the comple-tion time. We do not directly consider the coverage problems introduced by visiting the neighborhoods. This variant of the combinatorial traveling salesman problem is solved by introducing Self-Organizing Map (SOM) and Bézier curves, and the trajectory respecting the kinematic constraints of the UAV.

As we assumed that the UAV will return back to the initial location $p_1$ after the complete reconnaissance tour, $\delta = 0$ for the initial location. With the vantage waypoint set $P = \{p_1, ..., p_n\}, p_i \in \mathbb{R}^3$ generated above, and the given initial location $p_1$ and $\delta$, we can determine the final trajectory $\mathcal{T}$ as a sequence $\Sigma = (\sigma_1, \ldots, \sigma_n)$ of Bézier curves $\mathcal{T}_i, 1 \leq i \leq n$. The final trajectory $\mathcal{T} = (\mathcal{T}_{\sigma_1}, \ldots, \mathcal{T}_{\sigma_n}), 1 \leq \sigma_i \leq n$, we need to minimize the estimation of the travel time $\mathbf{E}(\mathcal{T})$, which can be determined from (5.10). In order to simplify the model, we employ the Model Predictive Controller (MPC) for path following, and the vertical and horizontal movements of the drone are individually considered. we denote $a_{ver}, v_{ver}, a_{hor}, v_{hor}$ as maximal vertical and horizontal accelerations and speeds, respectively. We also assume that the initial and final velocity of the drone is zero, so that the drone will start from the initial location $p_1$ with zero velocity and return back to it in the end. The estimation of the travel time $\mathbf{E}(\mathcal{T})$ and the profile of the velocity can be computed by the method in [261] by maximum possible tangent acceleration $a_{tan} = \sqrt{a_{hor}^2 - a_{rad}^2}$, $a_{rad}$ is the radial acceleration.

The adaptation can be considered as a movement of the waypoint locations towards the alternate location $s_p$ and a new location of each adapted waypoint location $\nu$ becomes $\nu'$ and it follows the standard SOM learning [180].

$$\nu' = \nu + \mu f(\sigma, d)\, (s_p - \nu) \tag{5.14}$$

where $\mu$ is the learning rate, $\sigma$ is the learning gain, $d$ is the distance of $\nu$ from the winner waypoint location $\nu^*$, and $f(\sigma, d)$ is the neighboring function

$$f(\sigma, d) = \begin{cases} e^{\frac{-d^2}{\sigma^2}} & \text{for } d < 0.2M \\ 0 & \text{otherwise} \end{cases} \qquad (5.15)$$

where $M$ is the current number of neurons in the SOM.

The surveillance planning algorithm will stop the adaptation if $i \leq i_{max}$ or $\nu^*$ are negligibly close to their respective $s_p$, or all winner waypoint locations are inside the $\delta$-neighborhood of the respective initial waypoint location. Local Iterative Optimization (LIO) [262] is a procedure that optimizes the whole trajectory locally, e.g., it can consecutively optimize $\theta_i$, $\psi_i$, $l_a^i$, and $l_b^i$ in the loop with waypoint $\nu_{i-1}$, $\nu_i$, and $\nu_{i+1}$. The reason (5.5) can optimize variables $\theta_i$, $\psi_i$, $l_a^i$ independently is tangent vector $t_a^i$ and $t_b^{i-1}$ implicitly satisfy the smooth constraint (5.12). $t_a^i$ and $t_b^{i-1}$ are related to the same waypoint $\nu_i$.

Otherwise, go to Step 3. An intersection of the straight-line segment $(s_p, p)$ with the sphere in $\mathbb{R}^3$ shaped $\delta$-neighborhood of p is used to determine the alternate location $s_p$ towards which the network is adapted instead of p to save the travel time.

We can easily extend the described mechanism into multiple UAV versions by two principal methods:

1 Once the surveillance path for single UAV is generated, we can distribute multiple UAVs travel along the same path as the single UAV, but with different initial position to avoid collisions. To avoid collisions, the initial deployment of UAVs must be coordinated with the drone's velocity and the length of the path, by, for instance, evenly spacing the appropriate number of UAVs along the determined trajectory. Thereafter, each UAV can perform its surveillance duty independently without further coordination. This method can markedly reduce the surveillance circle or duration, and significantly increase the frequency and intensity of surveillance.

2 The vantage waypoint set can be partitioned into several subsets, and dedicated UAV(s) can traverse through each subset independently. In the case of multiple UAVs, we may use the aforementioned method to perform collision-free monitoring tasks.

We also analyze the complexity of the algorithm. In each learning epoch, the computational complexity depends on the number of waypoints $n$ and the number of neurons in the SOM network $M$. Notice the SOM is a two-layered neural network, whose input layer is the locations of vantage waypoints $P = \{p_1, ..., p_n\}, p_i \in \mathbb{R}^3$, and output layer is an array of adapted waypoint locations $\mathcal{N} = \{\nu_1, ..., \nu_M\}$. As the algorithm regenerates the loop $N$ with winner waypoints between $n - 2n$ (see from line 8 of Algorithm 4), the complexity of the path-planning procedure is $O(n^2)$.

---

**Algorithm 4** UAV Path Planning Algorithm

---

1: Create the loop $\mathcal{N}$ with $n$ waypoint locations around the initial location $p_1$

2: Set the learning gain $\gamma = 12.41n + 0.6$, the learning rate $\mu = 0.5$, and the gain decreasing rate $\eta = 0.1$. Set the epoch counter $i = 1$, $i_{max} = 100$.

3: **while** the termination condition hasn't been reached **do**

4:     **for** $p \in \Pi(P)$ **do**

5:         **for** each learning epoch **do**

6:             determine $\nu^*$ and $s_p$

7:             Adapt $\nu^*$ and its neighbours towards $s_p$ using (7)

8:             remove all non-winner waypoint locations and perform LIO-based optimization of the trajectory

9:             Update learning parameters: $\gamma = (1 - \eta)\gamma, i = i + 1$

10:         **end for**

11:     **end for**

12: **end while**

13: **return** final trajectory $\mathcal{T}$

---

## 5.3.3  Simulation Results

To verify the effectiveness of the proposed algorithm, two simulation scenarios are given in this section. The target area is 20 m by 20 m terrain with $i = 3$ random shaped obstacles are shown in Figure 5.10a, and the obstacles have $n_1 = 3, n_2 = 4, n_3 = 4$ vertices, respectively. The average execution times for the above methods are shown in Table 5.2, and the average velocities and coverage times for the different algorithms are shown in Table 5.3. Furthermore, the simulation is conducted with the following parameters in Table 5.4.

Table 5.2: Average execution times (seconds) for the different algorithms

| Algorithm | Single-Area Single UAV | Single-Area Multiple UAVs (Two UAVs, two paths) |
|-----------|------------------------|--------------------------------------------------|
| TOS | 3.48 | 4.53 |
| 3DAA | 4.74 | 5.12 |
| EA | 5.30 | 5.91 |
| Proposed | 5.14 | 6.35 |

### 5.3.3.1   Single-Area Single UAV

To confirm the performance of our monitoring strategy in a complex environment, we performed validation in the following scenario. First, we apply the proposed Vantage Waypoint Set Generation in step one to obtain the waypoint set that can fully cover the target area. Figure 5.10b shows the $\delta$-neighborhood of all 10 waypoint locations at different heights from 6.4 m to 23.3 m. Secondly, we compare the proposed UAV path with the three mentioned methods' paths in terms of operation time and surveillance quality. In accordance with surveillance quality, we take the smoothness of the path, path length, and whether the uncovered area exits into consideration. Under the same velocity profiles as our proposed method, the time-optimal strategy (TOS) [51], evolutionary algorithm (EA) [263], and 3D Alternating algorithm (3DAA) then run under the same environment. The proposed algorithm (see Figure 5.10c) and evolutionary algorithm (see Figure 5.10e) are superior to the time-optimal surveillance path-planning algorithm (see Figure 5.10d) and 3D alternating algorithm see Figure 5.10f) in terms of operation time. The time-optimal surveillance path-planning algorithm is the most straightforward method and smoothest path among them. Nonetheless, it inevitably neglects the problems caused by occlusion, so we may need to apply the geometric calculation to calculate the uncovered part and deploy other unmanned aerial vehicles to cover it completely. On the other hand, the evolutionary algorithm has a longer path and more sharp

Table 5.3: Average velocity (meter/second) and coverage time (second) for the different algorithms

| Algorithm | Scenario | Average velocity (m/s) | Minimum coverage time (s) |
|---|---|---|---|
| TOS | Single-Area Single UAV | 6.5 | 22.2 |
| | Single-Area Multiple UAVs (Three UAVs, three paths) | 5.1 | 7.4 |
| 3DAA | Single-Area Single UAV | 6.5 | 13.6 |
| | Single-Area Multiple UAVs (Two UAVs, two paths) | 6.5 | 8.7 |
| EA | Single-Area Single UAV | 6.6 | 12.5 |
| | Single-Area Multiple UAVs (Two UAVs, two paths) | 6.8 | 6.5 |
| Proposed | Single-Area Single UAV | 6.8 | 7.9 |
| | Single-Area Multiple UAVs (Two UAVs, two paths) | 6.8 | 4.2 |

turns.

### 5.3.3.2 Single-Area Multiple UAVs

In this scenario, multiple UAVs are being used for full coverage reconnaissance and surveillance missions. The same waypoint positions are shown in Figure 5.10b. Figure 5.11a shows two drones in different initial positions share the same route. Figure 5.11b shows the situation where each UAV has a separate track to cover part of the terrain. Moreover, when both complete their monitoring circle, the coverage of the region of interest is realized. In both scenarios, the duration of covering the region of interest is obviously increased, so the average time of covering any point between two continuous times is reduced by about 47%. In other words, the points of interest are monitored more often. We also deploy three UAVs to cover each individual obstacle as a comparison. However, the uncovered part due to the overlapping of the flight surfaces is inevitable. Another two methods are applied for two UAVs to conduct the surveillance duty in Figure 5.11d,e. In this case, the evolutionary algorithm generates a smoother path than the proposed one. However, the proposed algorithm still outperforms the 3D alternating algorithm with a shorter length and fewer sharp turns.Table 5.3 shows the average velocity and minimum coverage time in both scenarios. The proposed approach shown in Figure 5.10c has the minimum coverage time in the single UAV scenario, and Figure 5.11a has the minimum coverage time in the multiple UAV scenario. Table 5.2 shows the computational time of the simulation above. The time-optimal strategy and 3D alternating algorithm take less time to cover the target area than the other two methods because of the lower computation load. Moreover, the evolutionary algorithm compromises the surveillance quality with computation time.

Table 5.4: Simulation Parameters

| camera | | | | |
|---|---|---|---|---|
| $\theta$ | $H_{min}$ | $c_1$ | $c_2$ | $c$ |
| $\frac{\pi}{2}$ | 4m | 1m | 0.5m | 0.2m |

| mobility of UAV for proposed method | | | | mobility of UAV for compared method | | |
|---|---|---|---|---|---|---|
| $a_{ver}$ | $a_{hor}$ | $v_{ver}$ | $v_{hor}$ | $a_z$ | $v_z$ | $v_f$ |
| $0.5m/s^2$ | $1.2m/s^2$ | $0.5m/s$ | $1.2m/s$ | $0.5m/s^2$ | $0.5m/s$ | $1.2m/s$ |

## 5.3.4 Section Summary

In this section, we consider the reconnaissance and surveillance problem for UAVs flying over geometrically complex environments, such as mountainous terrains and urban regions. We present a two-phase strategy to completely cover the target area at different altitude. The main contribution of this section is to develop an occlusion-aware UAV surveillance strategy regarding the kinematic constraints of the UAV and the obstacles' occlusions. The UAV will visit a certain vantage waypoint set as fast as possible and ensure that any point in the area is seen from some position. The simulation result demonstrated the validation of the algorithm. The problem of finding the vantage waypoint set can be viewed as a drone version of the 3D Art Gallery Problem. The unsupervised learning and Bézier curves are used to generate a smooth and fast trajectory for the drone.

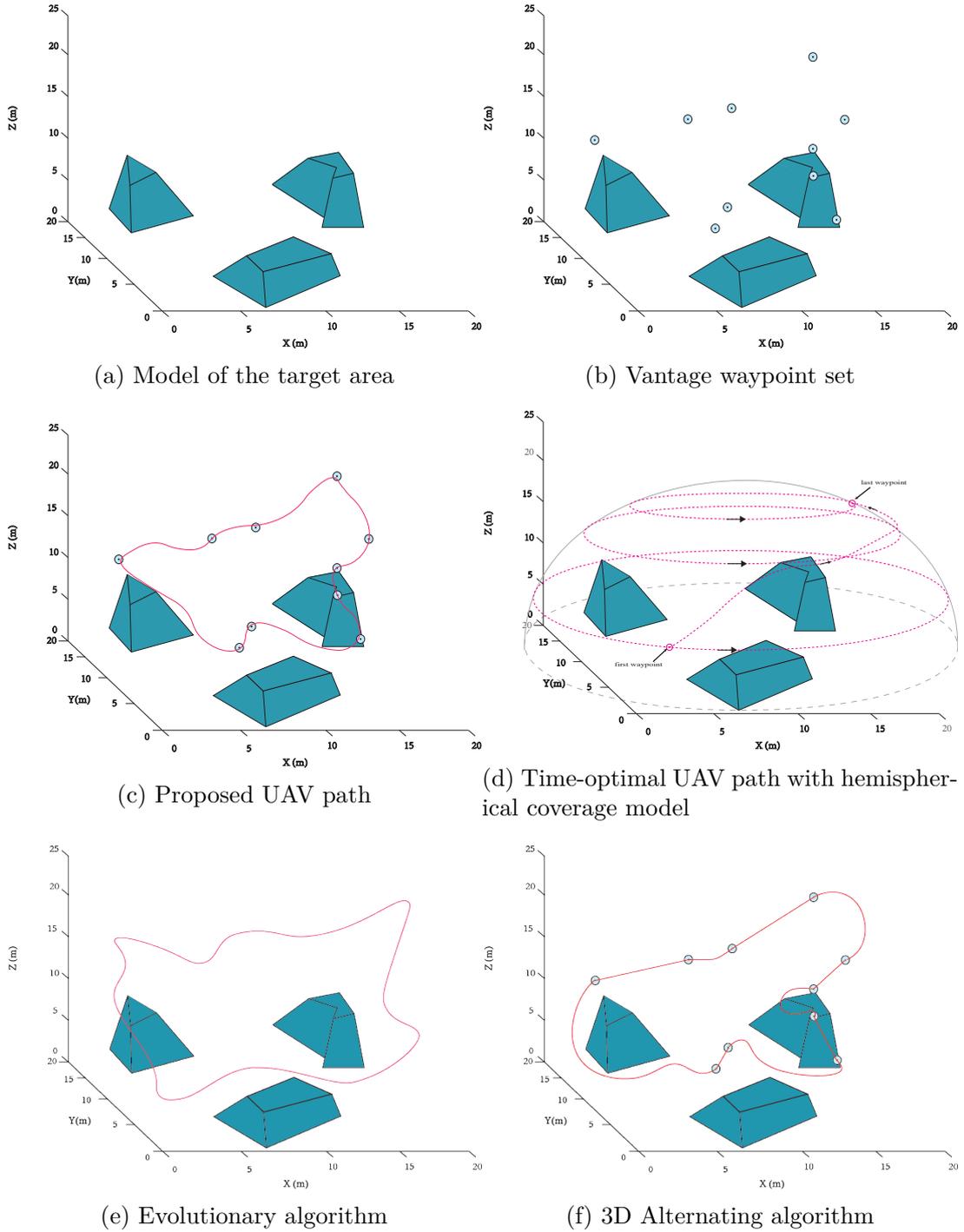

(a) Model of the target area

(b) Vantage waypoint set

(c) Proposed UAV path

(d) Time-optimal UAV path with hemispherical coverage model

(e) Evolutionary algorithm

(f) 3D Alternating algorithm

Figure 5.10: (a) the given environment, (b) the vantage waypoint set by the proposed method, UAV surveillance trajectory using (c) proposed strategy, (d) time-optimal strategy in [51], (e) evolutionary algorithm [263] and (f) 3D Alternating algorithm—Single UAV.

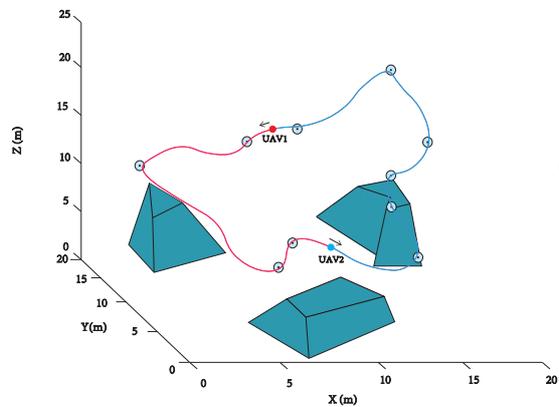

(a) Two UAVs share one path, the red/blue point is its initial position

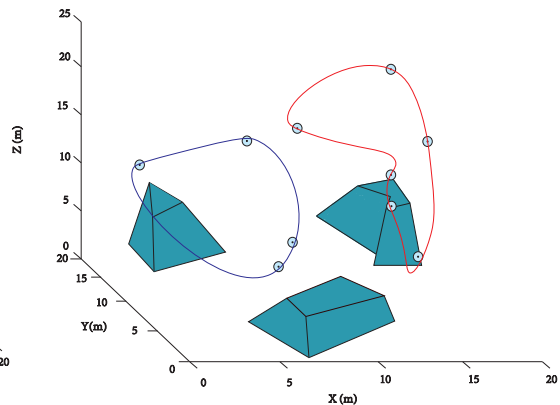

(b) Two UAV, two paths

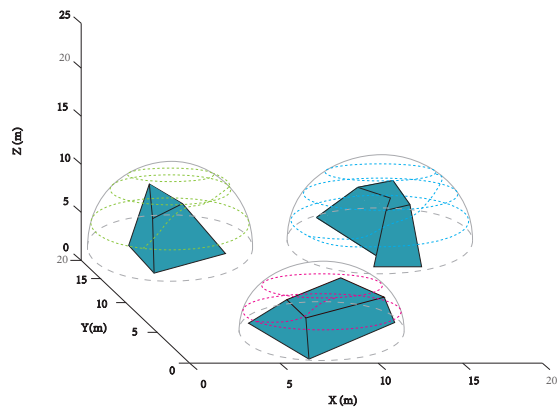

(c) Three UAV, three paths

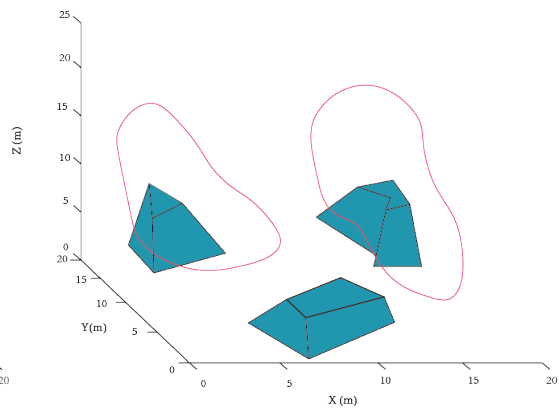

(d) Two UAV, two paths

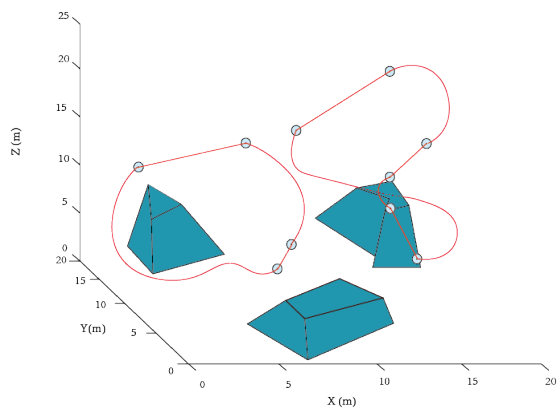

(e) Two UAV, two paths

Figure 5.11: UAV surveillance trajectories using (a,b) proposed strategy and (c) time-optimal strategy in [51], (d) evolutionary algorithm [263], and (e) 3D alternating algorithm—Multiple UAVs

# Chapter 6

# Path Planning for Self-Driving Cars with Convolutional Neural Network

## Contents

## 6.1   Motivation

This chapter is based on the publications [264, 265]. In this chapter, we study the path planning problem through another machine learning technique that we call convolutional neural network (CNN), which is different from previous chapters. As declared in the previous chapter, the convolutional neural network is one of the most popular deep neural network architectures, consisting different types of layers, including convolutional layers, flatten layers, pooling layers, fully connected layers. The main advantages of CNNs include higher accuracy for extracting features than standard flattened neural networks and save the time-consuming step to select intuitive features for the model. In other words, features can be learned automatically from training samples. The key drawback of CNNs is the requirement of vast amounts of training data. However, this problem can be solved properly by data augmentation.

This chapter employs a developed deep neural network CNN for an autonomous vehicle and navigates it by predicting the steering angle. The synthetic images used in our work are generated from the Udacity platform, which is an open-source simulator developed based on Unity. We develop a convolutional neural network that automatically extracts the features from the images and finds the dependencies for forecasting the steering angle to keep the vehicle running at the center of the lane. Our data augmentation methods provides a more generalized database for training, which saves the training time for individual conditions. The proposed model has been experimentally verified in several environments. Furthermore, the learned features can be transferred to environments that have never been seen before. It optimizes all processing steps in autonomous driving simultaneously and eases the burden on substantial programming efforts.

## 6.2   Problem Statement

In this chapter, we address the navigation problem of the autonomous vehicle, also known as the robot car, driverless or self-driving car, which shows profound potential to change to our society remarkably, not to mention the significant enhancements they could bring to the overall safety, efficiency, and convenience of transportation and transit systems. The final goal of the development of autonomous vehicles is to replicate the complex driving tasks of human. However, there are still many problems that need to be settled efficiently, including arduous challenges, including obstacle perception, decision-making, and control. Thus, the fully autonomous driving vehicles is still on the road [266].

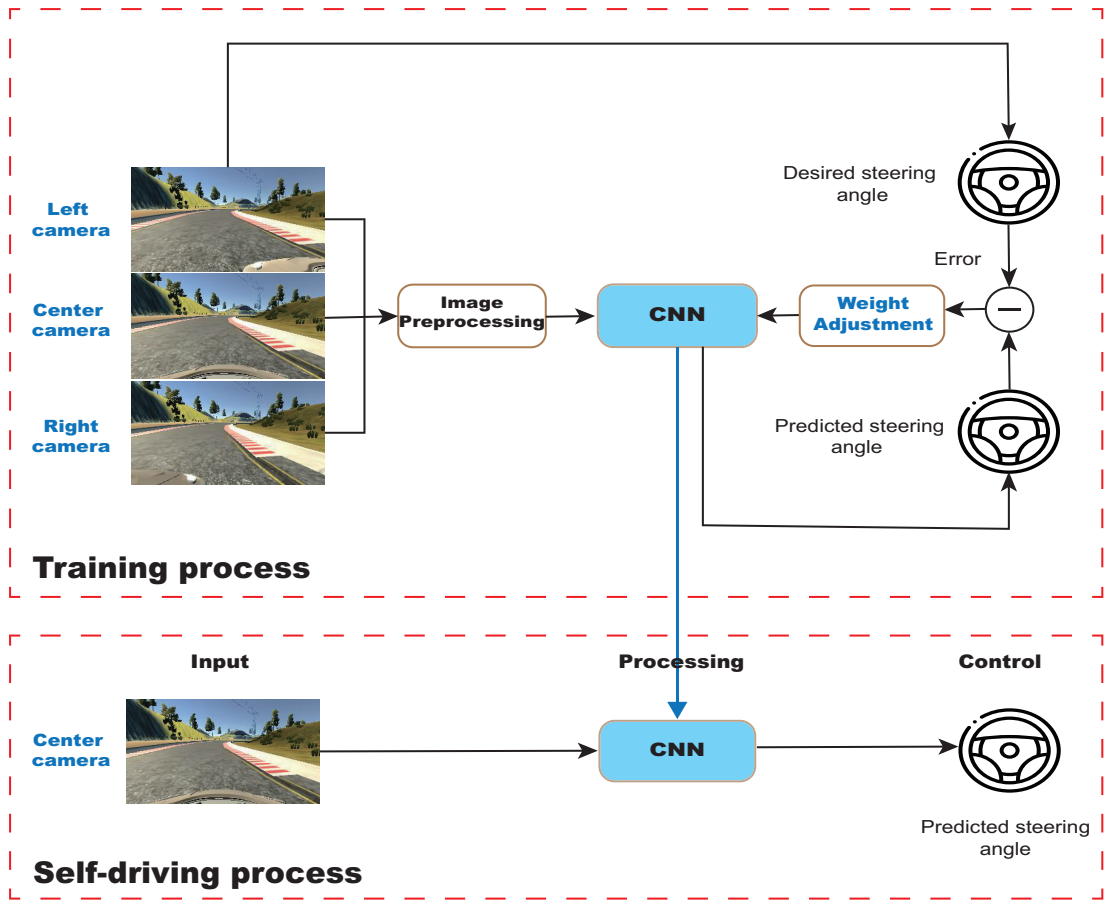

Figure 6.1: The overview proposed navigation scheme

This chapter proposes a deep neural network model to navigate the vehicle by predicting steering angles. Compared with the substantial programming burden of feature extractions and rule-settings, our model can learn human driver's behavior and generate its own prediction by extracting features from images taken from the vehicle's front windshield. The proposed scheme is decomposed into a training process and a self-driving process shown in Figure 6.1. The input data is obtained
from three cameras mounted on the left, right, and center of the front windshield
during the driving simulations. The collected images are processed by a series of
operations, such as resizing, balancing dataset, normalization, augmentation.

The preprocessed images with the corresponding labels of speed, steering angle,
acceleration rate, and braking rate are then sent into our convolutional neural net-
work for training. During this process, the steering angle is predicted and compared
with the desired one for that specific image. The weights of our model are adjusted
by the back propagation method to improve the accuracy of the CNN. Finally, the
trained network can automatically generate commands of steering angles from the
images captured in real-time during the self-driving process.

Udacity [267] is a Unity-based open-source platform for autonomous vehicle
simulation and convolutional neural network training. There are two different modes
available on this platform: training mode and autonomous mode. In training mode,
we can manipulate the car through keyboard input and record the needed driving
behavior as a sequence of images with labels. In autonomous mode, we upload
our trained network and test its self-driving performance. At the same time, the
platform acts as a server to collect the output data for analysis. Three cameras are
mounted on the left, right, and center of the car's front windshield to collect the data.
While driving, time-stamped images from those three cameras and the corresponding
labels of speed, steering angle, acceleration rate, and braking rate are recorded.
The maximum speed and the maximum steering angle of our simulation car are
30 miles per hour and 25° respectively. The collected data is then preprocessed
and augmented as the input data and sent into our neural network. The predicted
steering angle with throttle values is then collected as output data by the platform
to drive the car autonomously see Figure 6.2.

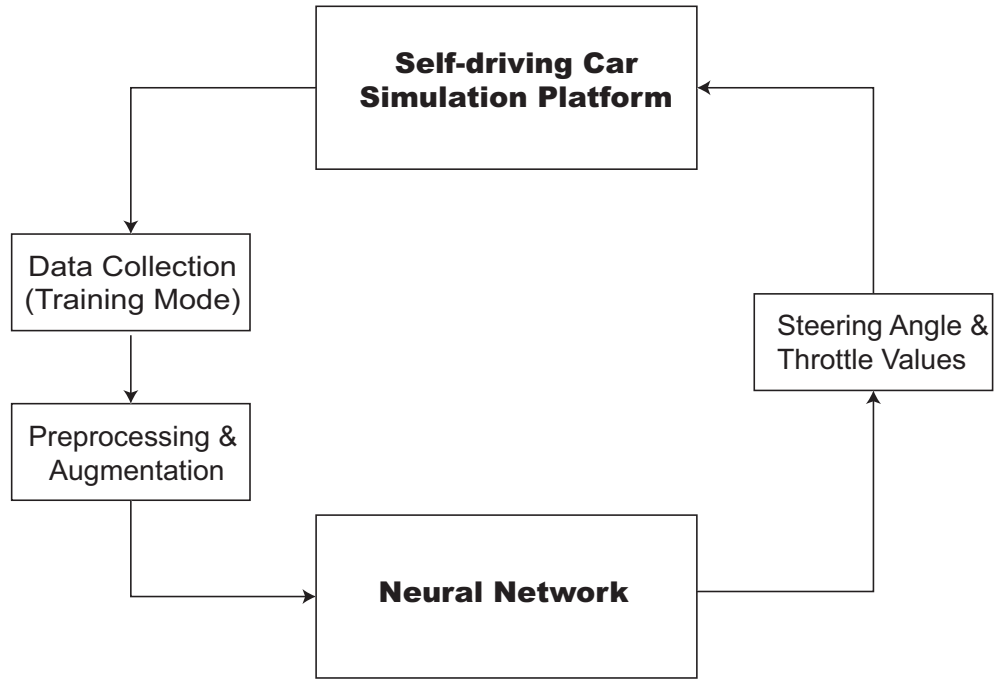

Figure 6.2: The system overview

## 6.3 Network Architecture

The artificial neural network structure is inspired by the human brain's neural structure, Figure 6.3(a) illustrate a simple standard neural network that can transfer information by layers of interconnected neurons. The neurons send their values layer by layer. Firstly, they go through the input layer on the left-hand side. Then, the second layer, which is a hidden layer. Although there is only one hidden layer, the number can be different. Finally, the neuron in the output layer represents the output of the network. We call the network in Figure 6.3(a) a fully connected neural network.

As there is a weight in each connection, the input of the next layer is the product of the weight and the output of the previous layer. In the training process, the weights among those connections are adjusted to optimize the network's performance. In general, we use a loss function to measure the error of a network. In this chapter, the weights of the network are adjusted by the mean square error methods, which minimizes the error between the predicted command from our network and the real steering command. The network architecture consists of one normalization layers as the first layer followed by four convolutional layers, a flatten layer, and three dense layers (Figure 6.4). The captured driving image with three channels is fed as the input, and the predicted steering angle is the output of the network.

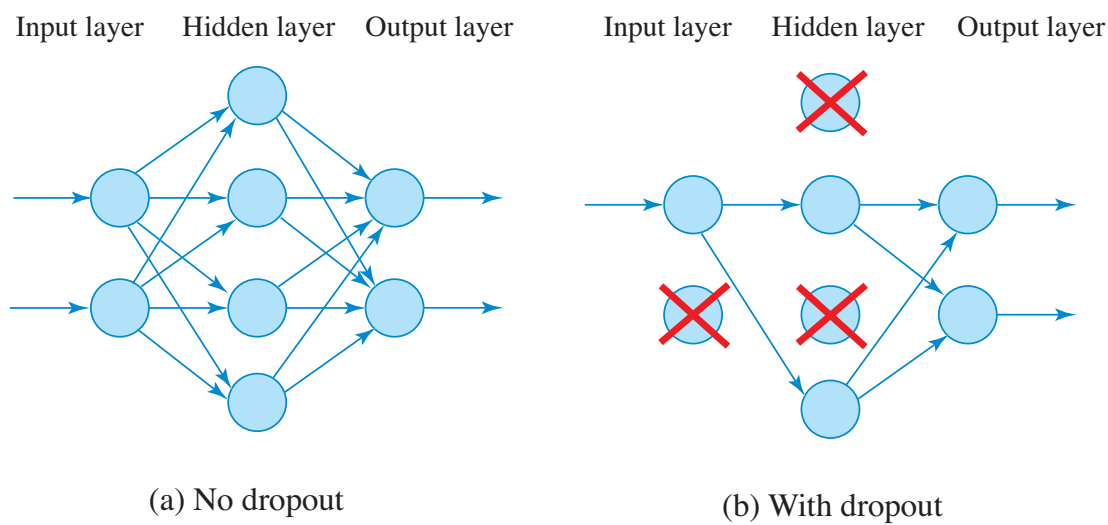

(a) No dropout                      (b) With dropout

Figure 6.3: The standard neural network (no dropout); and the neural network after dropout

The normalization layer unifies the dimension by normalizing all images with value pixels from -1 to 1. Meanwhile, the steering angles are also normalized to values between -1(far right) and 1(far left). This normalization process accelerates the network via GPU processing and allows the normalization scheme to be altered with the network architecture. The input images are then split and forwarded to the network.

In order to extract features, the convolutional layers are determined empirically

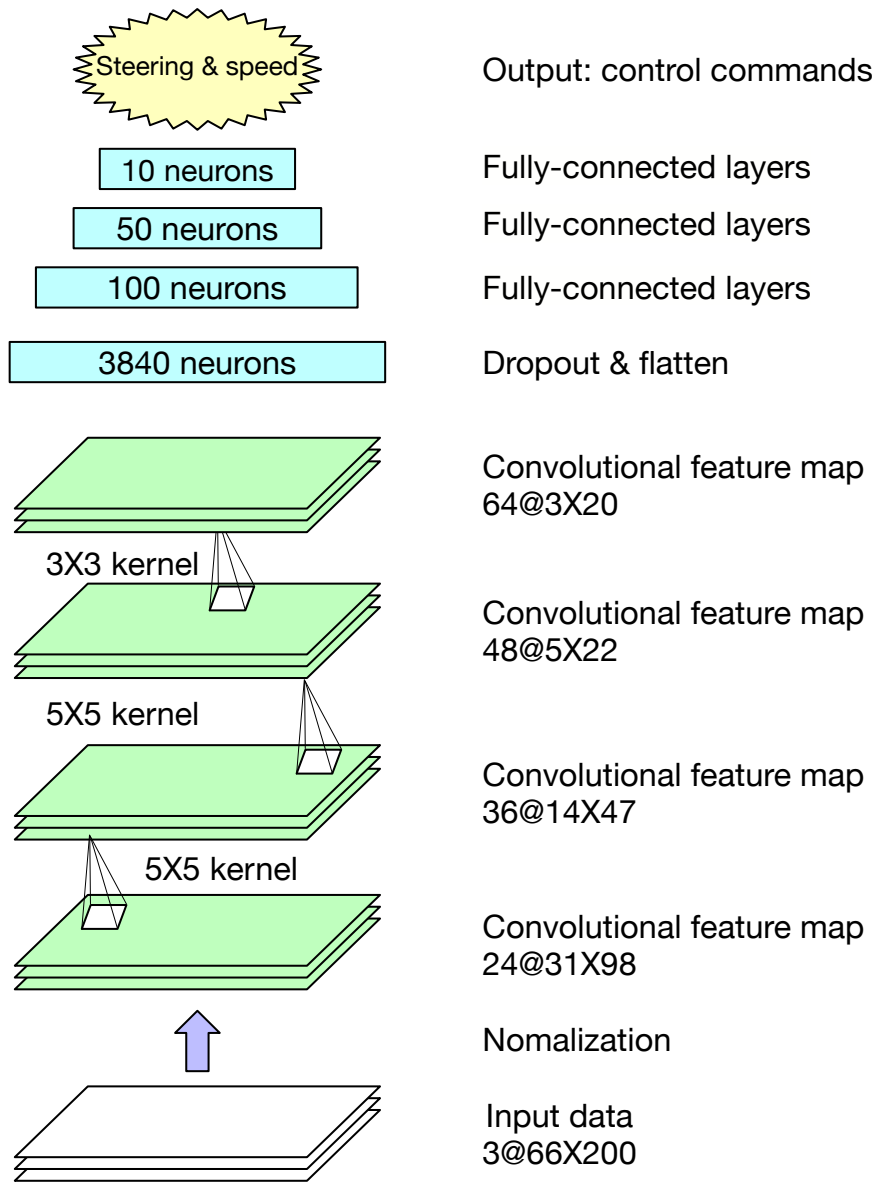

Output: control commands

Fully-connected layers

Fully-connected layers

Fully-connected layers

Dropout & flatten

Convolutional feature map
64@3X20

Convolutional feature map
48@5X22

Convolutional feature map
36@14X47

Convolutional feature map
24@31X98

Nomalization

Input data
3@66X200

Figure 6.4: The network architecture

by a series of experiments with different layer configurations. The first three layers
use strided convolutions with a $5 \times 5$ size kernel with stride of $2 \times 2$, and the last
one has a $3 \times 3$ size kernel with non-stride. The depths of each layer are 24, 36,
48, and 64, respectively. Then the multidimensional output generated from the last
convolutional layer converts to a one dimensional with 3840 neurons by a flatten
layer. The last part of our neural network is three dense layers, which are developed
with gradually reducing sizes: 100 50, 10 neurons, respectively. The output control
value contains two elements, one is steering angle, and another is throttle value
(Figure 6.5).

| Layer (type) | Output Shape | Param # |
|---|---|---|
| conv2d_8 (Conv2D) | (None, 31, 98, 24) | 1824 |
| conv2d_9 (Conv2D) | (None, 14, 47, 36) | 21636 |
| conv2d_10 (Conv2D) | (None, 5, 22, 48) | 43248 |
| conv2d_11 (Conv2D) | (None, 3, 20, 64) | 27712 |
| dropout_6 (Dropout) | (None, 3, 20, 64) | 0 |
| flatten_2 (Flatten) | (None, 3840) | 0 |
| dense_8 (Dense) | (None, 100) | 384100 |
| dropout_7 (Dropout) | (None, 100) | 0 |
| dense_9 (Dense) | (None, 50) | 5050 |
| dropout_8 (Dropout) | (None, 50) | 0 |
| dense_10 (Dense) | (None, 10) | 510 |
| dropout_9 (Dropout) | (None, 10) | 0 |
| dense_11 (Dense) | (None, 1) | 11 |

Total params: 484,091
Trainable params: 484,091
Non-trainable params: 0
--

Figure 6.5: The summary of CNN

To make the network architecture more robust and prevent overfitting, there is a dropout layer with a dropout rate of 50% implemented after flatten and fully connected layers. The key idea behind the dropout layer is to drop out some neurons from the network during training randomly (see Figure 6.3(b)). This procedure significantly reduces overfitting and prevent co-adapting between neurons, furthermore, improves the performance of neural networks [268].

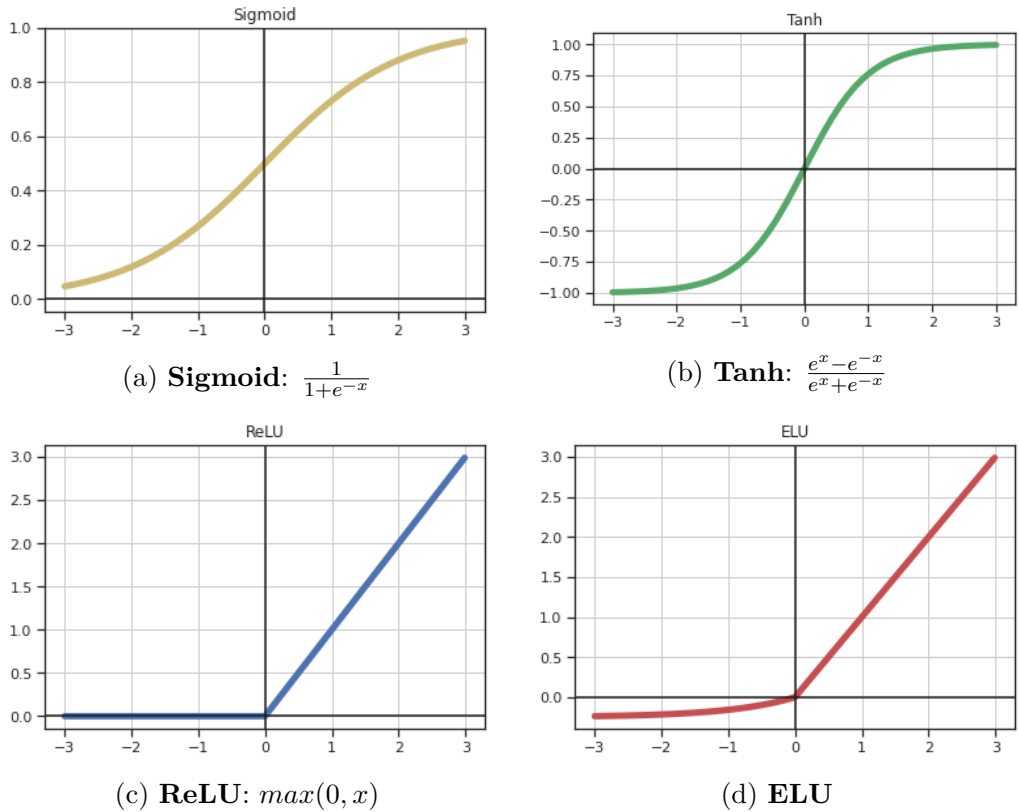

(a) **Sigmoid**: $\frac{1}{1+e^{-x}}$

(b) **Tanh**: $\frac{e^x - e^{-x}}{e^x + e^{-x}}$

(c) **ReLU**: $max(0, x)$

(d) **ELU**

Figure 6.6: Activation functions

In addition, all the convolutional layers are uniformly activated by ELU functions [154] for more nonlinearity in the network. ELU, also known as Exponential Linear Unit, is an activation function that tends to converge more quickly and produces more accurate results. Both ReLU (Rectified Linear Unit) [155] and ELU activation functions mitigate the problem of exploding or vanishing gradients. For the positive region, the functions simply outputs $x$. ReLU, ELU has non-zero values in the negative region, which means it can decrease the error by fixing the weight parameters, and more robust to input changes or noise. Furthermore, it does not suffer from the problem of dying neurons. ELU can obtain higher classification accuracy than ReLU, Sigmoid [156], and tanh [157] (see Figure 6.6).

$$ELU(x) = \begin{cases} x; x \geq 0 \\ \alpha(e^x - 1); \text{x} < 0 \end{cases} \tag{6.1}$$

where $\alpha$ is a hyper-parameter, and $\alpha \geq 0$.

The mean square error method is used as the loss function to minimize the error between the steering command from our network and the actual command and train the weights of our network. The loss function is:

$$MSE = \frac{1}{n} \sum (y_i - \hat{y}_i)^2 \tag{6.2}$$

where $y_i$ and $\hat{y}_i$ are the i-th recorded steering angle and the predicted steering angle, respectively.

## 6.4 Simulation Results

In this section, computer simulations are carried out to validate the performance of the proposed model considering different conditions and environments. We use both desert and mountain tracks in our simulations. The desert track is used for training purposes, and the desert and mountain tracks are used for testing (Figure 6.7).

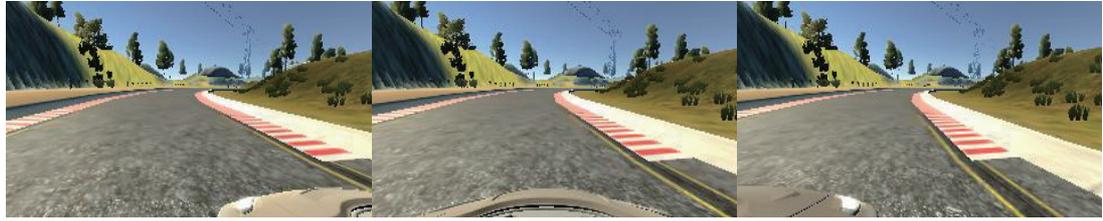

(a) Training environment: desert road

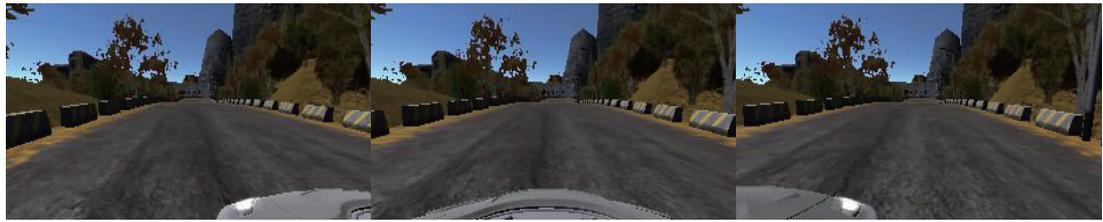

(b) Testing environment: mountain road

Figure 6.7: Images from left, center and right camera

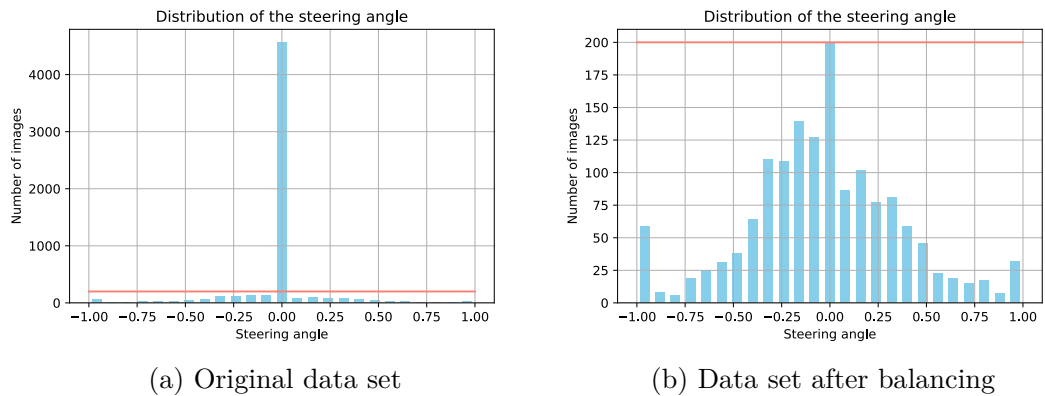

(a) Original data set       (b) Data set after balancing

Figure 6.8: Steering angle distribution histogram

Firstly, the data is collected by driving the vehicle manually in the desert track environment two laps forward and reverse, which can eliminate steering bias from driving in one direction. While driving, the time-stamped image frames are taken by cameras at the left, right, and center of the car's front windshield (see Figure 6.7(a)). Along with the images, the corresponding labels of speed, steering angle, acceleration rate, and braking rate are also collected in our data set. The total amount of data collected is 5,864, which consists of 17,592 individual frames from three cameras with corresponding labels.

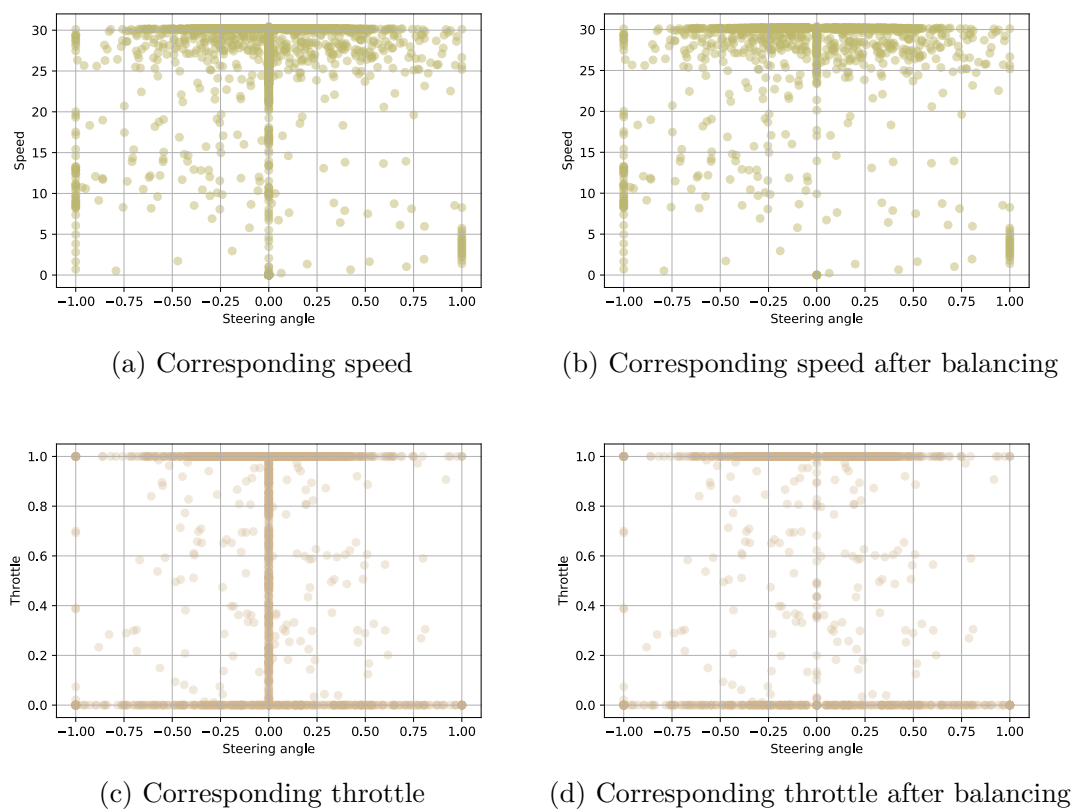

(a) Corresponding speed

(b) Corresponding speed after balancing

(c) Corresponding throttle

(d) Corresponding throttle after balancing

Figure 6.9: Distribution of the corresponding speed and throttle value

The resolution of the original frame is 160 x 320 pixels. The distribution of the steering angle is centered at 0° (go straight) and divided into 25 equal portions shown in Figure 6.8(a). And then, we set the threshold value (red line) at 400 to balance the overlarge portion. After balancing, 4,365 images are removed, while 1,499 images remain in our data set. The distribution of the steering angle after balancing is shown in Figure 6.8(b) and Table 6.1, and we can see that the majority of them fall upon 0°. The distribution of the corresponding speed and throttle values is shown in Figure 6.9. The image frames are further split randomly into 1,199 training samples and 300 validation samples in an 80/20 fashion (Figure 6.10).

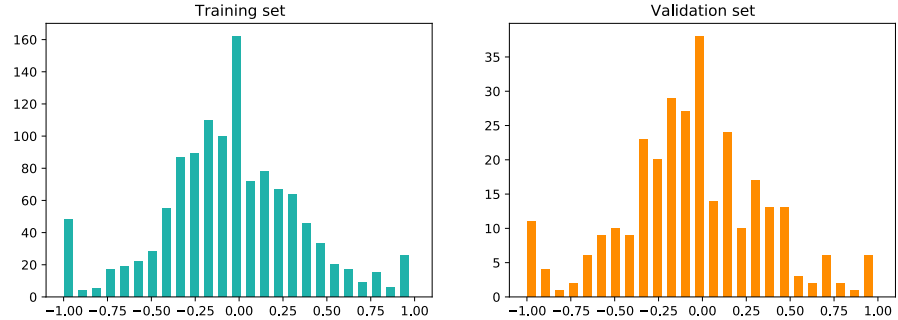

Figure 6.10: Steering angle distribution of two sets

It is clear that these operations deteriorate the lack of data to train the model with promising accuracy for testing. Hence data augmentation is applied to increase the instance of the data. It can help the model to extract more information and save computation power for a faster response. Besides, the data augmentation methods generate more generalized data without hand-coding individual features and rules. These methods add randomness and rearrange the sequence of data in the model train itself rather than simply memorizing the data.

The applied data augmentation techniques include flipping, panning, brightness change, and zooming.

- **Flipping:** By flipping frames and inverting the predicted steering angle's signs, we expand the data with instances in the reverse direction. Therefore, instances of different output can be taken place equally and prevent any bias on left or right (Figure 6.11(a)).

- **Panning/Shifting:** The images can be shifted/panned by -10 to +10% on both x- and y-axis independently to simulate the conditions of driving at different positions of the track. And the corresponding offset is added to the steering angle (Figure 6.11(b)).

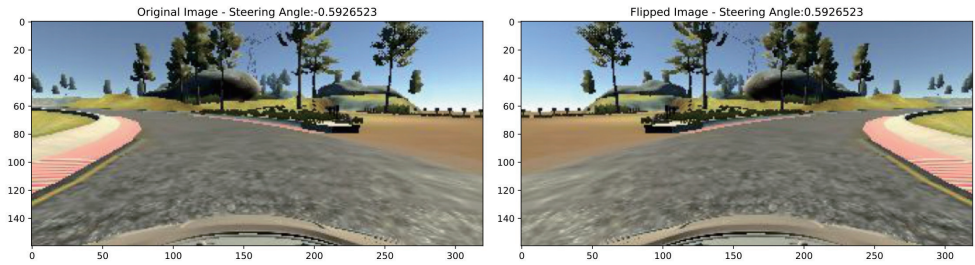

(a) Flipping

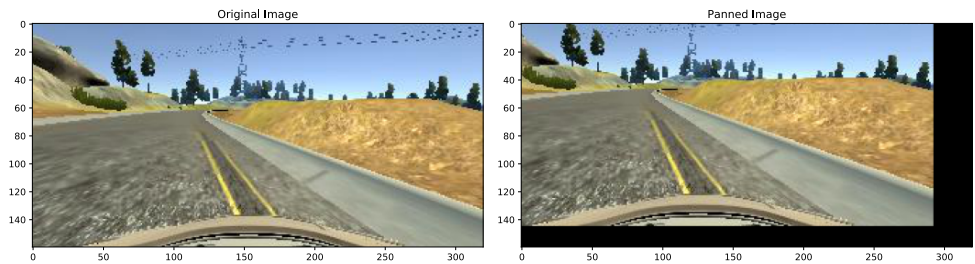

(b) Panning

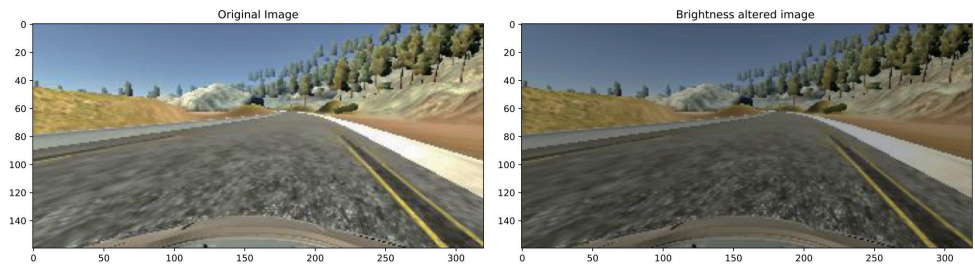

(c) Brightness augmentation

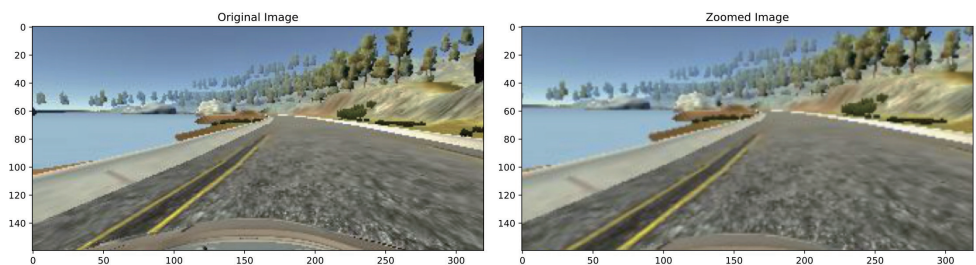

(d) Zooming

Figure 6.11: Data augmentation methods

- **Brightness:** The brightness technique increases current data by randomly changing the brightness of the images to simulate different kinds of weather (Figure 6.11(c)).

- **Zooming:** We set the zooming in to a value of 100% to 150% of the original size (Figure 6.11(d)).

After that, we do image preprocessing to calibrate images and pass them into our neural network for training. We cropped the sky and the car hood out of the image, and the new size is 60 x 200 pixels. By doing so, the computer can focus on the important parts of the image with faster processing speed. Additionally, we smooth them with Gaussian blur and convert color-space from RBG to YUV. Figure 6.12 shows an example of an original frame and the corresponding preprocessed image.

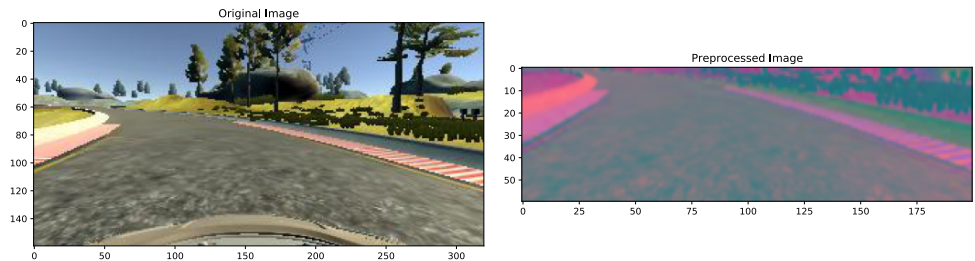

Figure 6.12: Original image and preprocessed image

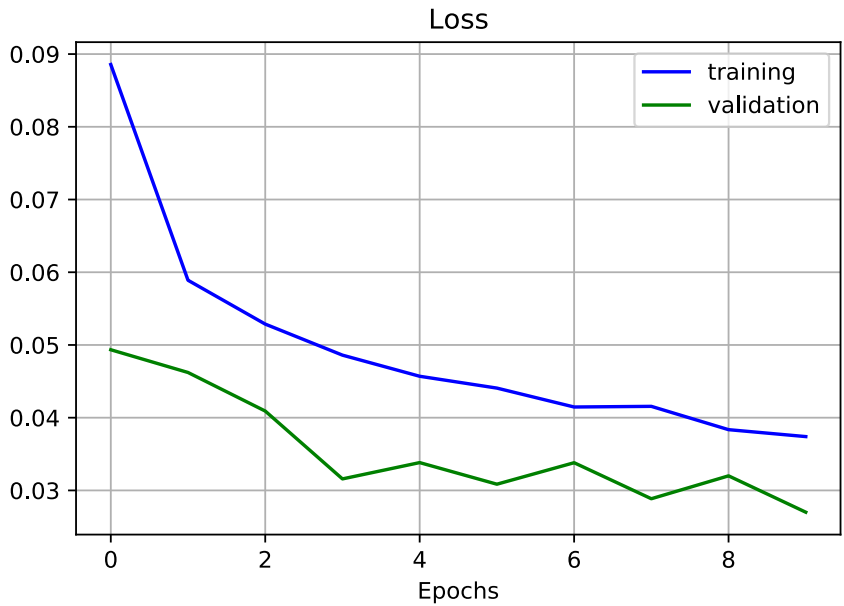

Figure 6.13: MSE Loss

Our neural network is trained on the GPU with TensorFlow and Keras by back-end Colaboratory. To optimize this loss, the Adam [168] is applied as our optimizer with 0.001 learning rate, and trained with batch size 300 in 10 epochs in Table 6.2. The best result among all epochs is around 0.028 (Figure 6.13). After training, the network is able to automatically predict the steering angles to navigate the vehicle on the track without any collisions. The trained model is then tested in both desert track and the unseen mountain track environments. The performance and effectiveness of our model have been confirmed by simulation results, the video can also be viewed on the YouTube website via the link at `https://youtu.be/z7t8FQQ7Ohg`, `https://youtu.be/Z6pAQsrdoPE`. It shows that the trained network can make driving commands even on a track that has never been seen before.The satisfactory performance also demonstrates the accuracy of the predicted driving commands.

Table 6.1: Steering angle after balancing

| Statistic | Value |
|-----------|-------|
| Maximum | 1 |
| Minimum | -1 |
| Mean | -0.0358 |
| Std. dev. | 0.4030 |

Table 6.2: Training parameters

| Parameter | Value |
|-----------|-------|
| Batch size | 100 |
| Epoch | 30 |
| Learning rate | 0.001 |

## 6.5 Summary

This chapter proposed a convolutional neural network (CNN) model to predict the steering angle from images taken by the center camera to drive a car autonomously. With the minimum training data from a human driver, the model learns how to drive in a known or unknown environment. The input data is captured by three cameras present on the right, left, and center of the vehicle's front windshield, and then preprocessed and passed through the convolutional neural network for training. Features can be extracted automatically for vehicle localization and used to make accurate predictions. The concept of back propagation adjusts our CNN model's weights to improve the accuracy of the CNN model. The trained model is then used to drive the car autonomously by predicting the steering angle of the car with the input images captured in real-time.

Compared with decomposing this problem into smaller pieces, such as lane detection, path planning, and control, this model optimizes all processing steps simultaneously. One of the main advantages of the proposed method is to eliminate the need for hand-coding tremendous amounts of different features and rules, and instead, we develop a model that learns how to drive by observing the human driver.

The simulation results show that our trained model can drive autonomously in both training and testing environments. With the satisfactory performance of the proposed convolutional neural network, we can see that the autonomous vehicle could be a possible remedy for traffic accidents. Despite that, we also see the need for further research on more challenging environments to accomplish the goal of full automation for public transportation.

# Chapter 7

# Conclusion and Future Directions

As one of the most essential and important abilities of the autonomous mobile robot, path planning has been and still is the focus of extensive research. Most of this research centers around techniques for decision making (learning and planning) to improve the performance of autonomous mobile robots, so as to enable robots to act intelligently even without complete knowledge of their environment. The main applications include planetary exploration, patrolling, navigation systems towards UAVs for reconnaissance and surveillance duty and autonomous vehicles for and rescue, delivering goods and merchandise, industrial automation, construction, entertainment, etc. This chapter summarises the research in this report and presents the possible future extensions.

## 7.1   Summary and Contribution

In this section, we conclude the report by highlighting the contributions. Three research questions are presented in Chapter 1. The first problem is the navigation of the ground mobile robot in an unknown environment, and how to improve the effectiveness and efficiency of generating the path. The second problem is the coverage problem of UAV deployment with the influence of obstruction. The third problem is about the path planning of autonomous vehicles, especially by extracting the characteristics of human drivers to predict the steering angle. In Chapter 2, we conduct a systematic review of state-of-the-art literature, specifically about path planning methods, different types of mobile robots, and machine learning techniques.

As mentioned in the previous chapter, the existing algorithms for robot path planning control can be classified into two main categories: global and reactive (local) [61]. In the global algorithms, the information of the environment is known *a priori*, and used to find the best possible solution [199, 269]. However, the major drawback of global path planning is its inability to deal with the uncertainty in the environment. Nevertheless, the dependence on environmental information and the computation complexity highly affect the final path planning performance. To solve the first question, Chapter 3 illustrates two collision-free navigation algorithms in unknown 2D and 3D environments, respectively. The proposed hybrid reactive path planning algorithm simplifies the completed challenge into policy selection. By introducing Q-learning, the robot "learns" the optimal policy by interacting with the environment. Simulations were carried out on V-REP, interfaced with MATLAB and Python. In Chapter 4, a new method is developed for mobile robots arriving at the optimal path without any collisions in an unknown dynamic environment. Unlike the other control algorithms cited in the literature, our approach can be applied to an unknown environment with dynamic obstacles with arbitrary or even time-varying shapes. It is not necessary to separate or approximate the shape of obstacles with polygons or disks. Furthermore, no deterministic knowledge of the obstacle velocity or even a moderate rate of its change is needed in our integrated environment representation and reinforcement learning-based control algorithm. The required comprehensive environmental information can be easily obtained. It involves Q-learning to construct the optimal path selection, and it is found to be both efficient and effective in terms of the shortest path, time. A large number of computer simulations verify the performance of the algorithm in different scenarios. The excellent performance of the proposed algorithm is shown in terms of path optimality when compared with the conventional control methods mentioned in this chapter.

Chapter 5 develops two coverage control algorithms for drones to tackle the second problem. The proposed methodology has addressed UAV kinematic constraints and camera sensing constraints into consideration. In other words, our algorithm solves the problem caused by occlusions of the obstacles, like vegetation, hills, buildings, walls, etc. The drones equipped with our proposed algorithm and camera can fly into bushfire-prone areas to conduct complete/full coverage, which means every point on the target area can be seen at least once in one complete surveillance circle. Similar application scenarios also include wildlife monitoring and conservation, intrusion detection, and aerial distribution of agricultural chemicals.

In order to promote the social transformation from human drivers to self-driving vehicles, a convolutional neural network is developed in Chapter 6 to extract the features from the images and find the dependence of predicting steering angle and throttle values of self-driving vehicle. In addition, the performance of the developed system has been verified and evaluated through a large number of tests in different scenarios.

## 7.2 Future Work

The COVID-19 pandemic is already ushering in a host of challenges to international industrial manufacturers. However, many countries still have significant strengths to take advantage of future commercial opportunities. It is a critical time to explore a proactive deployment of automation and robotics technologies. This work opens up several potential directions to do our part for this new trend. In this section, we summarize some directions of future work.

In Chapter 3 and 4, an important direction for future research is to extend the obtained results to the case on a team consisting of several robots, for instance, in the problem of navigation for sweep coverage in cluttered environments [270]. Furthermore, implementing the proposed methods in the real robots and conducting experiments in real-time testbeds. We believe that the new experimental data could provide more insights to improve the proposed algorithms.

In Chapter 5, we aim to develop techniques for coverage control of mobile networks, especially drones. The developed algorithm can be implemented in a prototype of an early bushfire detection system, which picks up changes in the atmosphere that will likely cause a bushfire. One interesting direction for future research is to take the realistic pan-tilt capability of the onboard camera into consideration. In other words, adding the flexibility in the camera's orientation [263]. Another future direction might be to apply the coverage approaches in a geographically vast area, employing a network of UAVs. Three most common coverage problems are blanket coverage [271], barrier coverage [272], and sweep coverage [273]. It would be interesting to consider ways to combine current coverage algorithms with collision avoidance strategies in cluttered environments [222, 274, 275].

In Chapter 6, we will keep continuous attention to the state-of-the-art bioinspired intelligent self-driving algorithms. There will be more related research conducted to facilitate the transition from human drivers to autonomous vehicles in the future. For example, the last-mile delivery and viable self-driving haulage vehicles to transport material in practice.

In addition, another direction for future research is to combine the proposed autonomous navigation algorithms with advanced method of robust control and state estimation in vehicle systems such as H-infinity based robust control [276–279], sliding mode control [280–284], robust Kalman state estimation [285–290], and hybrid dynamical systems/switched controller systems [291–296].

The dynamics of longitudinal vehicle behaviors are also an interesting aspect to investigate the future mixed-autonomy traffic. A platoon of vehicles are said to be stable if the state fluctuations (e.g., position error) of vehicles are not amplified when propagating along the vehicle stream [297]. A vast amount of literature has shown it can be influenced by the limitation of the vehicle's acceleration, and driver's timid or aggressive accelerations may contribute to the stop-and-go waves that produce traffic flow oscillations or even traffic congestion. The dynamics of longitudinal vehicle behaviors have a microscopic impact on one following vehicle, but also lead to more significant impact on reducing traffic congestion, enhancing highway safety, improving traffic efficiency, and reducing fuel consumption [298]. It is therefore essential to have a better understanding of vehicle longitudinal behaviors in order to predict their impacts. Compared with human drivers, the autonomous vehicle has potential advantages in behavior prediction and impact analysis.

The new trends of automation and robotics are led by artificial intelligence, autonomous driving, network communication, cooperative work, nanorobotics, friendly human-robot interfaces, safe human-robot interaction, and emotion expression and perception [299]. Meanwhile, the automation and robotics and other fields are complementary. These new trends can boost the development of fields like health care, logistics, and industry, and then feed back to itself.